\RequirePackage{fix-cm}
\documentclass[smallextended]{svjour3}       % onecolumn (second format)
\smartqed  % flush right qed marks, e.g. at end of proof
\usepackage{graphicx}
\graphicspath{{Figures/}} % Specifies the directory where pictures are stored
\usepackage{amsmath}
\usepackage{amsfonts}
\usepackage{enumitem}
\usepackage{algorithm}
\usepackage{algorithmic}
\usepackage{booktabs}  
\usepackage{footnote}
\usepackage{rotating}
\usepackage{lscape}
\usepackage[square,numbers]{natbib}

\usepackage{caption}
\usepackage{subcaption}
\newcommand*\xbar[1]{%
	\hbox{%
		\vbox{%
			\hrule height 0.5pt % The actual bar
			\kern0.4ex% % Distance between bar and symbol
			\hbox{%
				\kern-0.15em% % Shortening on the left side
				\ensuremath{#1}%
				\kern-0.16em% % Shortening on the right side
			}%
		}%
	}%
} 
\journalname{arXiv.org}
\begin{document}
	\raggedbottom

\title{Change Detection in Noisy Dynamic Networks: A Spectral Embedding Approach%\thanks{Grants or other notes
}


\author{Isuru Udayangani Hewapathirana \textsuperscript{1}         \and
        Dominic Lee \textsuperscript{2}\and
        Elena Moltchanova \textsuperscript{2}\and
        Jeanette McLeod \textsuperscript{2}
         %etc.
}

\authorrunning{Hewapathirana et al.} % if too long for running head

\institute{1. \at
              Faculty of Science, \\
              University of Kelaniya, \\
              Sri Lanka. \\
              \email{ihewapathirana@kln.ac.lk}           %  \\
           \and
           2. \at
              School of Mathematics and Statistics, \\
              University of Canterbury, \\
              New Zealand.
}

\date{}

\maketitle

\begin{abstract}
Change detection in dynamic networks is an important problem in many areas, such as fraud detection, cyber intrusion detection and health care monitoring. It is a challenging problem because it involves a time sequence of graphs, each of which is usually very large and sparse with heterogeneous vertex degrees, resulting in a complex, high dimensional mathematical object. Spectral embedding methods provide an effective way to transform a graph to a lower dimensional latent Euclidean space that preserves the underlying structure of the network. Although change detection methods that use spectral embedding are available, they do not address sparsity and degree heterogeneity that usually occur in noisy real-world graphs and a majority of these methods focus on changes in the behaviour of the overall network. 

In this paper, we adapt previously developed techniques in spectral graph theory and propose a novel concept of applying Procrustes techniques to embedded points for vertices in a graph to detect changes in entity behaviour. Our spectral embedding approach not only addresses sparsity and degree heterogeneity issues, but also obtains an estimate of the appropriate embedding dimension. We call this method CDP (change detection using Procrustes analysis). We demonstrate the performance of CDP through extensive simulation experiments and a real-world application. CDP successfully detects various types of vertex-based changes including (i) changes in vertex degree, (ii) changes in community membership of vertices, and (iii) unusual increase or decrease in edge weight between vertices. The change detection performance of CDP is compared with two other baseline methods that employ alternative spectral embedding approaches. In both cases, CDP generally shows superior performance.

\end{abstract}

\section{Introduction}
\label{intro}
A network is a collection of entities, that have inherent relationships. Some examples include a social network of friendships among people, a communication network of company employees connected by phone calls, emails or text messages, and a biological network of neurons connected by their synapses.  A network can be mathematically conceptualized as a graph by associating entities with vertices, and relationships with edges connecting vertices in the graph. For example, in the graph representation of a social network like Facebook, vertices may represent friends and edges represent friendship connections.

Most real-world networks evolve as time progresses. That is, the entities and their relationships keep evolving with time. This type of relational data can be represented as a dynamic network. For example, a communication network of a company is a dynamic network because new employees (entities) join the network and communication patterns (relationships) are modified continuously. Although both the entities and the relationships in a network can vary over time, in this paper, we assume that a dynamic network consists of a fixed set of entities with time varying relationships between them. A dynamic network can be represented as a time sequence of graphs, each representing the entities (as vertices) and their relationships (as edges) at a given time instant. \textit{Change detection} is the process of continuously monitoring a dynamic network for deviations in entities and their relationship structure. A clear illustration of the change detection process based on a toy example is given in \cite{hewapathirana2019change}. Given a dynamic network conceptualized as a time sequence of undirected, weighted graphs, we address the problem of detecting vertex-based changes at each time instant. Detecting vertex-based changes is important in areas such as fraud detection, cyber intrusion detection and spam detection. For example, consider the time varying email communications between a set of employees in an organisation. A sudden collaboration between a set of employees who rarely communicated during the recent past, may indicate some unusual motivation or a major event involving the organisation \cite{sricharan2014localizing}. Such changes in entity behaviour can be detected by monitoring the behaviour of vertices in the corresponding sequence of graphs.  

\begin{figure}
	\centering
	\begin{subfigure}{.5\textwidth}
		\centering
		\includegraphics[trim = 70mm 0mm 50mm 0mm, scale=0.5]{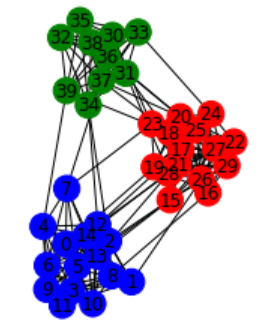}
		\caption{Original graph.}
		\label{Illustrate_embedding1}
	\end{subfigure}%
	\begin{subfigure}{.5\textwidth}
		\centering
		\includegraphics[trim = 50mm 0mm 50mm 0mm, scale=0.5]{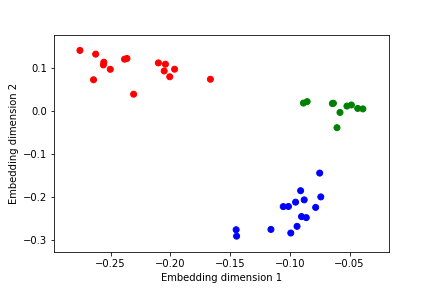}
		\caption{Two-dimensional embedding.}
		\label{Illustrate_embedding2}
	\end{subfigure}
	\caption{\textbf{Illustration of an embedding of a network using a toy example}. The two-dimensional embedding preserves the edge-based closeness in the original graph.}
	\label{Illustrate_embedding}
\end{figure}

Monitoring the behaviour of every vertex in the graph is a challenging problem because each graph in the time sequence contains a large number of vertices resulting in a high-dimensional mathematical object. Spectral embedding methods provide an effective solution to the high dimensionality problem. These methods can be used to obtain a low dimensional representation of the graph that excludes noise and redundant information and retain important structural information \cite{skillicorn2007understanding}. Our goal for spectral embedding is to obtain a low dimensional representation of vertices which maintains their edge-based closeness in the graph. In Figure \ref{Illustrate_embedding}, we give an illustration of an embedding of a small graph. The left figure (a) shows a graph where the length of each edge is drawn proportionally to the closeness between the corresponding pair of vertices. We can observe three clusters of vertices in this graph. The right figure (b) gives the two-dimensional embedding, where each vertex is represented as a point in a two dimensional Euclidean space. We can see how the edge-based closeness of vertices in the graph in (a) is maintained by the embedded points in (b). This characteristic emphasizes the \textit{clustering} property of the embedded points \cite{saerens2004principal}.

%\begin{figure}
%% Use the relevant command to insert your figure file.
%% For example, with the graphicx package use
%	\includegraphics[trim = 0mm 0mm 50mm 0mm, scale=0.6]{Figures/graphVis}
%\caption{Small graph where the length of each edge is drawn proportionally to the closeness between
%	the corresponding pair of vertices}
%\label{fig:embeddingIllustrate}       % Give a unique label
%\end{figure}

In literature, we can find numerous approaches that detect vertex-based changes in a time series of graphs \cite{neil2013scan,heard2010bayesian,papadimitriou2010web,priebe2005scan,gupta2012integrating,yu2018detecting}. However, only a few utilize spectral methods. For example,  \cite{akoglu2010event,ide2007computing,sun2008less} apply matrix-based spectral embedding while \cite{sun2006window,papalexakis2012parcube} use a tensor-based spectral embedding method. The majority of the real-world graphs are sparse and contain vertices with heterogeneous
degrees \cite{sengupta2015Hetro}. Currently available spectral-based change detection methods do not simultaneously address sparsity and degree heterogeneity issues prior to obtaining an embedding from the graph. Consequently, changes involving only a few vertices, or changes
involving low degree vertices, tend to be missed by these methods.

In this paper, we propose a novel method called CDP (change detection using Procrustes analysis) to detect changes in vertex behaviour. In our method, we first obtain a low dimensional embedding from the weighted adjacency matrix representing the graph at each time instant. Each embedded point characterizes the behaviour of a vertex in the graph at a given time instant. We use statistical Procrustes analysis techniques \cite{dryden1998statistical} to compare embeddings across time instants and calculate change scores for vertices. We evaluate the performance of CDP using extensive simulation experiments and the dynamic network for the Enron email dataset \cite{klimt2004introducing}. By carefully structuring the simulation experiments, we fully evaluate the performance of the method in detecting various types of changes that occur in real world networks. In all our experiments, we formally compare CDP to two other methods. Based on the results, we conclude that CDP efficiently and effectively identifies various vertex-based changes that are considered in our experiments.

The rest of the paper is organized as follows. We first provide a brief overview of our overall change detection method in Section \ref{Brief Overview}. In Section \ref{Problem Framework}, we provide a detailed description of our change detection framework. In Section \ref{CDP Algorithm}, we summarize our change detection procedure and present the CDP algorithm. We evaluate the performance of CDP using simulation experiments (Section \ref{Simulation Experiments}) and a real-world application (Section \ref{Real World Data}). In each experiment, the performance of CDP is compared with two other change detection approaches which are discussed in Section \ref{Comparison Methods}. Finally, we conclude by summarizing our findings in Section \ref{Summary}. 

\section{Brief Overview}
\label{Brief Overview}

Our proposed method, CDP (change detection using Procrustes analysis), aims to detect vertex-based changes in a dynamic network. A dynamic network is represented  as a time sequence of undirected graphs, where each graph is then represented as a symmetric, weighted adjacency matrix. We apply spectral methods to the weighted adjacency matrix and embed the vertices into a $d$-dimensional Euclidean space that preserves the {closeness} between vertices in the original graph representation. The embedded points also highlight important vertex properties such as transitivity, homophily by attributes, and clustering, that are present in most real-world graphs \cite{hoff2002latent,nickel2007random}. In this paper, we define these embedded points as \textit{features} for vertices characterizing vertex behaviour at each time instant. Vertices in sparse and heterogeneous graphs depict entities with different abilities to establish connections. It is difficult to achieve a good representation if we ignore sparseness and degree heterogeneity when obtaining a low dimensional embedding \cite{joseph2013impact}. By employing ideas from spectral graph theory \cite{chung1997spectral}, combined with the graph regularization technique introduced in \cite{amini2013pseudo}, we formulate a strategy to effectively embed sparse and heterogeneous graphs into low dimensional Euclidean spaces. It is important to identify an optimum value for the low dimension $d$ in order to obtain a highly accurate representation of the inherent clusters of the data using the embedded space \cite{brand2003unifying}. CDP adapt the low-rank matrix approximation method in \cite{achlioptas2007fast} to automatically estimate the proper embedding dimension.

Generalized orthogonal Procrustes analysis (GPA) methods can be used to calculate an average from a set of matrices after removing Euclidean similarity transformations \cite{dryden1998statistical,stegmann2002brief}. We adjust the standard GPA technique to extract \textit{profile features} during the recent past time instants, and calculate change scores for vertices at each time instant. A {profile feature}, which is also a vector, represents the average behaviour of the vertex in the recent past time instants (previous $w$ time instants). Our idea of applying Procrustes analysis techniques to compare embeddings for the purpose of change detection in dynamic networks is new and is inspired by \cite{tang2012community}. Using a moving window approach, the change score calculation procedure is repeated over time to detect changes for all time instants.

\begin{figure}[htbp]
	\centering
	\includegraphics[trim = 10mm 0mm 0mm 0mm, scale=0.4]{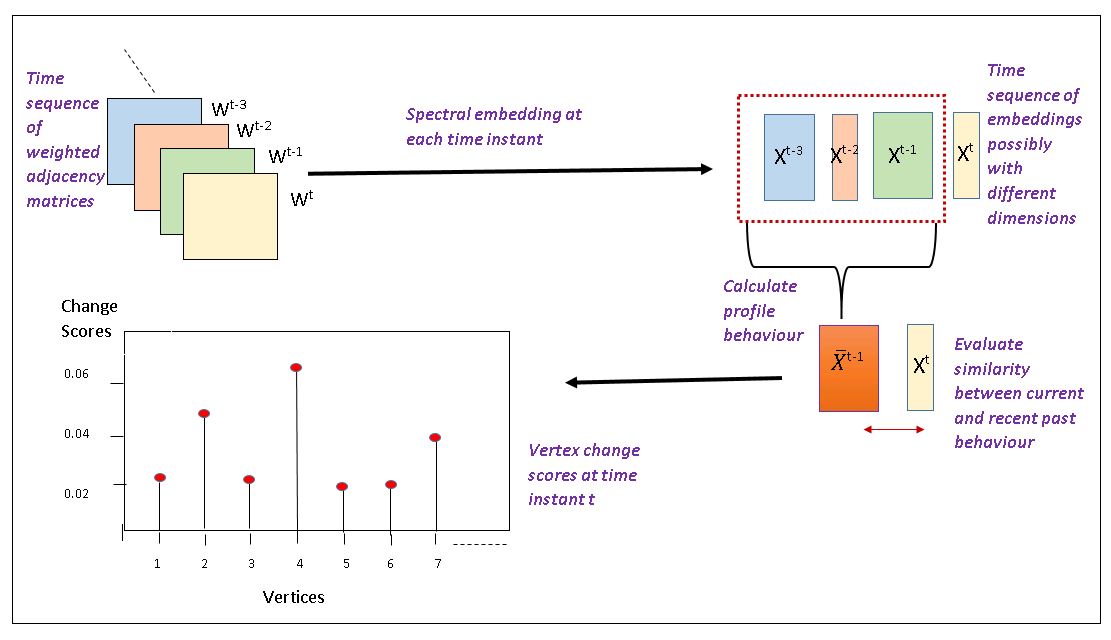}
	\caption{\textbf{Illustration of the overall CDP framework}. The time sequence of graphs is first represented as a time sequence of weighted adjacency matrices. At each time instant, we perform spectral embedding on the matrix and obtain an embedding where each row corresponds to a feature representing a vertex's behaviour. Next, we define a window of length, $w \in \mathbb{Z}_+$, over the
		previous $w$ embeddings, and use GPA to obtain the profile embedding, where each row corresponds to a vertex's profile feature. The dissimilarity between the current embedding and the profile embedding is then obtained to compute the change scores of the vertices at the current time instant. The window is moved along all preceding time instants to calculate vertex change scores for the whole time period.  \label{overall plan2}}
\end{figure}

Figure \ref{overall plan2} provides an illustration of the overall CDP framework. In order to evaluate the performance of CDP, we apply it to both synthetic and real-world datasets. We compare our method with two baseline change detection methods that are also based on different spectral embedding procedures. The results show that CDP performs better than the others in various change scenarios considered.

\section{Problem Framework}
\label{Problem Framework}

\subsection{Notation and Terminology}
\label{Notation and Terminology}

Let $G^1,G^2,\ldots,G^\mathcal{T}$ be a sequence of graphs defined over time instants, $t=1,2,\ldots,\mathcal{T}$. Each $G^t$ is a weighted and undirected graph with a fixed set of vertices, $V=\{v_1,\ldots v_n\}$. In our discussions, we also refer to $v_i$ as vertex $i$. Define the edge set  of graph, $G^t$, as $E^t$, where $|E^t|\leq n^2$, and $E^t$ contains edge, $e_{i,j}$, if there is an edge between vertex $i$ and vertex $j$. Each graph is represented by a symmetric weighted adjacency matrix, $W^t$, of dimension ${n \times n}$, where each element, $W_{i,j}^t\geq 0$. If $W_{i,j}^t=0$, then the vertices $i$ and $j$ are not connected in $G^t$. The degree of each vertex $i$ at time instant $t$ is defined as
\begin{equation*}
d_i^t=\sum_{j=1}^{n}W_{i,j}^t.
\end{equation*}
The degree matrix, $D^t$, is the diagonal matrix containing the vertex degrees, $d_1^t,\ldots,d_n^t$, on the diagonal. Let $\hat{\lambda}^t$ be the average vertex degree of graph, $G^t$, where $\hat{\lambda}^t=\frac{1}{n}\sum_{i}d_{i}^t$. From \cite{amini2013pseudo}, we define a network as sparse when $\hat{\lambda}^t<5$. 

\subsection{Problem Statement}
\label{Problem Statement}

At each time instant $t$, our goal is to calculate a change score for each $v_i$ in $G^t$, relative to the recent past behaviour. Our definition of the change score for $v_i$ at time instant $t$ is defined as follows.
\begin{definition}\label{def:changeScoreCh3} 
	The change score, $z^t_i$, for $v_i$ at time instant $t$ is 
	\begin{equation}
	z^t_i = f(\bar{\mathbf{x}}^{t-1}_i,\mathbf{x}^t_i),
	\end{equation}
	where $\mathbf{x}^t_i$ is the feature vector representing the behaviour of $v_i$ at time instant $t$, $\bar{\mathbf{x}}^{t-1}_i$ is the profile feature vector representing the behaviour of  $v_i$ in the recent past time instants, and $f$ is a dissimilarity function.
\end{definition}
According to this definition, our overall change detection procedure can be summarized as follows. 
\begin{enumerate}[noitemsep,topsep=0pt]
	\item Obtain a feature, $\mathbf{x}^t_i$, for $v_i$ from each $G^t$, where $\mathbf{x}^t_i~\in ~\mathbb{R}^{d^t}$ and $d^t \in \mathbb{Z}_+$.
	
	\item Obtain a profile feature, $\bar{\mathbf{x}}^{t-1}_i$, for $v_i$ from recent $w$ past time instants, $G^{t-w},\ldots,G^{t-1}$, where $\bar{\mathbf{x}}^{t-1}_i \in \mathbb{R}^{\bar{d}^{t-1}}$ and ${\bar{d}^{t-1}}~ \in \mathbb{Z}_+$.
	
	\item Calculate the dissimilarity between $\mathbf{x}^t_i$ and $\bar{\mathbf{x}}^{t-1}_i$, and obtain the \textit{change scores}, $z^t_i$, for $v_i \in V$ by using a suitable dissimilarity function $f$.
\end{enumerate}
In Sections \ref{Obtaining a Representative Summary at Each Time Instant}, \ref{Obtaining the Profile Behaviour at Each Time Instant}, and \ref{Change Score Calculation}, we discuss how these steps are implemented respectively.

\subsection{Feature Extraction at Each Time Instant}
\label{Obtaining a Representative Summary at Each Time Instant}

In this section, we formulate our spectral embedding strategy for each $G^t$ (Note that in this paper, for discussions focused on one time instant, we drop the superscript $t$ to simplify notation. For example, we use $W$ instead of $W^t$ to denote the matrix of $G^t$). The \textit{embedding} of a graph is an $n \times d$ matrix, where rows correspond to the $d-$dimensional embedded points for vertices. Our spectral embedding procedure consists of three main steps.

\begin{enumerate}
	\item Pre-processing the weighted adjacency matrix, $W$, of $G$.\\ 
	As we consider weighted, heterogeneous graphs, some edges possess considerably higher weights than the other edges and can turn out to be very influential during the embedding process. These edges are called \textit{dominant edges}. The elements of the corresponding weighted adjacency matrix, $W$, also show high variability. The presence of dominant edges may also hinder the detection of unusual edges that have lower weights, preventing the change from being detected. Applying a transformation on $W$, such as the logarithm, helps to mitigate this problem. After the log transformation, we scale each element, so that all elements in the resulting matrix are between zero and one. Below we state our two preprocessing steps in more detail.
	
\begin{enumerate}
	\item Apply a log transformation to each element in $W$, and obtain $\Ddot{W}$, where 
	\begin{equation}
	\label{logtransform}
	\Ddot{W}_{i,j} = \log_{10}(W_{i,j} + 1) \quad \forall i,j \in \{1,\ldots,n\}.
	\end{equation}
	
	\item Scale the elements of $\Ddot{W}$ by its maximum element, and obtain $\Grave{W}$, where 
	\begin{equation}
	\label{scale}
	\Grave{W}_{i,j} = \frac{\Ddot{W}_{i,j}}{\max_{i,j}\{\Ddot{W}_{i,j}\}}.
	\end{equation} 
Note that the methodology discussed in this paper is also applicable to an unweighted graph, where the representation matrix is the binary adjacency matrix, $A$, with elements that are $0$'s or $1$'s. However, performing log transformation followed by scaling would make no difference, hence can be omitted in this case.

\end{enumerate}

\item Obtaining a suitable representation matrix.\\
The mapping of edge weights into a suitable representation matrix is an essential task when using the embedded points to study the structure of the underlying graph \cite{skillicorn2007understanding}. In sparse and heterogeneous graphs possessing power law degree distributions, the embeddings from the weighted adjacency matrix will only focus on vertices with the highest degrees, resulting in an inaccurate representation of the underlying connectivity structure \cite{mihail2002eigenvalue}. To account for sparsity and degree heterogeneity, we construct the \textit{regularized degree normalized weighted adjacency matrix}, $M$, as the representation matrix. Let the regularizer, $\tau$, be 
\begin{equation}
\label{regularize}
\tau = \frac{1}{4n^2}\sum_{i,j}\Grave{W}_{i,j}.
\end{equation}
Then $M$ is given by
\begin{equation}\label{eq:regDegNormWAdj}
M = D_{\tau}^{-1/2}W_{\tau}D_{\tau}^{-1/2},
\end{equation}
where
\begin{equation}\label{eq:regularize}
W_{\tau} = \Grave{W} + \tau \mathbf{1}\mathbf{1}^T,
\end{equation}
where $\mathbf{1}$ is an $n$-dimensional column vector containing all ones, and $D_{\tau}$ is the degree matrix for $W_{\tau}$. The regularization step (Equation \ref{eq:regularize}) addresses sparseness by adding $\tau$ to each element in $\Grave{W}$, while the degree normalization step (Equation \ref{eq:regDegNormWAdj}) further adjusts for the irregularity in the degree distribution by dividing each element, $[W_{\tau}]_{i,j}$, by $\sqrt{[d_{\tau}]_{i}[d_{\tau}]_{j}}$. For a detailed theoretical justification on using $M$ as the representation matrix to obtain an embedding, we refer the reader to \cite{amini2013}.

\item Obtaining a low dimensional embedding from the representation matrix, $M$, using spectral decomposition.\\ 

A low dimensional embedding, $X$, from the representation matrix, $M$ can be seen as a solution to the optimization function,
\begin{equation}
\max_X \parallel X^TMX \parallel^2_F,
\end{equation}
subject to $X^TX = I$, where $X \in \mathbb{R}^{n \times d}$ for $d \ll n$ \cite{ng2001spectral}. The embedding, $X$, can be estimated by performing the singular value decomposition (SVD),  $M=U\Sigma V^T$, and extracting $d$ principal singular vectors. In order to determine $d$ we employ the low-rank matrix approximation procedure in \cite{achlioptas2007fast}, which proposes to retain those singular vectors capturing the strongest structure in $M$ based on the $L_2$ norm. The $L_2$ norm of matrix $M$ is defined as,

\begin{equation}\label{l2norm}
\parallel M \parallel_2 = \max_{\parallel \mathbf{{v}} \parallel_F=1} \parallel M\mathbf{{v}} \parallel_F,
\end{equation}

where $\parallel \mathbf \parallel_F$ denotes the Frobenius norm.

We refer the intersted reader to \cite{achlioptas2007fast} for a detailed and theoretical description of the method. In this section, we summarize our implementation of their method in Algorithm \ref{alg:estimate dachlioptas}.

\begin{algorithm}[]
	\caption{Optimal Low-Rank $d$ Approximation}
	\label{alg:estimate dachlioptas}
	\begin{minipage}{12cm}
		\begin{algorithmic}[1]

			\renewcommand{\algorithmicrequire}{\textbf{Input:}}
			\renewcommand{\algorithmicensure}{\textbf{Output:}}
			
			\REQUIRE (i) Symmetric matrix, $M$, with dimensions ${n \times n}$, where $\textrm{rank}(M)=r$, (ii) threshold, $\epsilon$
			\ENSURE $d$
			\STATE Compute SVD, $M = U \Sigma {U}^T$
			\STATE Update $M=M - \sigma_1 \mathbf{u}_1 \mathbf{u}_1^T$
			\STATE Compute SVD, $M = U \Sigma {U}^T$
			\STATE Initialize k=1, $\rho = \inf$
			\WHILE{$\rho>\epsilon$ \textbf{and} $k\leq r$}
			\STATE $\hat{M}_k= \sum_{j=1}^{k}\sigma_j \mathbf{u}_j \mathbf{{u}}_j^T$
			\STATE $R_k=M-\hat{M}_k$
			\STATE Calculate $\tilde{R_k}$ by randomly flipping the signs of elements in $R_k$ such that, $\mathbb{P}\left[[\tilde{R}_k]_{i,j} = [R_k]_{i,j} \right] = \frac{1}{2}$ and $\mathbb{P}\left[[\tilde{R}_k]_{i,j} = -[R_k]_{i,j} \right] = \frac{1}{2}$
			\STATE Update $\rho=\frac{\lvert \parallel R_k \parallel_2 - \parallel \tilde{R_k} \parallel_2 \rvert}{\parallel R_k \parallel_F}$\\
			\STATE Set $k = k + 1$
			\ENDWHILE
			\IF {$k=r$}
			\STATE Set the converged value $d=r$
			\ELSE
			\STATE Set the converged value $d=k-1$
			\ENDIF
		\end{algorithmic}
	\end{minipage}
\end{algorithm}

It is important to note that the regularization step (Equation \ref{eq:regularize}) inserts edges between all disconnected components and creates a connected graph. For such a graph, the first principle singular vector, $\mathbf{u}_1$ (with corresponding singular value $\sigma_1$), of $M$,  is a constant vector and therefore not useful for the embedding \cite{von2007tutorial}. Thus, to obtain the embedding dimension, $d$, we initially remove the first reconstruction in step 2 of Algorithm \ref{alg:estimate dachlioptas}. Hence, the output $d$ returned by the algorithm is the number of principal singular vectors that should be kept starting from the second principal singular vector onwards\footnote{The Frobenius norm of a matrix measures its \textit{average linear trend} \cite{achlioptas2007fast}. Hence, the division by the Frobenius norm of $R_k$ in step 9 of the algorithm provides a standardization to each $\rho_k$ \cite{skillicorn2007understanding}.}. Once $d$ is obtained, the low dimensional embedding, $X \in \mathbb{R}^{n \times d}$, is given by
\begin{equation*}
X = [\mathbf{u}_2,\ldots \mathbf{u}_{d+1}].
\end{equation*}
Each row vector, $\mathbf{x_i} \in \mathbb{R}^d$, is the feature for $v_i$ at a given time instant.  

Furthermore, an important input parameter for Algorithm \ref{alg:estimate dachlioptas}  is the convergence threshold, $\epsilon$. Since there is no definitive method for choosing $\epsilon$ discussed in \cite{achlioptas2007fast}, we conduct extensive experiments and decide $\epsilon=0.005$ (See Appendix). 

\end{enumerate}

By following steps 1, 2, and 3 in Section \ref{Obtaining a Representative Summary at Each Time Instant}, each graph, $G^t$, in the time sequence is represented as a low dimensional embedding, $X^t \in \mathbb{R}^{n \times d^t}$, where $d^t$ is the {embedding dimension} returned by Algorithm \ref{alg:estimate dachlioptas}. After following the three steps discussed in this section, the sequence of graphs, $G^1,\ldots,G^\mathcal{T}$, is reduced to a sequence of low dimensional embeddings, $X^1, \ldots, X^{\mathcal{T}} $.

\subsection{Obtaining the Profile Features at Each Time Instant}
\label{Obtaining the Profile Behaviour at Each Time Instant}

After performing the steps stated in Section \ref{Obtaining a Representative Summary at Each Time Instant}, we have a set of $w$ embeddings from the $w$ recent past time instants. From the \textit{uniqueness} property of SVD \cite{skillicorn2007understanding}, the embedding obtained at each time instant is unique up to Euclidean similarity transformations such as scale, rotation and reflection. Thus, we cannot directly average the embeddings from the recent past time instants to obtain profile features. Generalized orthogonal Procrustes analysis (GPA) can be used to obtain an average from a set of matrices after adjusting for Euclidean similarity transformations. In this section, we show how we employ GPA to obtain an average embedding, $\xbar{X}^{t-1}$, from the set of embeddings, $X^{t-w},\ldots,X^{t-1}$. We call $\xbar{X}^{t-1}$, the \textit{profile embedding} for time instant $t$. Let us first state the GPA procedure.

The \textit{pre-shape}, $\widetilde{X}$, of a matrix, $X \in \mathbb{R}^{n \times d}$, is defined as 
\begin{equation}\label{eq:preShape}
\widetilde{X} = \frac{X_c}{\parallel X_c \parallel_F},
\end{equation}
where
\begin{equation}
X_c = CX,
\end{equation}
and the centering matrix, $C = I - \frac{1}{n}\mathbf{j}\mathbf{j}^T$. Here,  $I$ is an $n \times n$ identity matrix, and $\mathbf{j}$ is an $n - $dimensional vector of ones. Let $X_1,\ldots,X_w$ be $w$ matrices, each of dimension $n \times d$. GPA involves the optimization of the least squares objective function
\begin{equation}\label{eq:GPAObj}
\min_{\Gamma, \mu}\sum_{i=1}^{w}\parallel \widetilde{X}_i\Gamma_i-\mu \parallel_F^2,
\end{equation}
where $\Gamma_i \in \mathbb{R}^{d \times d}$ is the orthogonal rotation/reflection matrix corresponding to $X_i$, and $\widetilde{X}_i$ is the preshape corresponding to $X_i$ as given in Equation \ref{eq:preShape}. In Algorithm \ref{alg:GPA} we summarize our implementation of the iterative algorithm that solves the GPA objective function. 

\begin{algorithm}[]
	\caption{Generalized Procrustes Distance Calculation}
	\label{alg:GPA}
	\begin{algorithmic}[1]
		
		\renewcommand{\algorithmicrequire}{\textbf{Input:}}
		\renewcommand{\algorithmicensure}{\textbf{Output:}}
		
		\REQUIRE $X_1,\ldots,X_w \in \mathbb{R}^{n \times d}$, threshold $\epsilon$
		\ENSURE $\hat{\mu}$, $\widehat{X}_i$ for $i=1,\ldots, w$
		\vspace{0.5cm}
		\STATE Initialize $\mu_0=X_1$, $\mathcal{D}=\inf$
		\WHILE{$\mathcal{D}>\epsilon$}
		
		\FOR{$i\in \{1,\ldots,w\}$}
		\STATE Calculate $\widetilde{X}_i=\frac{CX_i}{\parallel CX_i \parallel_F}$, where $C = I - \frac{1}{n}\mathbf{j}\mathbf{j}^T$
		\STATE Calculate SVD, $\mu_0^T\widetilde{X}_i=U\Sigma {V}^T$ 
		\STATE Calculate $\hat{\Gamma}_i={V}U^T$ 
		\STATE Obtain $\hat{X}_i=\widetilde{X}_i\hat{\Gamma}_i$
		\ENDFOR
		\STATE Calculate mean embedding, $\hat{\mu}=\frac{1}{w}\sum_{i=1}^{w}\widehat{X}_i$, using the aligned embeddings
		\STATE Update ${\mathcal{D}}=\parallel \mu_0-\hat{\mu}\parallel_F^2$
		\STATE Update: $\mu_0 = \hat{\mu}$
		\ENDWHILE 
	\end{algorithmic}
\end{algorithm}

There is one limitation in applying GPA to the embeddings obtained at different time instants. GPA assumes that all matrices, $X^{t-w},\ldots,X^{t-1}$, are of the same dimension, but the embeddings resulting from our methods discussed in Section \ref{Obtaining a Representative Summary at Each Time Instant}, can be of different dimensions. We find two possible solutions to address this problem. Let $d_{\max}=\max \{d^{t-w},\ldots, d^{t-1}\}$. 
\begin{enumerate}[topsep=0pt]
	\item For any $X^t$ with $d^t<d_{\max}$, append columns of zeros to $X^t$ to make it of size $n \times d_{\max}$.
	\item For any $X^t$ with $d^t>d_{\max}$, truncate the additional columns of $X^t$ to make it of size $n \times d_{\max}$.
\end{enumerate} 
Truncating extra dimensions causes us to drop singular vectors that may describe important structure of the graph. Appending columns of zeros does not cause loss of information, and is thus preferred.  Thus, whenever the dimensions of the embeddings to be compared are different from each other, we append the low dimensional embedding with columns of zeros before fitting the generalized Procrustes model. 

Therefore, the profile embedding, $\xbar{X}^{t-1}$, is calculated as follows:
\begin{enumerate}
	\item Let $d_{\max}= \textrm{max} \{d^{t-w}\ldots,d^{t-1}\}$. Append $d_{\max}-d^t$ columns of zeros to each $X^t$ and obtain $X^t_{\text{padded}}$. 
	\item Perform the generalized Procrustes analysis procedure and estimate the mean embedding, $\xbar{X}^{t-1}=\hat{\mu}$. To do this, we input $X^{t-w}_{\text{padded}},\ldots,X^{t-1}_{\text{padded}}$ into Algorithm \ref{alg:GPA}, and estimate the mean embedding, $\hat{\mu}$.
\end{enumerate}
At each time instant $t$, the $n$ rows of $\xbar{X}^{t-1}$ give the profile features for the $n$ vertices in the graph. 

\subsection{Change Score Calculation}
\label{Change Score Calculation}

After applying the methods discussed in Sections \ref{Obtaining a Representative Summary at Each Time Instant} and \ref{Obtaining the Profile Behaviour at Each Time Instant}, at each time instant $t$, we end up with the profile embedding, $\xbar{X}^{t-1}\in \mathbb{R}^{n \times \bar{d}^{t-1}}$, and current embedding, $X^t \in \mathbb{R}^{n \times d^{t}}$. Vertex change scores are calculated by computing the dissimilarity between $X^t$ and $\xbar{X}^{t-1}$. Procrustes analysis can be used to compare two matrices after adjusting for Euclidean similarity transformations. From Section \ref{Obtaining the Profile Behaviour at Each Time Instant}, when $\bar{d}^{t-1} \neq d^t$, we append columns of zeros to the lower dimensional embedding. Thus, the change score, $z^t_i$, for vertex $i$ at time instant $t$ is calculated as follows:
\begin{enumerate}
	\item Let $d_{\max}= \textrm{max} \{d^t,\bar{d}^{t-1}\}$. Append $d_{\max}-d^t$ and $d_{\max}-\bar{d}^{t-1}$ columns of zeros to $X^t$ and $\xbar{X}^{t-1}$, respectively and obtain $X^t_{\text{padded}} \in \mathbb{R}^{n \times d_{\textrm{max}}}$ and $\xbar{X}^{t-1}_{\text{padded}} \in \mathbb{R}^{n \times d_{\textrm{max}}}$. 
	
	\item  Perform GPA using Algorithm \ref{alg:GPA} and obtain the transformed embeddings, $\widehat{X}^t$ and $\widehat{\xbar{X}}^{t-1}$, and the average of the transformed embeddings, $\hat{\mu}^t$. 
\item For each vertex $i$, calculate the change score
\begin{equation}
{z}_i^t = \frac{\parallel \widehat{X}_{i,.}^t - \widehat{\xbar{X}}_{i,.}^{t-1} \parallel^2_F}{\parallel \hat{\mu}^t \parallel_F}.
\end{equation}

\end{enumerate}

\subsection{Proposed Algorithm - CDP (Change Detection using Procrustes Method)}
\label{CDP Algorithm}

We have now constructed the three main steps of our change detection procedure. These include, at each time instant, extracting features for vertices through graph embedding (Section \ref{Obtaining a Representative Summary at Each Time Instant}), calculating profile features for vertices by applying GPA on the recent past embeddings (Section \ref{Obtaining the Profile Behaviour at Each Time Instant}), and finally calculating change scores for vertices through generalized Procrustes distance calculation between current and profile embeddings (Section \ref{Change Score Calculation}). The steps are listed in Algorithm \ref{alg:CDP}.

\begin{algorithm}[]
	\caption{Change Detection using Procrustes Analysis - CDP}
	\label{alg:CDP}
	\begin{algorithmic}[1]

		\renewcommand{\algorithmicrequire}{\textbf{Input:}}
		\renewcommand{\algorithmicensure}{\textbf{Output:}}

		\REQUIRE (i) Time sequence of symmetric, weighted adjacency matrices, $W^1,W^2,\ldots W^\mathcal{T}$, where each $W^t$ has dimension ${n \times n}$ (ii) window size, $w$
		\ENSURE Time sequence of vertex change scores, $\mathbf{z}^1, \mathbf{z}^2, \ldots \mathbf{z}^\mathcal{T}$. Each $\mathbf{z}^t$ is a vector of dimension $n$  
		%\vspace{0.5cm}
		%\STATE Obtain lower dimensional embeddings: 
		%\vspace{0.2cm}
		\FOR{$t=1$ \TO $\mathcal{T}$ }
		\STATE Update: $W^t=\log_{10}(W^t+\mathbf{1}_{n\times 1}\mathbf{1}_{n\times 1}^T)$
		\STATE Update: $W^t=\frac{W^t}{\max_{i,j}\{W_{i,j}^t\}}$
		\STATE Calculate $\tau^t = \frac{ 1}{4n^2}\sum_{i,j}W_{i,j}^t$\\
		Update: $W_{\tau}^t = W^t + \tau^t \mathbf{1}_{n\times 1}\mathbf{1}_{n\times 1}^T$,\\
		Calculate  $D_{\tau}^t \in \mathbb{R}^{n \times n}$, where {$[D_{\tau}^t]_{i,i} = \sum_{j=1}^n [W_{\tau}^t]_{i,j}$}\\
		Calculate $M^t = (D_{\tau}^t)^{-1/2}W_{\tau}^t (D_{\tau}^t)^{-1/2}$
		
		%\STATE Calculate {$M^t = (D_{\tau}^t)^{-1/2}W_{\tau}^t (D_{\tau}^t)^{-1/2}$}, where\\
		%$W_{\tau}^t = W^t + \tau^t \mathbf{1}_{n\times 1}\mathbf{1}_{n\times 1}^T$,\\
		%$\tau^t = \frac{ 1}{4n^2}\sum_{i,j}W_{i,j}^t$, and \\
		% $D_{\tau}^t \in \mathbb{R}^{n \times n}$ be the diagonal matrix with elements, {$[D_{\tau}^t]_{i,i} = \sum_{j=1}^n [W_{\tau}^t]_{i,j}$}
		%\ENDFOR    
		%\vspace{0.5cm}    
		%\FOR{$t=1$ \TO $T$ } 
		\STATE{{Input $M^t$ to Algorithm \ref{alg:estimate dachlioptas}}} and estimate $d^t$
		\STATE{{Perform SVD}: $M^t = U\Sigma U^T$}
		\STATE{{Obtain the low dimensional embedding} ${X_t}\in \mathbb{R}^{n \times d^t}$}, where\\
		${X^t} = [\mathbf{u}_2,\ldots \mathbf{u}_{d^t+1}]$
		\ENDFOR
		
		%\vspace{0.5cm}    
		%\STATE 2. Obtaining the profile behaviour and change score calculation:
		\FOR{$t=w+1$ \TO $\mathcal{T}$ }
		\STATE{Let $d_{\max_1}=\max \{d^{t-w},\ldots,d^{t-1}\}$}
		\FOR{$t'\in \{t-w,\ldots,t-1\}$}
		\IF{$d^{t'} < d_{\max_1}$}
		\STATE append $d_{\max_1}-d^{t'}$ columns of zeros to $X^{t'}$  
		\ENDIF
		\ENDFOR
		\STATE{Input $X^{t-w},\ldots,X^{t-1}$ into Algorithm \ref{alg:GPA}, and estimate the profile embedding $\xbar{X}^{t-1}$}
		\STATE {Let $d_{\max_2}=\max \{d_{\max_1},d^{t}\}$}
		\IF{$d_{\max_1} < d_{\max_2}$}
		\STATE append $d_{\max_2}-d_{\max_1}$ columns of zeros to $\xbar{X}^{t-1}$   
		\ENDIF
		\IF{$d^{t} < d_{\max_2}$}
		\STATE append $d_{\max_2}-d^t$ columns of zeros to $X^t$ 
		\ENDIF
		\STATE{Align $X^t$ and $\xbar{X}^{t-1}$ with each other using Algorithm \ref{alg:GPA}, and obtain the adjusted embeddings, $\widehat{X}^t$ and $\widehat{\xbar{X}}^{t-1}$}, and the mean $\hat{\mu}^t$
		\STATE Calculate vertex change scores,
		${z}_i^t = \frac{\parallel \widehat{X}_{i,.}^t - \widehat{\bar{X}}_{i,.}^{t-1} \parallel^2_F}{\parallel \hat{\mu}^t\parallel_F}$
		\ENDFOR
	\end{algorithmic}
\end{algorithm}

After describing our algorithm, we evaluate its performance by conducting experiments on simulated dynamic networks and a real-world dataset.

\section{Simulation Experiments}
\label{Simulation Experiments}

Simulated networks enable us to comprehend not only how and when a specific technique is doing well, but also when a technique is not doing well \cite{yu2019monitoring}. We conduct such an investigation by generating different synthetic datasets that mimic several real-world change scenarios. Within each scenario, a subset of vertices under-go change from recent past behaviour. 

\subsection{Overall Setting}
\label{Overall Setting}

For each change scenario we generate a time sequence of symmetric weighted adjacency matrices, $W^1,W^2, \ldots, W^\mathcal{T}$, to represent a time sequence of weighted graphs. Similar to \cite{wang2017fast}, we assume that each network is generated from a certain recognized underlying model that determines the process of generation. We assume that the edges of the graphs have distribution $F_0$ and when a change occurs the distribution becomes $F_1$. We consider two types of changes.

\begin{enumerate}
	\item Change occurs at a given time instant: \textit{change-point}.
	A change point is injected to the time sequence of graphs by defining the edge distribution as
	\begin{equation}\label{eq:point}
	W^t \sim
	\begin{cases}
	F_1 & \text{if } t = t^*,\\
	F_0 & \text{otherwise },
	\end{cases}
	\end{equation}
	for $w < t^* \leq \mathcal{T}$.
	
	\item Change occurs at a time instant, and persists for some time period: \textit{change-interval}. A change-interval is generated by defining the edge distribution as
	\begin{equation}\label{eq:step}
	W^t \sim
	\begin{cases}
	F_1 & \text{if } t^*_1 \leq t \leq t^*_2,\\
	F_0 & \text{otherwise },
	\end{cases}
	\end{equation}
	for $ w < t^*_1 < t^*_2 \leq \mathcal{T}$.
\end{enumerate}
In Section \ref{Generative Model for Network Generation}, we discuss the model that is used to generate graphs for our experiments.

\subsection{Random Graph Model Used for Synthetic Network Generation}
\label{Generative Model for Network Generation}

The degree corrected stochastic block model (DCSBM) \cite{karrer2011stochastic} is a commonly used model because it can closely mimic the community structure of real-world networks. In our simulation experiments, we employ the DCSBM to define the probability distribution of the edges of a graph. By adjusting the model parameters, we obtain a wide variety of edge distributions. 

Let $c_i \in \{1,2,\ldots, k\}$ denote the block membership of vertex $i$. Then the vector, $\mathbf{c} \in \{1,2,\ldots, k\}^{n}$, of dimension $n$ denotes the block memberships of the $n$ vertices in the graph. In terms of the weighted adjacency matrix, $W$, its distribution under the DCSBM is given by
\begin{equation}\label{eq:DCSBMch3}
\mathbb{P}[W|\pmb{\theta},\psi,\mathbf{c}] = \prod_{i\neq j}\frac{(\theta_i\theta_j\psi_{c_i,c_j})^{W_{i,j}}}{W_{i,j}!}\exp(-\theta_i\theta_j\psi_{c_i,c_j}),
\end{equation}
where $\psi_{c_i,c_j}$ is the expected number of edges between a vertex in block $c_i$ and a vertex in block $c_j$, and $\pmb{\theta}$ is an $n$-dimensional vector of degree parameters. Each element, $W_{i,j}$, is a Poisson random variable with mean $\theta_i\theta_j\psi_{c_i,c_j}$. 
In order to mimic the degree distribution of real-world graphs, the vector, $\pmb{\theta}$, is generated from a power-law distribution \cite{clauset2009power} defined as
\begin{equation*}
\mathbb{P}(\pmb{\theta}|\theta_{\textrm{min}},\beta) = \prod_{i=1}^{n}\frac{\beta - 1}{\theta_{\textrm{min}}}\left(\frac{\theta_i}{\theta_{\textrm{min}}}\right) ^{-\beta},
\end{equation*}
where $\theta_{\textrm{min}}$ is the lower bound of the support of $\theta_i$, $\beta$ is the shape parameter. The ${\theta}_i$'s are normalized to sum to one for vertices in the same block, i.e., $\sum_{i} \theta_i \delta_{c_i,r}=1$ (where $\delta_{c_i,r}=1$ if vertex $i$ belongs to block $r$). 

To specify what $\psi$ is, let $B\in [0,1]^{k \times k}$ be the block probability matrix where each element, $B_{c_ic_j}$, denotes the probability of an edge between vertices in blocks $c_i$ and $c_j$.  Using $\mathbf{c}$, we can obtain $\mathbf{g} \in \mathbb{R}^{k \times 1}$, where each element, $g_r=\sum_{i=1}^{n}\delta_{c_i,r}$, denotes the number of vertices in block $r$. Using $B$ and $\mathbf{g}$ we can calculate the expected number of edges, $\psi_{r,s}$, between a vertex in block $r$ and a vertex in block $s$ giving
\begin{equation*}
\psi_{r,s}=B_{r,s}{g_r}{g_s}.                    \end{equation*}
We select $B$ to have the form 
\begin{equation}\label{eq:lambda}
B= \lambda B^{\text{planted}} + (1-\lambda)B^{\text{random}},
\end{equation}
where $\lambda \in [0,1]$. For example, for a graph with three blocks, $B^{\textrm{planted}}$ can take the form,
\begin{equation}\label{eq:wplanted}
B^{\text{planted}} = \begin{bmatrix}
\alpha  & 0 & 0 \\
0 & \beta & 0 \\
0 & 0  & \gamma
\end{bmatrix},
\end{equation}
where $\alpha, \beta, \gamma \in [0,1]$ give the intra-block probabilities. $B^{\textrm{random}}$ is given by
\begin{equation}\label{eq:wrandom}
B^{\text{random}}= \nu\mathbf{1}_k\mathbf{1}_k^T,
\end{equation}
where $\mathbf{1}_k$ is the $k \times 1$ vector of ones, and $\nu \in [0,1]$. $\nu$ can be regarded primarily as an inter-block probability. Thus, by varying $\lambda$, we can vary the level of noise in the generated graphs, which makes it more difficult to identify the blocks. 

The Equations (\ref{eq:DCSBMch3} to \ref{eq:wrandom}) for the distributions of probability make the DCSBM a strong, flexible and popular tool for analyzing complex networks \cite{de2016detection,yu2019monitoring}. The distributions, $F_0$ and $F_1$, for the edges are obtained using different sets of parameter values. Each set of parameter values is chosen to mimic real-world change scenarios involving vertices. In Table \ref{tab:modelparas}, we summarize the parameter settings of different DCSBM models used to generate graphs in our experiments. 

\subsection{Change Scenarios}
\label{Change Scenarios}

A detailed review on numerous change scenarios studied in previous research is given in \cite{hewapathirana2019change}. Based on these ideas, we come up with the following change scenarios to evaluate our change detection method.

\begin{enumerate}
	\item Change in block membership - \textit{group-change}.\\
	A set of vertices in a block change their block (group) membership.
	\item Change in block Structure,
	\begin{enumerate}
		\item \textit{split} - a block in the graph splits into two blocks,
		\item \textit{merge} - the reverse of split:  two blocks join together and form one block,
		\item \textit{form} - a high increase in connections in a block that was previously sparse,
		\item \textit{fragment} - the reverse of form: a dense block becomes sparse.
	\end{enumerate}
	\item Change in degree, 
	\begin{enumerate}
		\item Heterogeneous degrees to homogeneous degrees - \textit{hetero-to-homo}.\\
		The degree parameters of a block of vertices in the graph change from heterogeneous to homogeneous. 
		
		\item Homogeneous degrees to heterogeneous degrees - \textit{homo-to-hetero}.\\
		The reverse of hetero-to-homo: the degree parameters of a block of vertices change from homogeneous to heterogeneous.
	\end{enumerate}
	\item Change in connectivity patterns:
	\begin{enumerate}
		\item Clear block structure to complex structure - \textit{simple-to-complex}.\\
		Two blocks add inter-block edges, disrupting the clear block structure in the graph.
		\item Complex block structure to clear block structure - \textit{complex-to-simple}.\\
		The reverse of simple-to-complex: most inter-block edges between two blocks vanish, resulting in a graph with a clear block structure.
	\end{enumerate}
\end{enumerate}

In Table \ref{tab:scenarios}, we give a detailed description of how we mimic these change scenarios through transitions of the underlying generative models. Each scenario corresponds to changes in the connectivity patterns of a subset of vertices in the graph. For each scenario, we visualize an example of $W$'s generated from the models corresponding to $F_0$ and $F_1$. 

For each change scenario, we generate a sequence of $30$ graphs, that is, we set $\mathcal{T}=30$. The parameters for the two types of changes defined in Section \ref{Overall Setting} are as follows.
\begin{enumerate}
	\item change-point (Equation \ref{eq:point}): $t^*=21$,
	\item change-interval (Equation \ref{eq:step}): $t^*_1 = 21, t^*_2=30$. 
\end{enumerate}
We use windows of sizes $1,5$, and $10$, and calculate change scores for all vertices. We repeat this $100$ times, and calculate our performance measures (Section \ref{Performance Measure}).

\begin{table}[]
	%  \centering
	\caption{Parameter settings of different models with fixed parameters, $n=900,\lambda=0.8$, $B^{\textrm{planted}}$ with $\alpha=0.01,\beta=0.02,\gamma=0.03$, and $B^{\textrm{random}}$ with $\nu=0.0025$.}\label{tab:modelparas}
	\begin{minipage}{12cm}
		\begin{tabular}{cccccccc}
			\toprule
			Model\footnote{Different values of $\lambda$ were tested ($\lambda=0,0.1,0.2,\ldots,1$), but the same $\lambda$ value is used for the pair of models involved in a given change scenario.}
			& $B^{\textrm{planted}}$ & Distribution of $\theta$ & $\mathbf{g}$ & $k$ \\
			\midrule
			$\mathcal{M}1$ & $\begin{bmatrix}
			\alpha  & 0 & 0 \\
			0 & \beta & 0 \\
			0 & 0  & \gamma
			\end{bmatrix}$ & $\pmb{\theta} \sim \mathbb{P}(\pmb{\theta}|1,2.5)$ & $[300,300,300]$ & 3\\
			& & & & \\   
			$\mathcal{M}2$ & $\begin{bmatrix}
			\alpha  & 0 & 0 & 0 \\
			0 & \alpha & 0 & 0\\
			0 & 0 & \beta & 0 \\
			0 & 0 & 0  & \gamma
			\end{bmatrix}$ & $\pmb{\theta} \sim \mathbb{P}(\pmb{\theta}|1,2.5)$ & $[150,150,300,300]$ & 4\\
			& & & & \\  
			$\mathcal{M}3$ & $\begin{bmatrix}
			\alpha  & 0 & 0 \\
			0 & \beta & 0 \\
			0 & 0  & 0.1(\gamma)
			\end{bmatrix}$ & $\pmb{\theta} \sim \mathbb{P}(\pmb{\theta}|1,2.5)$ & $[300,300,300]$ & 3\\
			& & & & \\
			$\mathcal{M}4$ & $\begin{bmatrix}
			\alpha  & 0 & 0 \\
			0 & \beta & 0 \\
			0 & 0  & \gamma
			\end{bmatrix}$ & $\pmb{\theta} \sim \mathbb{P}(\pmb{\theta}|1,2.5)$ & $[150,450,300]$ & 3\\
			& & & & \\
			$\mathcal{M}5$ & $\begin{bmatrix}
			\alpha  & 0 & 0 \\
			0 & \beta & 0 \\
			0 & 0  & \gamma
			\end{bmatrix}$ & $\pmb{\theta}_{V_c} = \mathbf{c}$ \footnote{$\pmb{\theta}_{V_c}$ is a vector of degree parameters of the set of vertices, $V_c = \{v_1,v_2,\ldots ,v_{300}\}$, and $\mathbf{c}$ denotes a positive vector of constants.} & $[300,300,300]$ & 3\\
			&  & $\pmb{\theta}_{\bar{V_c}}\sim \mathbb{P}(\pmb{\theta}|1,2.5)$ \footnote{$\pmb{\theta}_{\bar{V_c}}$ is a vector of degree parameters of the set of vertices, $\bar{V_c}= \{v_{301},v_{302},\ldots ,v_{900}\}$.} & & \\
			& & & & \\
			$\mathcal{M}6$ & $\begin{bmatrix}
			0.5\alpha & 0.5\alpha & 0 \\
			0.5\alpha & \beta-0.5\alpha & 0 \\
			0 & 0 & \gamma 
			\end{bmatrix}$ & $\pmb{\theta} \sim \mathbb{P}(\pmb{\theta}|1,2.5)$ & $[300,300,300]$ & 3\\
			& & & & \\
			\bottomrule
		\end{tabular}
	\end{minipage}
\end{table}

\begin{table}[]
	\centering
	\caption{\textbf{Illustration of change scenarios}. Each scenario corresponds to a change in the connectivity patterns of a subset of vertices in the DCSBM graph and is visualized using the pixel-plots of the adjacency matrices generated. }
	\label{tab:scenarios}
	\begin{tabular}{|c|c|c|c|c|}
		\toprule
		No. & Change Scenario & $F_0$ & $F_1$ & Changed Vertices\\
		\midrule
		1 & group-change & $\mathcal{M}1$ & $\mathcal{M}4$ & $\{1\ldots600 \}$ \\
		&  & \includegraphics[trim = 0mm 0mm 0mm 0mm, scale=0.11]{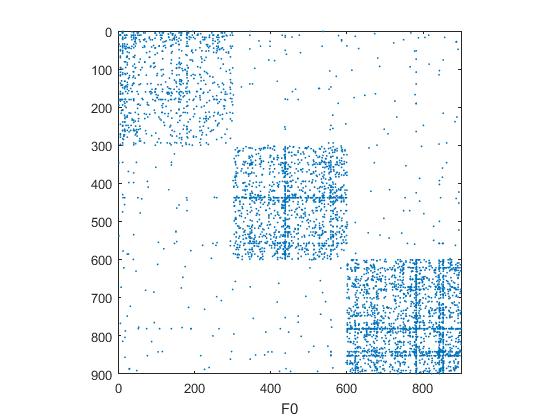} & \includegraphics[trim = 0mm 0mm 0mm 0mm, scale=0.11]{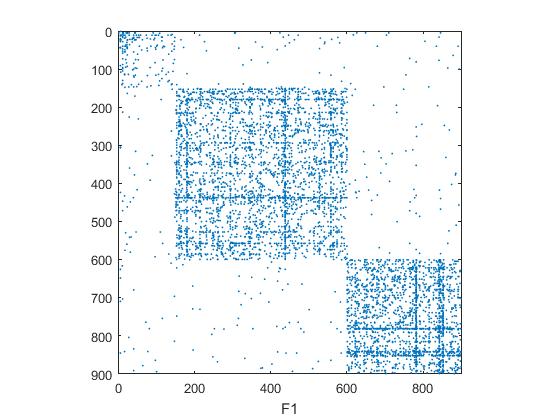} & \\ \midrule
		2 & split & $\mathcal{M}1$ & $\mathcal{M}2$ & $\{1 \ldots300 \}$ \\
		&  & \includegraphics[trim = 0mm 0mm 0mm 0mm, scale=0.11]{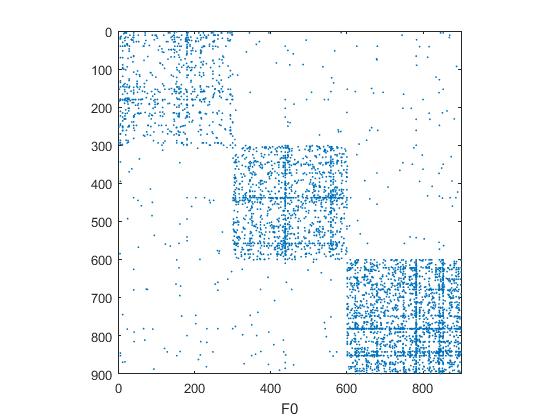}  & \includegraphics[trim = 0mm 0mm 0mm 0mm, scale=0.11]{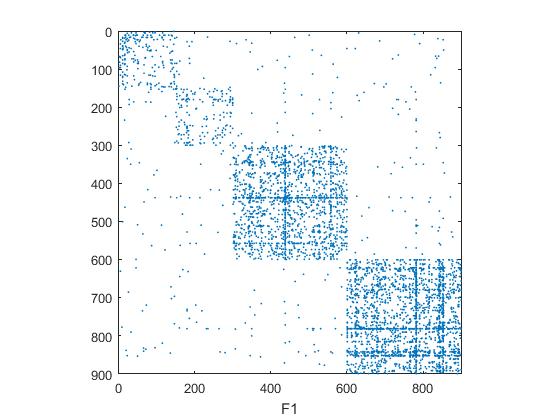} &  \\ \midrule
		3 & merge & $\mathcal{M}2$ & $\mathcal{M}1$ & $\{1 \ldots 300 \}$ \\ 
		&  & \includegraphics[trim = 0mm 0mm 0mm 0mm, scale=0.11]{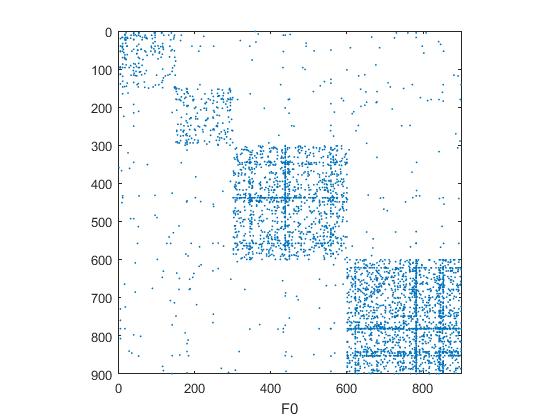} & \includegraphics[trim = 0mm 0mm 0mm 0mm, scale=0.11]{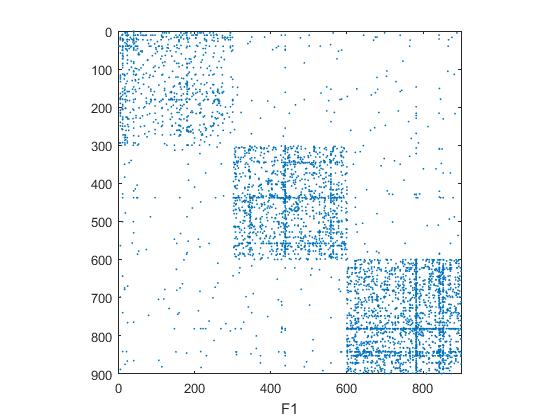} &  \\ \midrule
		4 & form & $\mathcal{M}3$ & $\mathcal{M}1$ & $\{601 \ldots 900 \}$ \\
		&  & \includegraphics[trim = 0mm 0mm 0mm 0mm, scale=0.11]{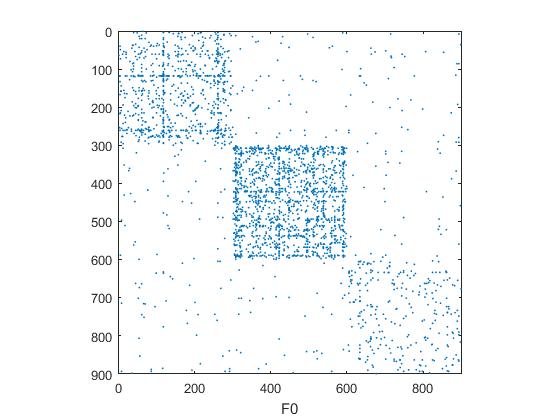} & \includegraphics[trim = 0mm 0mm 0mm 0mm, scale=0.11]{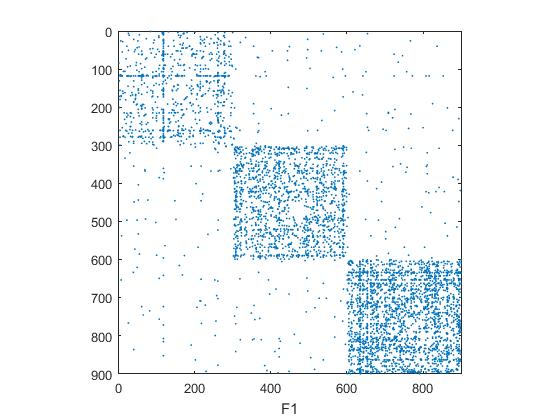} &  \\ \midrule
		5 & fragment & $\mathcal{M}1$ & $\mathcal{M}3$ & $\{601 \ldots 900 \}$  \\
		&  & \includegraphics[trim = 0mm 0mm 0mm 0mm, scale=0.11]{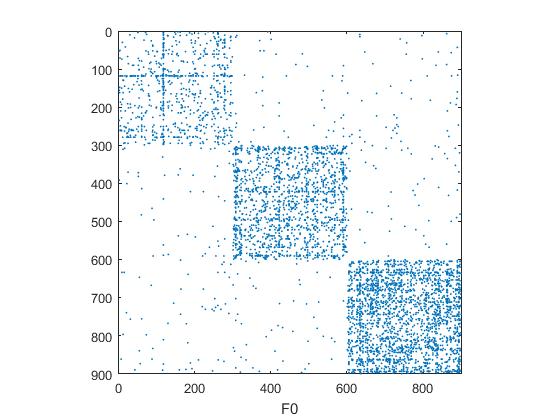}  & \includegraphics[trim = 0mm 0mm 0mm 0mm, scale=0.11]{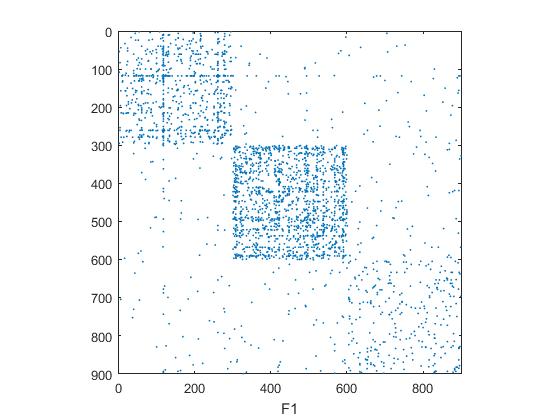} &  \\ \midrule
		6 & hetero-to-homo & $\mathcal{M}1$ & $\mathcal{M}5$ & $\{1 \ldots 300 \}$ \\
		&  & \includegraphics[trim = 0mm 0mm 0mm 0mm, scale=0.11]{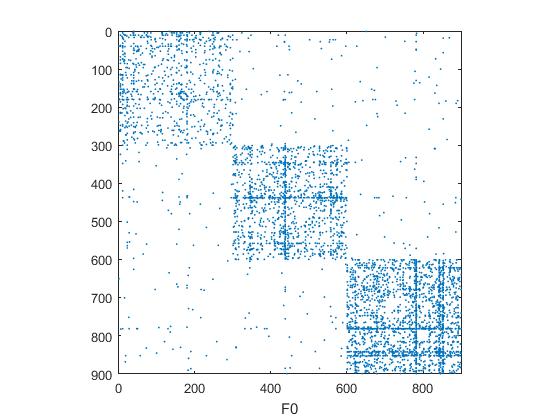} & \includegraphics[trim = 0mm 0mm 0mm 0mm, scale=0.11]{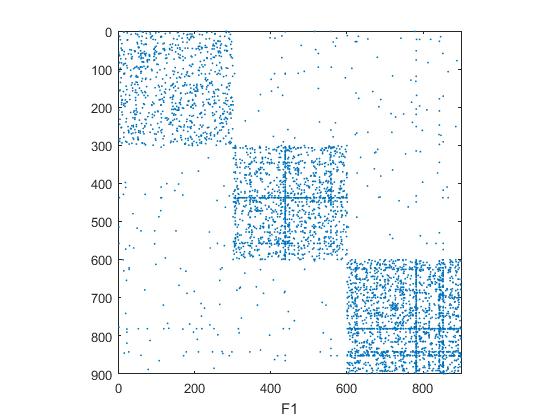} &  \\ \midrule
		7 & homo-to-hetero & $\mathcal{M}5$ & $\mathcal{M}1$ & $\{1 \ldots 300 \}$\\
		&  & \includegraphics[trim = 0mm 0mm 0mm 0mm, scale=0.11]{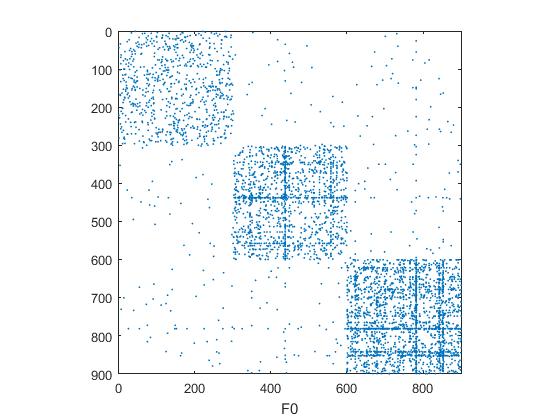} & \includegraphics[trim = 0mm 0mm 0mm 0mm, scale=0.11]{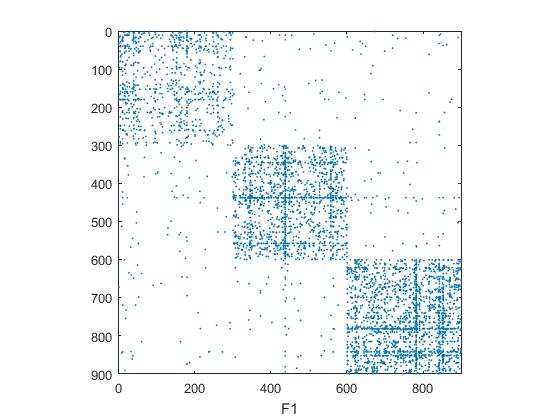}  &  \\ \midrule
		8 & simple-to-complex & $\mathcal{M}1$ & $\mathcal{M}6$ & $\{1 \ldots 600\}$ \\
		&  & \includegraphics[trim = 0mm 0mm 0mm 0mm, scale=0.11]{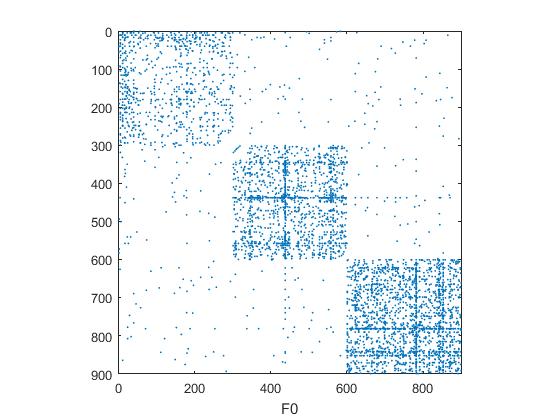} & \includegraphics[trim = 0mm 0mm 0mm 0mm, scale=0.11]{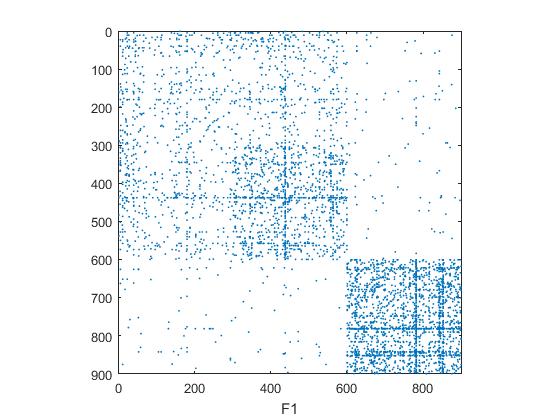} &  \\ \midrule
		9 & complex-to-simple & $\mathcal{M}6$ & $\mathcal{M}1$ & $\{1 \ldots 600\}$  \\
		&  & \includegraphics[trim = 0mm 0mm 0mm 0mm, scale=0.11]{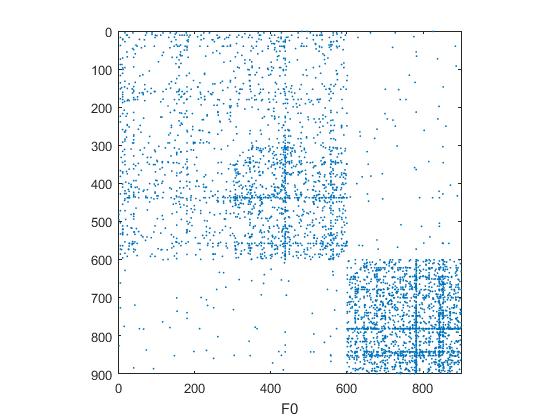} & \includegraphics[trim = 0mm 0mm 0mm 0mm, scale=0.11]{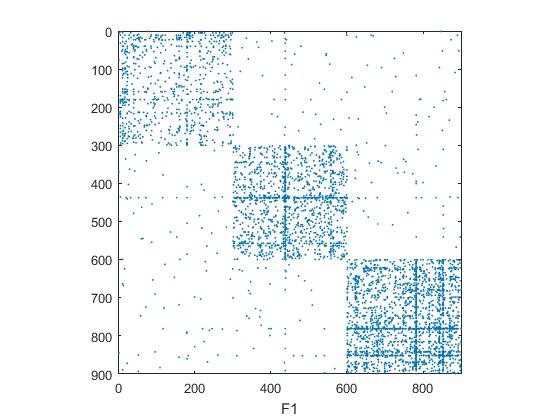}  &  \\ 
		
		\bottomrule
	\end{tabular}
\end{table}

\subsection{Performance Measure}
\label{Performance Measure}

Since our goal is to detect vertices that have changed their behaviour with respect to the recent past, we measure the performance of CDP with respect to the ability of the change scores produced to discriminate between changed and unchanged vertices. Each change scenario discussed in Section \ref{Change Scenarios} involves a set of vertices, $V_c$, changing their behaviour. Let $|V_c|=n_c$. If our method performs well, the change scores for vertices in $V_c$ should be higher than the change scores for the rest of the vertices in $V_{\bar{c}}$, especially at the time instant corresponding to a change. Note that $|V_{\bar{c}}|=n_{\bar{c}}$, $V_c \cup V_{\bar{c}} = V$, and $n_c+n_{\bar{c}}=n$.

Let us consider a time sequence of vertex change scores, $\mathbf{z}^1, \mathbf{z}^2, \ldots \mathbf{z}^{30}$, where each $\mathbf{z}^t$ is a vector of length $n$ obtained from a single simulation run of a change scenario. Let $\mathbf{\tilde{z}}^t$ be the $n_c \times 1$ vector of change scores obtained for $V_c$, and let $\mathbf{{\bar{z}}}^t$ be the $n_{\bar{c}} \times 1$ vector of change scores obtained for $V_{\bar{c}}$. We use a sampling procedure to estimate
\begin{equation*}
\phi^t=\mathbb{P}[{\tilde{z}}_i^t>{{\bar{z}}}_j^t],
\end{equation*}
which is the probability that vertex, $i$, in $V_c$ has a higher change score than vertex, $j$, in $V_{\bar{c}}$. We separately sample (with replacement) a vector of $N$ elements, $\hat{\tilde{\mathbf{z}}}^t$, from $\tilde{\mathbf{z}}^t$ and a vector of $N$ elements, $\hat{\bar{\mathbf{z}}}^t$, from $\bar{\mathbf{z}}^t$; then $\phi^t$ is calculated by counting the proportion of entries in $\hat{\tilde{\mathbf{z}}}^t$ that are larger than the corresponding entries in $\hat{\bar{\mathbf{z}}}^t$ as
\begin{equation*}
\phi^t\approx \frac{1}{N}\sum_{i=1}^{N}\delta_{\hat{\tilde{{z}}}_i^t>\hat{\bar{{z}}}_i^t},
\end{equation*}
where $\delta_{\hat{\tilde{{z}}}_i^t>\hat{\bar{{z}}}_i^t}$ is one if $\hat{\tilde{{z}}}_i^t>\hat{\bar{{z}}}_i^t$ and zero otherwise. In our experiments we use $N=100000$.

A proportion greater than $0.5$ indicates a higher chance of a change score for a vertex in $V_c$ being greater than a change score for a vertex in  $V_{\bar{c}}$. By repeating this for all $100$ simulation runs, we obtain a $100 \times 1$ vector of probabilities, ${\pmb{\phi}}^t$. If all elements of ${\pmb{\phi}}^t$ are greater than $0.5$ and closer to one at a changed time instant, good change detection performance is indicated. Instead of directly using ${{\phi}}_i^t$, we use the log odds
\begin{equation}\label{eq:odds}
{\eta}_i^t =   \log \left (\frac{{{\phi}}_i^t}{1-{{\phi}}_i^t}\right ),
\end{equation}
which measures the odds that a vertex in $V_c$ has higher change scores than a vertex in $V_{\bar{c}}$. When a change occurs, we expect the values of $\pmb{\eta}^t$ to lie above zero and be strongly positive. 
After calculating ${\eta}_i^t$, we further calculate the log odds ratio between time instants $t$ and $t-1$ which gives
\begin{equation}\label{eq:perf}
\bar{{\eta}}_i^t = \log\left ( \frac{\phi_i^t/(1-\phi_i^t)}{\phi_i^{t-1}/(1-\phi_i^{t-1})}\right ).
\end{equation}
In our experiments we calculate both $\pmb{\eta}^t$ and $\bar{\pmb{\eta}}^t$ to measure detection performance.

\subsection{Comparison Methods}
\label{Comparison Methods}

We compare our CDP algorithm with two baseline methods.
\begin{enumerate}
	\item \textbf{ACT}\\
	This is the \textit{activity} (ACT) vector-based change detection algorithm developed by \cite{ide2004eigenspace}. They employ a spectral embedding procedure, and represent a time sequence of graphs as a time sequence of activity vectors, $\mathbf{u}^t$, for $t\in \{1,2,\ldots,\mathcal{T}\}$. A profile vector, $\mathbf{r}^{t-1}$, is calculated from recent past $w$ activity vectors. The change score, ${z}_i^t$, can be calculated as
	\begin{equation}\label{eq:ide_nodescore}
	{z}_i^t = |{r}_i^{t-1} - {u}_i^t|,
	\end{equation} 
	where $|.|$ denotes absolute value. The elements of the activity vector, $\mathbf{u}^t$, denote the eigenvector centrality scores of the vertices in the graph. \cite{ide2004eigenspace} developed ACT to perform change detection in a time sequence of dense graphs. However in the majority of the applications we encounter, the graph obtained at each time instant is sparse and heterogeneous. As discussed in Section \ref{Brief Overview}, such a graph consists of vertices with very high degree (\textit{hubs}) as well as very low degree (sometimes zero; resulting in disconnected vertices in the graph). According to \cite{martin2014localization}, eigenvector centrality is a poor performance measure of centrality of vertices in sparse graphs. They show that the centrality scores are concentrated only on hubs and fail to capture the centrality of lower degree vertices. While this situation might be useful for some applications, for our current requirement of detecting changes in the behaviour of all vertices in the graph, it is inadmissible. Thus, we find \cite{ide2004eigenspace}'s approach cannot be generalized to most real-world graphs.
	
	\item \textbf{ACTM}\\
	We make a slight improvement to the profile vector calculation step in \cite{ide2004eigenspace} and call this method the \textit{modified activity} (ACTM) vector-based algorithm. Recall that \cite{ide2004eigenspace} represent the recent past behaviour using the profile vector, $\mathbf{r}^{t-1}$. However, $\mathbf{r}^{t-1}$ is only the first vector, $\mathbf{r}_1$, from the $w$ singular vectors,   $[\mathbf{r}_1,\mathbf{r}_2,\ldots,\mathbf{r}_w]$, resulting from the SVD of the $n \times w$ matrix of activity vectors, $[\mathbf{u}^{t-1},\ldots, \mathbf{u}^{t-w}]$, representing the recent past. The $w$ left singular vectors, $[\mathbf{r}_1,\mathbf{r}_2,\ldots,\mathbf{r}_w]$, define an orthonormal basis for the subspace defined by the $w$ activity vectors, $[\mathbf{u}^{t-w},\ldots, \mathbf{u}^{t-1}]$. Selecting only the first vector, $\mathbf{r}_1$, might cause us to loose information. Hence, a more representative profile vector can be obtained by projecting $\mathbf{u}^t$ onto the $w$ dimensional orthonormal subspace defined by $[\mathbf{r}_1,\mathbf{r}_2,\ldots,\mathbf{r}_w]$, where 
	\begin{equation}
	\bar{\mathbf{r}}^{t-1} = \left ( {\mathbf{r}_1 \cdot \mathbf{u}^t}\right ) \mathbf{r}_1 + \ldots \left ( {\mathbf{r}_w \cdot \mathbf{u}^t}\right ) \mathbf{r}_w.
	\end{equation}
	The profile vector, $\bar{\mathbf{r}}^{t-1}$ is also the best approximation to $\mathbf{u}^t$ in the subspace spanned by  $[\mathbf{u}^{t-w},\ldots, \mathbf{u}^{t-1}]$ \cite{poole2014linear}. The error vector, $\mathbf{e}^t = \bar{\mathbf{r}}^{t-1} - \mathbf{u}^t $, gives an indication of the deviation of $\mathbf{u}^t$ from its recent past. Thus, the change score, ${z}_i^t$, is
	\begin{equation}
	{z}_i^t = |\bar{{r}}_i^{t-1} - {u}_i^t|.
	\end{equation}
\end{enumerate}

\subsection{Results}
\label{Results}

\begin{figure}[]
	\centering
	\includegraphics[trim = 60mm 0mm 0mm 0mm, scale=0.25]{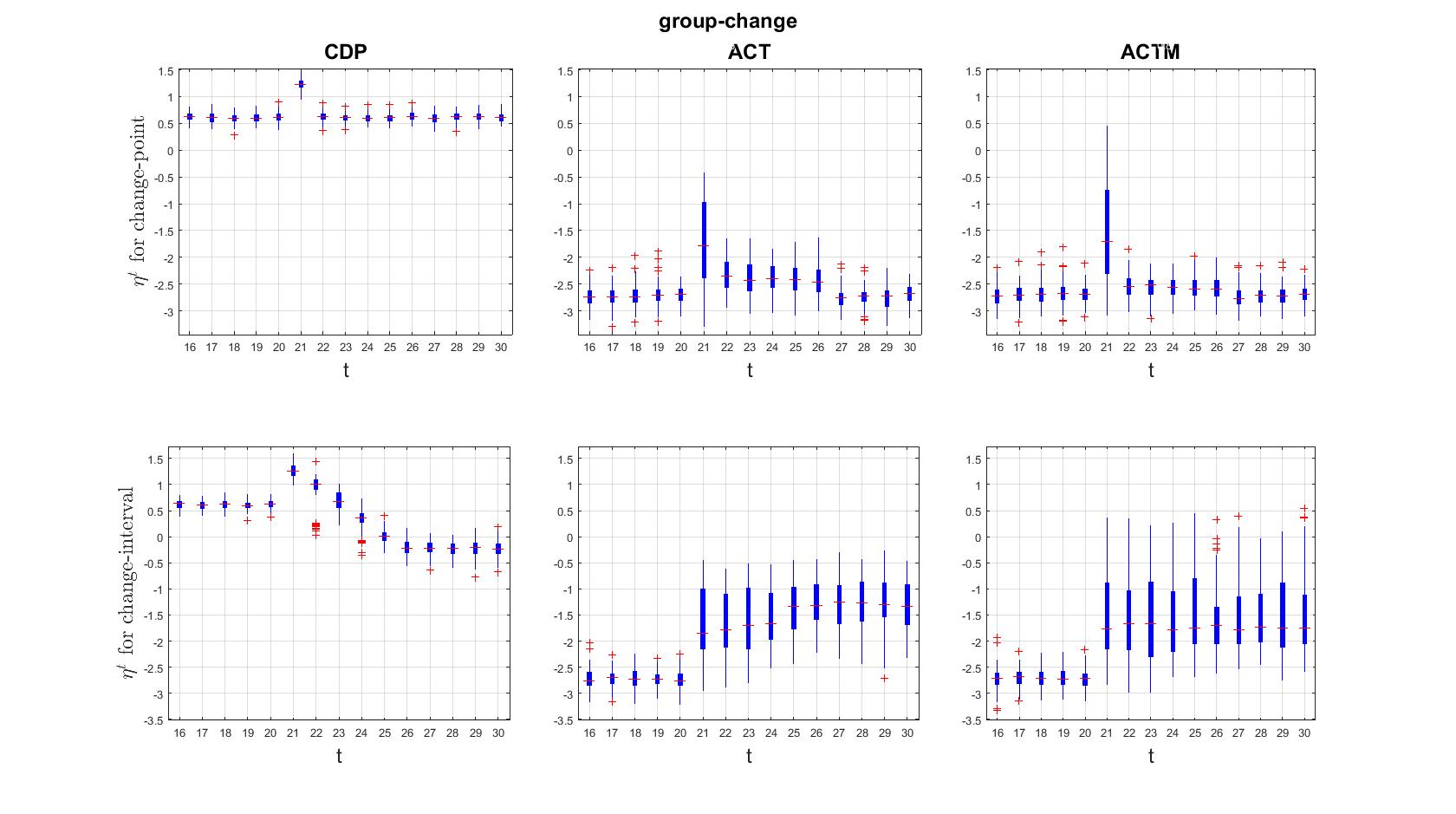}
	\caption{\textbf{Observing $\pmb{\eta}^{t}$ on CDP (left), ACT (middle), and ACTM (right) over time on group-change for $w=5$ change point(top) and change-interval (bottom)}. CDP shows a clear detection at $t=21$ for both change point and change-interval. Although ACT and ACTM methods also show an increase at $t=21$, the intervals still lie below zero.}
	\label{fig:grpOdds}	
\end{figure}

For each change scenario discussed in Section \ref{Change Scenarios}, we first calculate the performance measure ${\pmb{\eta}}^t$ for several time instants before and after $t=21$ for CDP, ACT, and ACTM for both change-point and change-interval. In Figure \ref{fig:grpOdds}, we show the corresponding results for group-change with $w=5$ for  $t=17,18,\ldots,30$.
Let us first discuss the results of CDP. For all time instants, before $t=21$, the ${\pmb{\eta}}^t$'s are centred at a given level. All graphs generated before $t=21$ are from the same model $\mathcal{M}1$. Since there is no model change, the odds of each $\mathbf{z}_{n_c}^t$ being greater than $\mathbf{z}_{n_{\bar{c}}}^t$ are similar during these time instants. At $t=21$, the generative model changes to $\mathcal{M}4$, and we see a clear increase of ${\pmb{\eta}}^{21}$ compared to ${\pmb{\eta}}^{20}$. This shows that there is a clear increase in $\mathbf{z}_{n_c}^{21}$. From $t=22$ onwards, we observe different patterns for change-point and change-interval.
\begin{itemize}[noitemsep,topsep=0pt]
	\item Change-point: the generative model returns to $\mathcal{M}1$ at $t=22$ and persists for all time instants, $t=23,\ldots,30$.
	\begin{enumerate}
		\item There is a big decrease in ${\pmb{\eta}}^{22}$ compared to ${\pmb{\eta}}^{21}$. Our window is $w=5$. Inside the window, there are four graphs generated from $\mathcal{M}1$, and one graph generated from $\mathcal{M}4$. Thus, unlike at $t=21$, there is less change compared to the recent past. Thus, $\pmb{\eta}^{22}$ is less than $\pmb{\eta}^{21}$.
		%		Hence, $\mathbf{z}_{n_c}^{22}$ decreases resulting in $\pmb{\eta}^{22}$ being less than $\pmb{\eta}^{21}$.
		
		\item For $t=22,23,24,25,26$, the window contains four graphs generated from $\mathcal{M}1$, and one graph generated from $\mathcal{M}4$. Thus, the change occurring in these time instants is similar. So, the $\pmb{\eta}_{t}$'s are generally centred at the same level. 
		%		So, $\mathbf{z}_{n_c}^{t}$'s and the resulting $\pmb{\eta}_{t}$'s are also similar. 
		
		\item At $t=27$, the window contains graphs generated purely from $\mathcal{M}1$, and the comparison is also done with a graph generated from $\mathcal{M}1$. So the change involving the set of vertices, $V_c$, at $t=27$ is  less than the change at $t=26$. Hence, $\pmb{\eta}^{27}$ decreases. 
		
		\item For $t=27,28,\ldots$, the window contains graphs generated purely from $\mathcal{M}1$, and the comparison is also done with a graph generated from $\mathcal{M}1$. Thus, the change occurring in these time instants is similar. So, the $\pmb{\eta}_{t}$'s are generally centred at the same level. 
	\end{enumerate}
	\item Change-interval: the generative model is $\mathcal{M}4$ for time instants, $t=22,\ldots,30$.
	\begin{enumerate}[nosep]
		\item There is a decrease in ${\pmb{\eta}}^{22}$ compared to ${\pmb{\eta}}^{21}$. Inside the window there are four graphs generated from $\mathcal{M}1$, and one graph generated from $\mathcal{M}4$. So there is less change involving the set of vertices, $V_c$, compared to their change at $t=21$. Hence, $\pmb{\eta}^{22}$ is less than $\pmb{\eta}^{21}$.
		
		\item For all time instants, $t=23,24,25$, the change becomes less and less as the window (recent past) contains more time instants which are similar to the current time instant. So $\mathbf{z}_{n_c}^{t}$ decreases with time, causing $\pmb{\eta}^{t}$ to decrease accordingly. 
		
		\item At $t=26$, the window contains graphs generated purely from $\mathcal{M}4$, and the comparison is also done with a graph generated from $\mathcal{M}4$. So the change is less compared to the change at $t=25$. Thus, $\pmb{\eta}^{26}$ is less than $\pmb{\eta}^{25}$. 
		
		\item For $t=26,27,\ldots$, the window contains graphs generated purely from $\mathcal{M}4$, and the comparison is also done with a graph generated from $\mathcal{M}4$. Thus, the change occurring in these time instants is similar. So, the $\pmb{\eta}_{t}$'s are generally centred at the same level. 
		
	\end{enumerate}
\end{itemize}

For change detection methods ACT and ACTM, ${\pmb{\eta}}^{t}$'s are wider. Furthermore, the bulk of ${\pmb{\eta}}^{t}$ lies below zero for all time instants. Thus, although we see an increase in the ${\pmb{\eta}}^{21}$ intervals for ACT and ACTM, these methods do not perform well in detecting the change.

\begin{figure}[]
	\centering
	\includegraphics[trim = 60mm 0mm 0mm 0mm, scale=0.25]{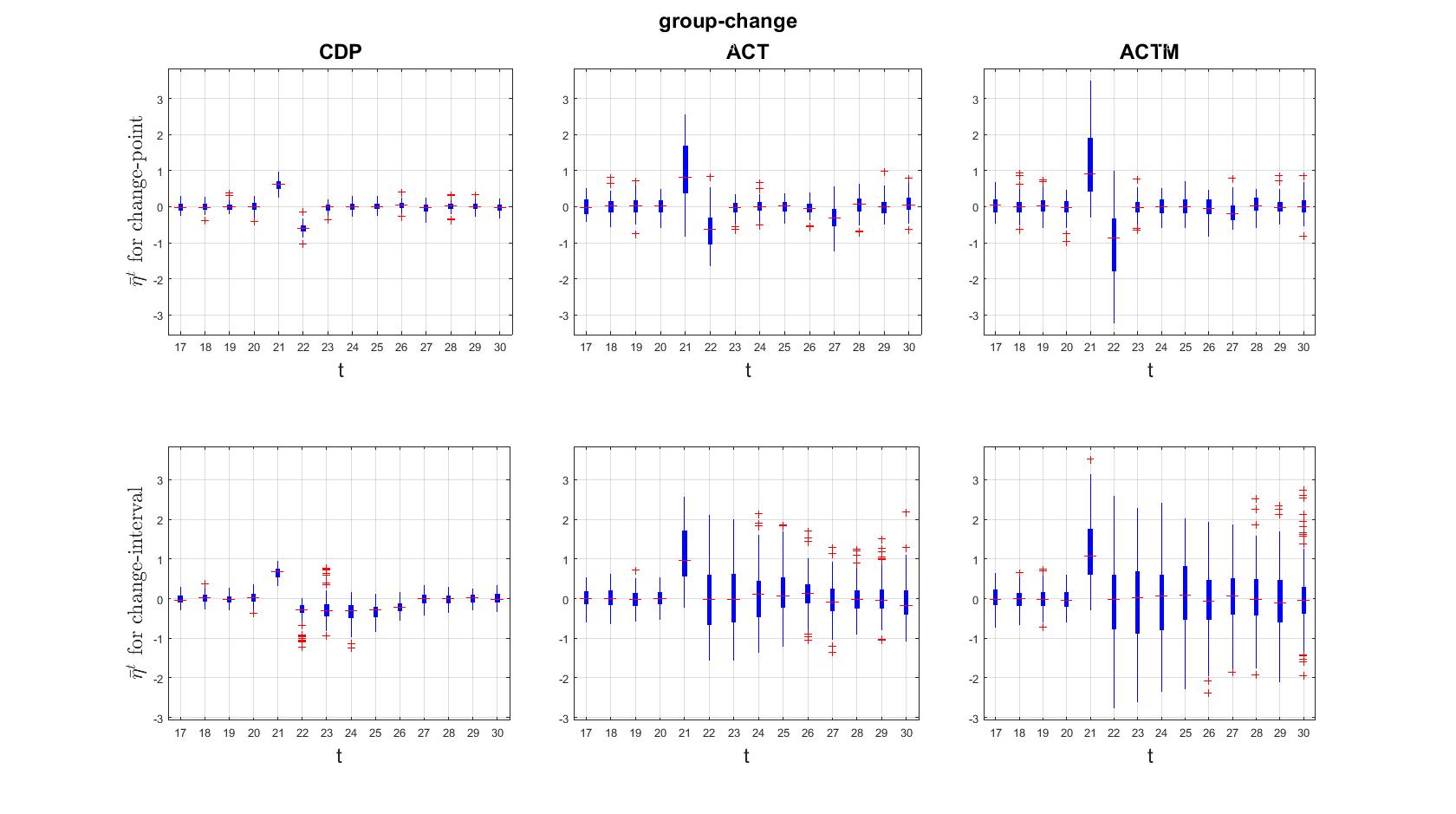}
	\caption{\textbf{Observing $\bar{\pmb{\eta}}^t$ on CDP (left), ACT (middle), and ACTM (right) over time on group-change for $w=5$ change point(top) and change-interval (bottom)}. }
	\label{fig:grpOddsRatio}
	
	\includegraphics[trim = 50mm 0mm 0mm 0mm, scale=0.22]{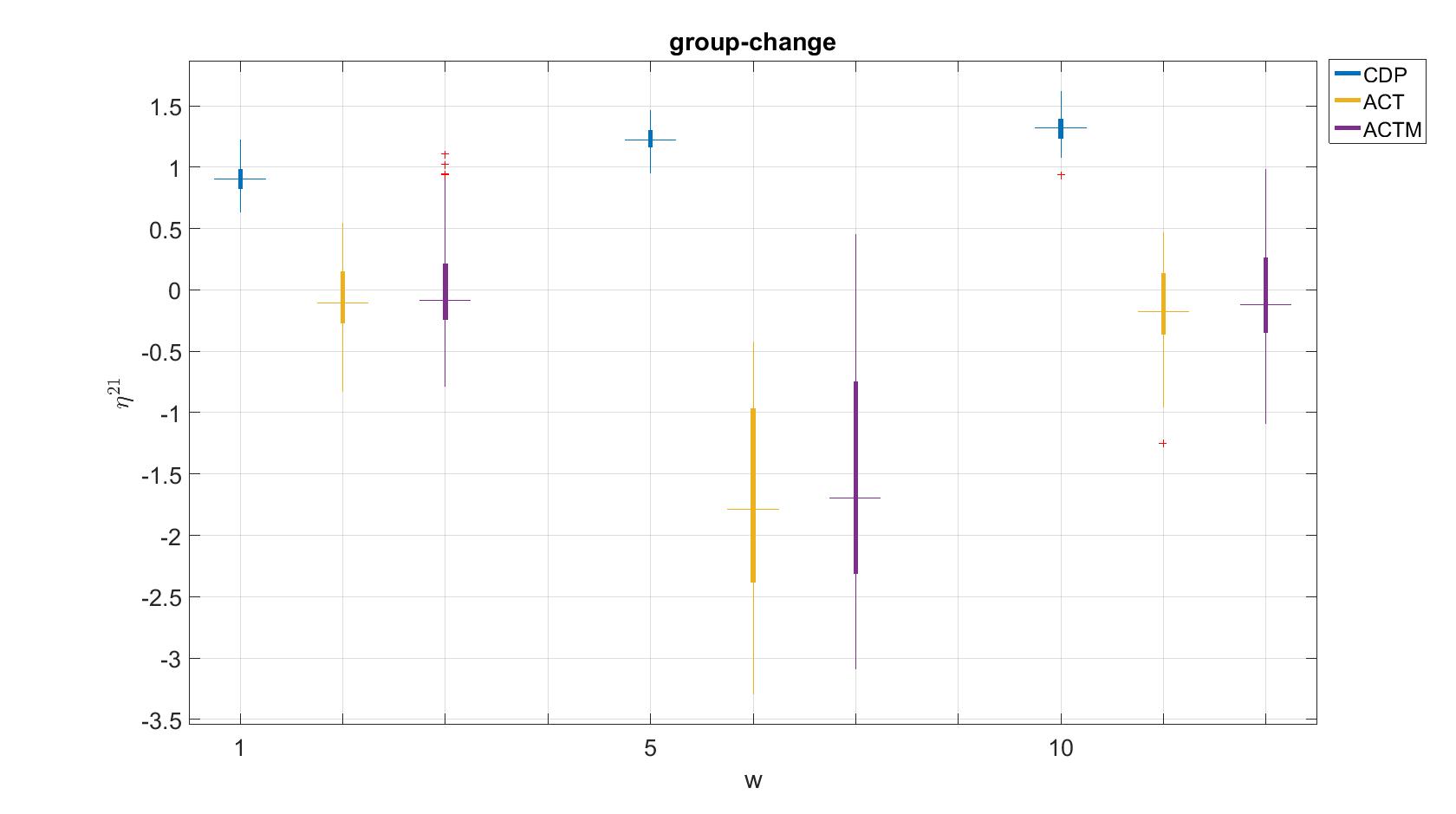}
	\caption{\textbf{Plot of $\pmb{\eta}^{21}$ on CDP, ACT, and ACTM for group-change for $w=1,5,10$ at change-point}. For all $w$, ${\pmb{\eta}^{21}}_{CDP}$ is positive and increases with $w$. For all $w$, a majority of elements of ${\pmb{\eta}^{21}}_{ACT}$ and ${\pmb{\eta}^{21}}_{ACTM}$ are negative.}
	\label{fig:grpOddsWindow}	
\end{figure}

\begin{figure}[]
	\centering
	%\includegraphics[trim = 50mm 0mm 0mm 0mm, scale=0.3]{Figures/split-odds}
	%  \caption{Observing $\text{odds}(\hat{\pmb{f}}_{t})$ on CDP, ACT and ACTM for split for $w=5$. CDP scores are positive and shows an increase at $t=21$ for both change-point and change-interval. ACT and ACTM give higher scores for non changed vertices, hence negative $\text{odds}(\hat{\pmb{f}}_{t})$ scores. ACT and ACTM methods fail to detect split.}
	%  \label{fig:splitOdds}
	\includegraphics[trim = 50mm 0mm 0mm 0mm, scale=0.22]{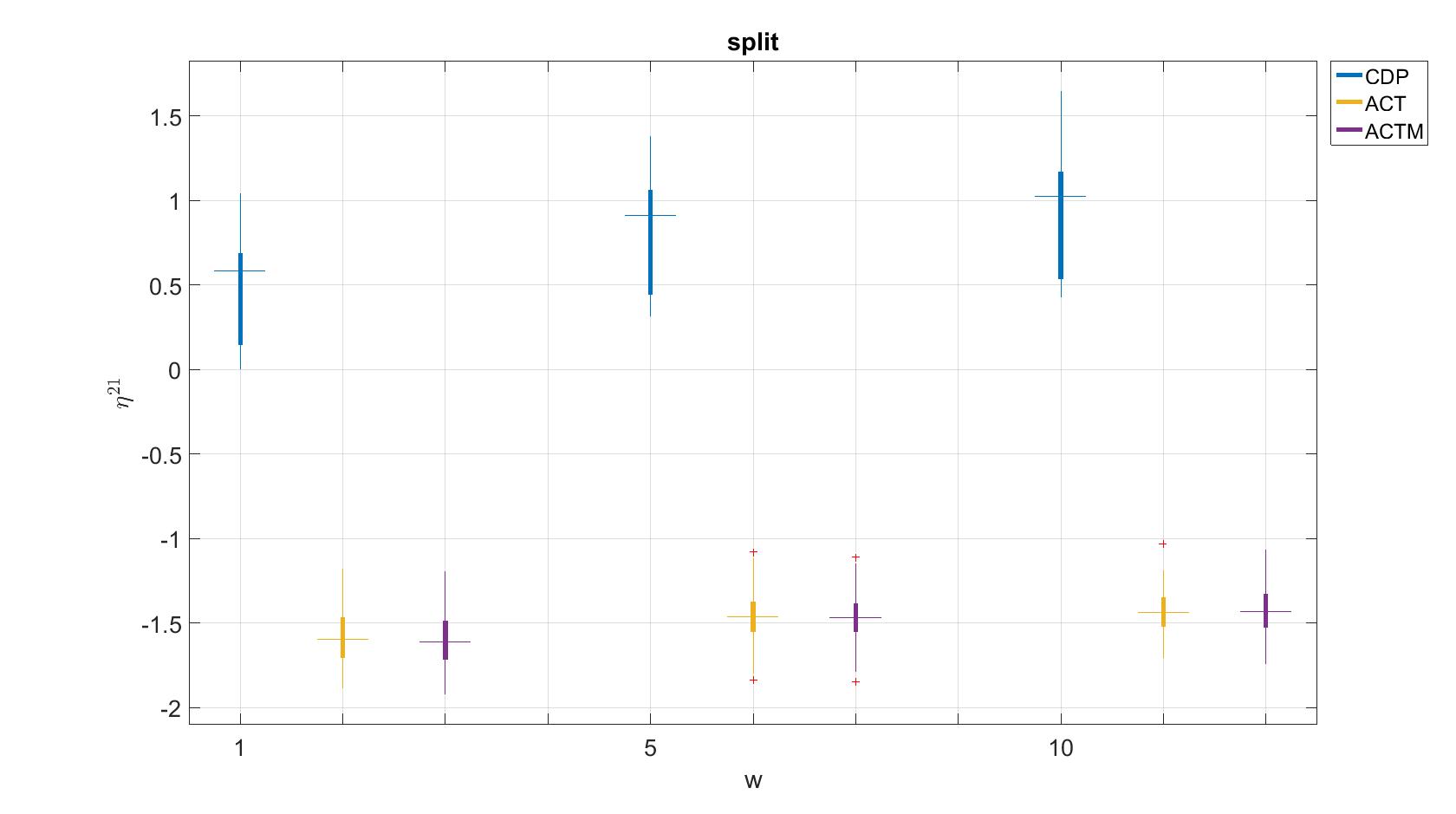}
	\caption{\textbf{Plot of $\pmb{\eta}^{21}$ on CDP, ACT, and ACTM for split for $w=1,5,10$ at change-point}. For all $w$, ${\pmb{\eta}^{21}}_{CDP}$ is positive and increases with $w$. For all $w$, ${\pmb{\eta}^{21}}_{ACT}$ and ${\pmb{\eta}^{21}}_{ACTM}$ are negative.}
	\label{fig:splitOddsWindow}
	\includegraphics[trim = 60mm 0mm 0mm 0mm, scale=0.25]{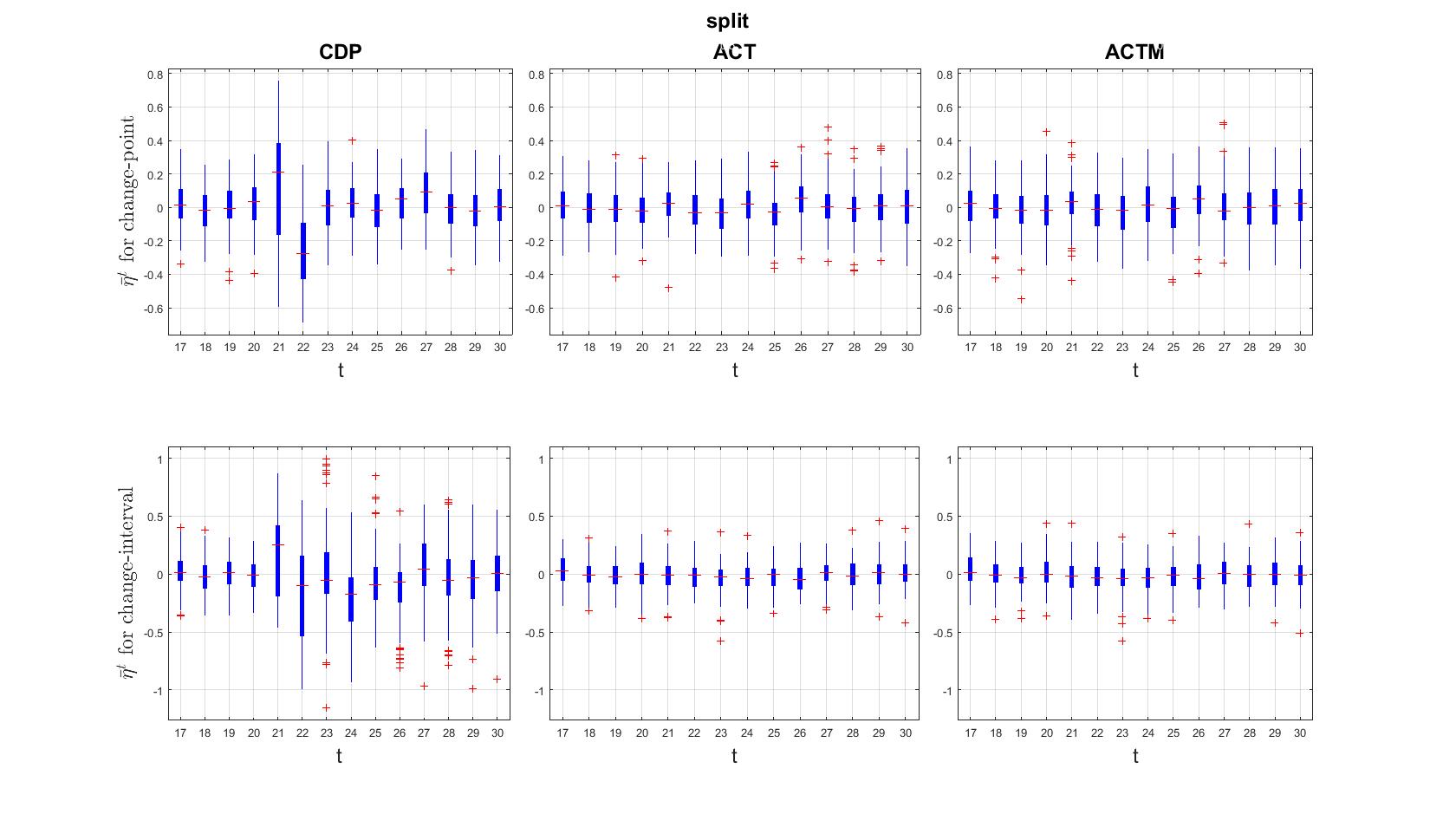}
	\caption{\textbf{Observing $\bar{\pmb{\eta}}^t$ on CDP (left), ACT (middle), and ACTM (right) over time on split for $w=5$ change point(top) and change-interval (bottom)}. Although ${\bar{\pmb{\eta}}}^{t}_{CDP}$ shows an increase at $t=21$ for both change point and change-interval, ${\bar{\pmb{\eta}}}^{21}_{CDP}$ shows high variability further extending below zero. ACT and ACTM do not show an increase at $t=21$.}
	\label{fig:splitOddsRatio}
	
\end{figure}

\begin{figure}[]
	\centering
	%\includegraphics[trim = 50mm 30mm 20mm 20mm, scale=0.3]{Figures/merge-odds}
	%  \caption{Observing $\text{odds}(\hat{\pmb{f}}_{t})$ on CDP, ACT and ACTM for merge for $w=5$. CDP scores do not show an increase at $t=21$ for both change-point and change-interval. CDP fails to detect merge scenario. ACT and ACTM give higher scores for non changed vertices, hence negative $\text{odds}(\hat{\pmb{f}}_{t})$ scores. ACT and ACTM methods fail to detect merge.}
	%  \label{fig:mergeOdds}
	\includegraphics[trim = 50mm 0mm 0mm 0mm, scale=0.22]{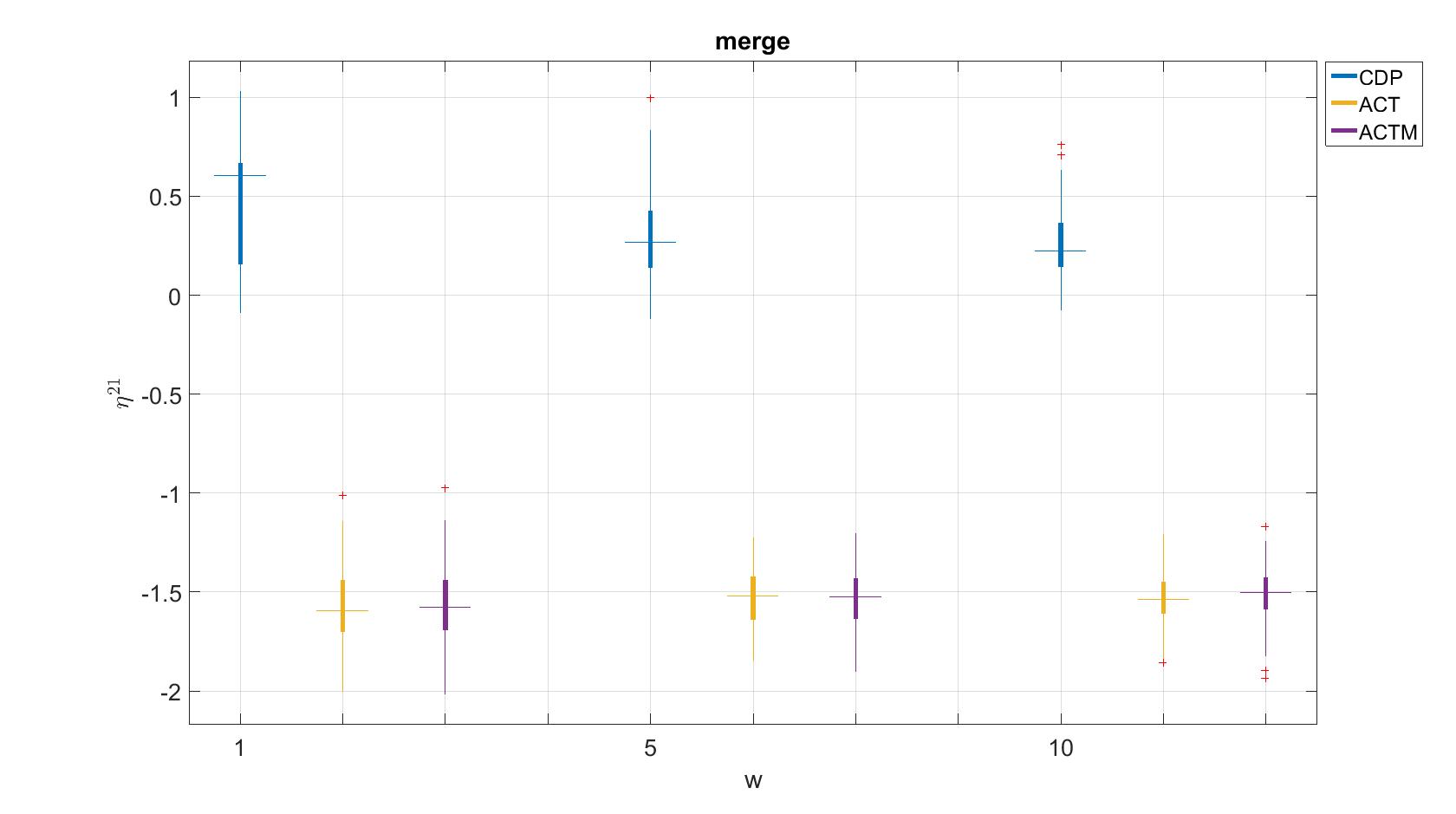}
	\caption{\textbf{Plot of $\pmb{\eta}^{21}$ on CDP, ACT, and ACTM for merge for $w=1,5,10$ at change-point}. For all $w$, ${\pmb{\eta}^{21}}_{CDP}$ is positive and decreases with $w$. For all $w$, ${\pmb{\eta}^{21}}_{ACT}$ and ${\pmb{\eta}^{21}}_{ACTM}$ are negative.}
	\label{fig:mergeOddsWindow}
	\includegraphics[trim = 60mm 0mm 00mm 0mm, scale=0.25]{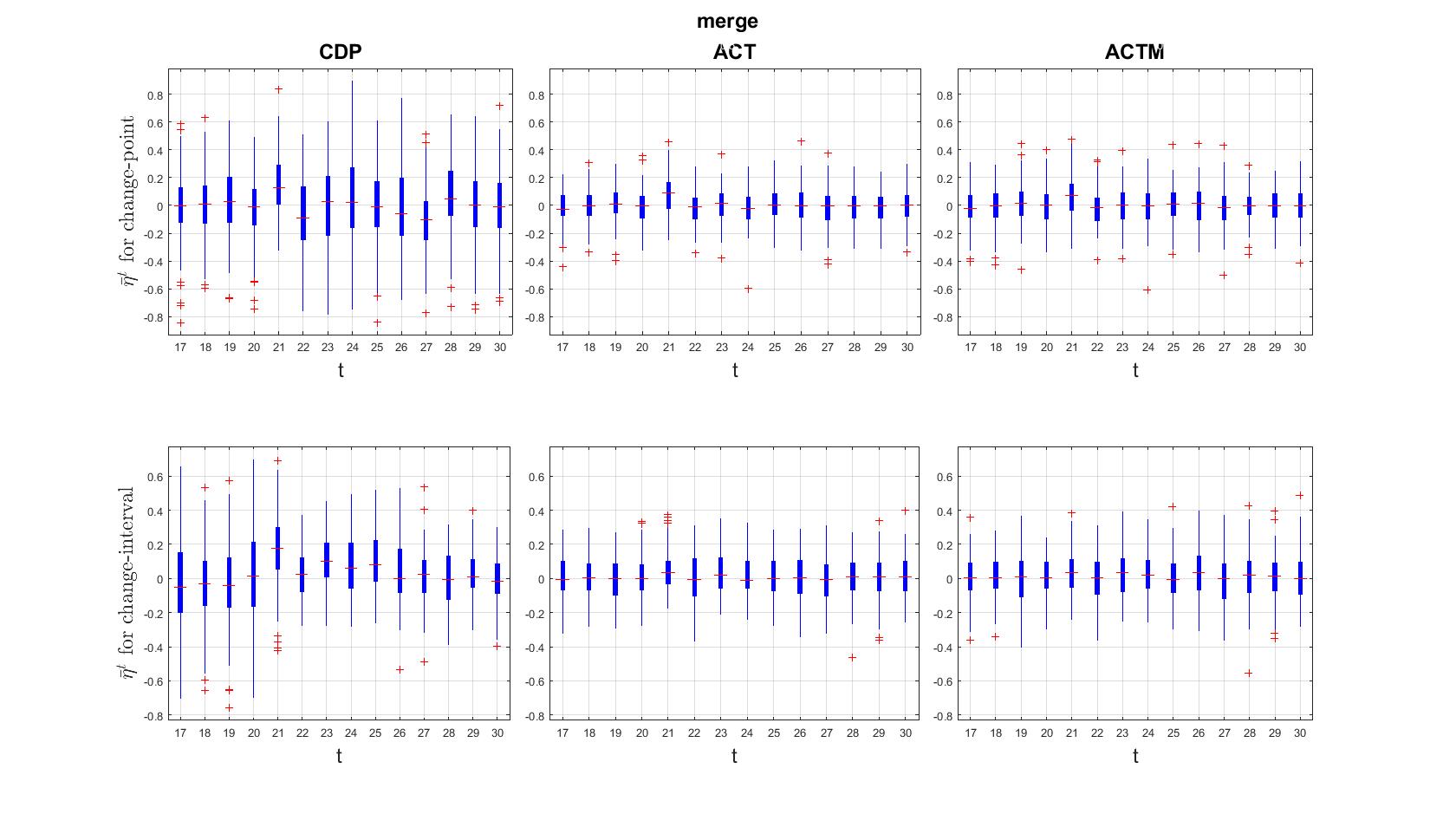}
	\caption{\textbf{Observing $\bar{\pmb{\eta}}^t$ on CDP (left), ACT (middle), and ACTM (right) over time on merge for $w=5$ change point(top) and change-interval (bottom)}. Although ${\bar{\pmb{\eta}}}^{t}_{CDP}$ shows an increase at $t=21$ for both change point and change-interval, ${\bar{\pmb{\eta}}}^{21}_{CDP}$ shows high variance, further extending below zero. ACT and ACTM do not show a clear increase at $t=21$ for both change-point and change-interval.}
	\label{fig:mergeOddsRatio}
	
\end{figure}

\begin{figure}[]
	\centering
	%\includegraphics[trim = 50mm 30mm 20mm 20mm, scale=0.27]{Figures/form-odds}
	%  \caption{Observing $\text{odds}(\hat{\pmb{f}}_{t})$ on CDP, ACT and ACTM for form for $w=5$. CDP scores are positive and shows a clear increase at $t=21$ for both change-point and change-interval. ACT and ACTM also show high positive scores at change-point as well.}
	%  \label{fig:formOdds}
	\includegraphics[trim = 50mm 0mm 0mm 0mm, scale=0.22]{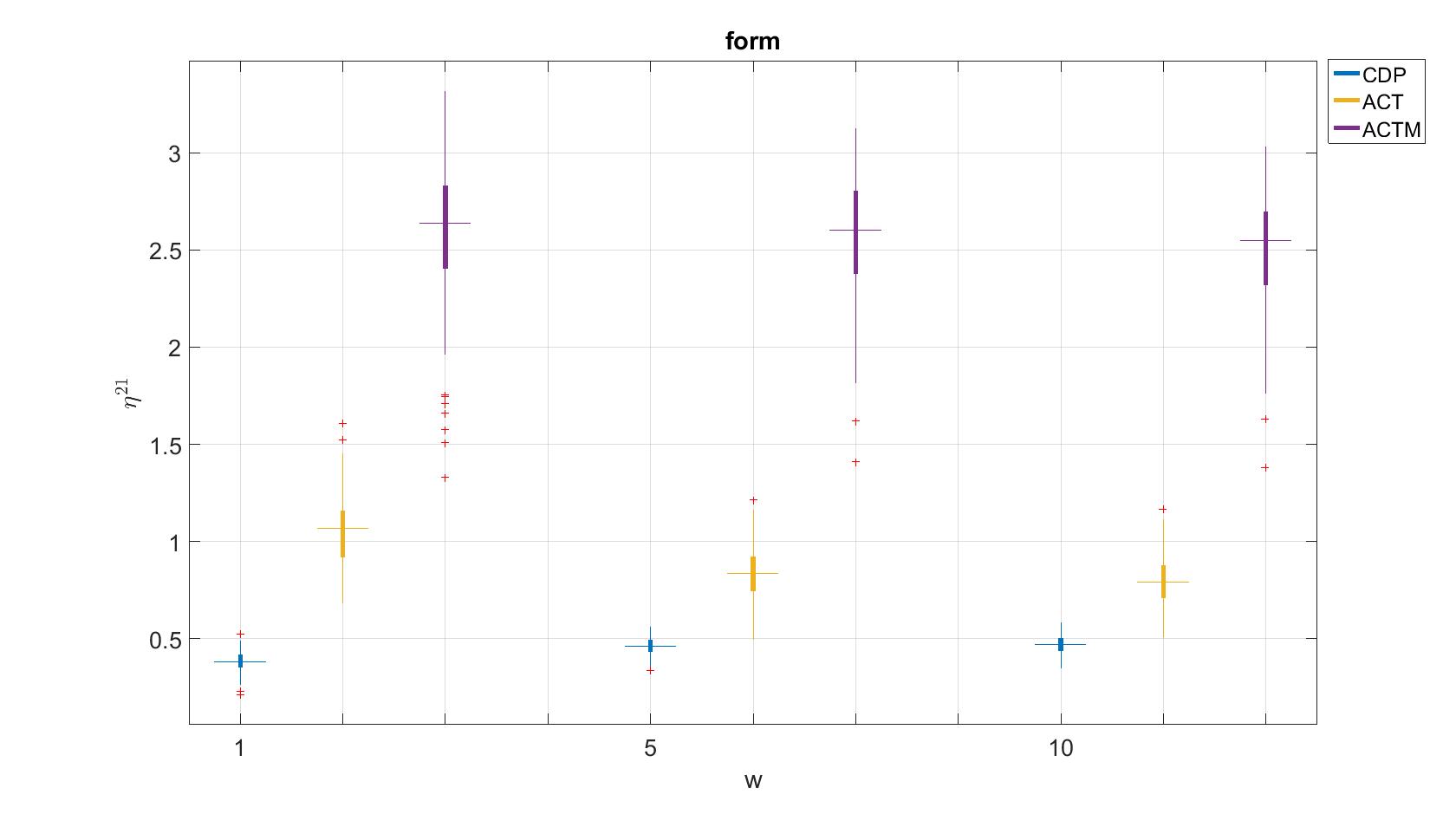}
	\caption{\textbf{Plot of $\pmb{\eta}^{21}$ on CDP, ACT, and ACTM for form for $w=1,5,10$ at change-point}. For all $w$, ${\pmb{\eta}^{21}}_{CDP}$, ${\pmb{\eta}^{21}}_{ACT}$, and ${\pmb{\eta}^{21}}_{ACTM}$ are positive and increase with $w$. However, ${\pmb{\eta}^{21}}_{ACTM}$ is wider and has more outliers.}
	\label{fig:formOddsWindow}
	\includegraphics[trim = 60mm 00mm 0mm 00mm, scale=0.25]{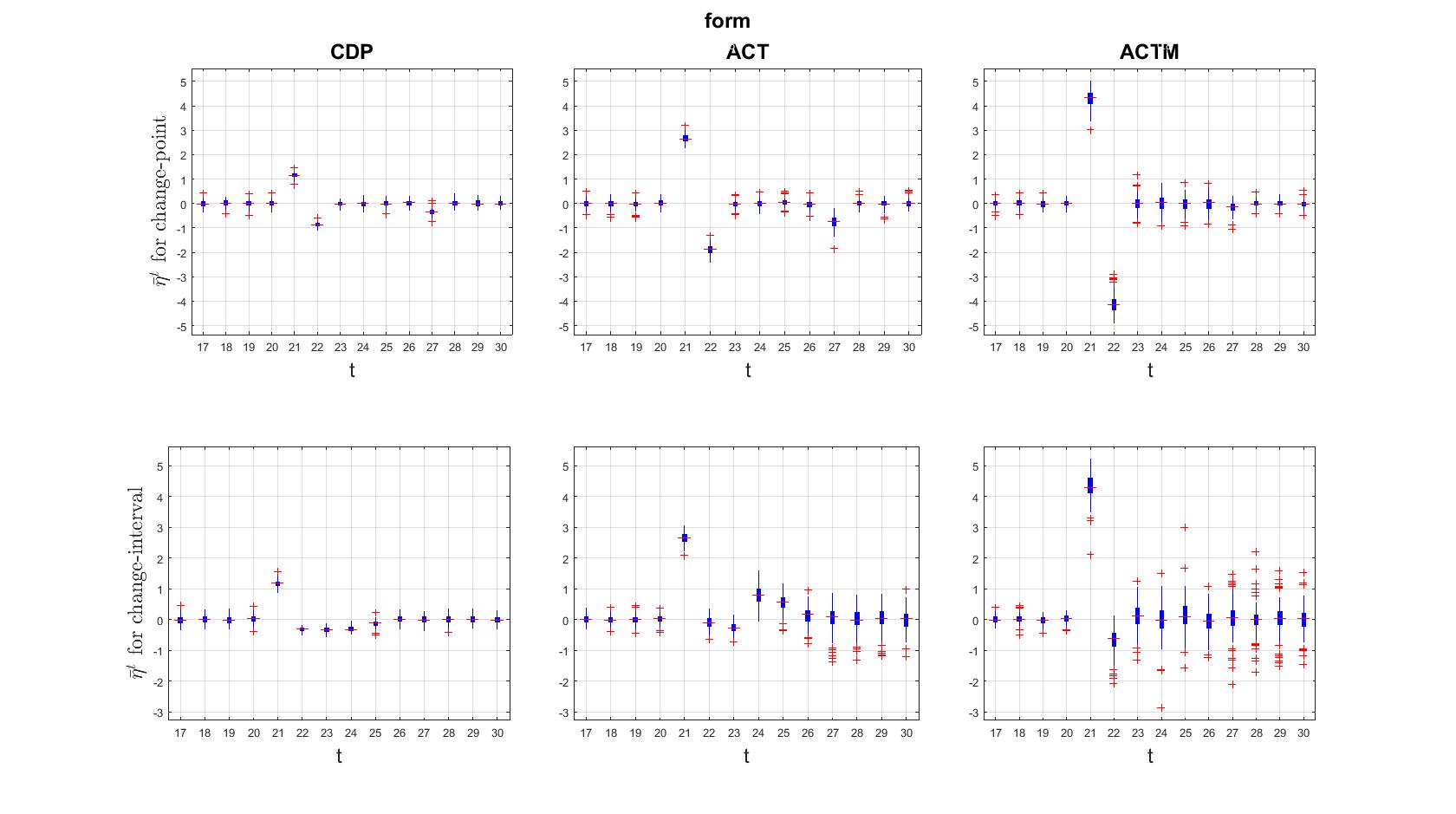}
	\caption{\textbf{Observing $\bar{\pmb{\eta}}^t$ on CDP (left), ACT (middle), and ACTM (right) over time on form for $w=5$ change point(top) and change-interval (bottom)}. All ${\bar{\pmb{\eta}}}^{t}_{CDP}$, ${\bar{\pmb{\eta}}}^{t}_{ACT}$ and ${\bar{\pmb{\eta}}}^{t}_{ACTM}$ show a clear increase at $t=21$, while ${\bar{\pmb{\eta}}}^{21}_{ACTM}$ shows the highest increase.}
	\label{fig:formOddsRatio}
	
\end{figure}

\begin{figure}[]
	\centering
	%\includegraphics[trim = 50mm 0mm 0mm 0mm, scale=0.3]{Figures/fragment-odds}
	%  \caption{Observing $\text{odds}(\hat{\pmb{f}}_{t})$ on CDP, ACT and ACTM for fragment for $w=5$. CDP scores show a clear increase at $t=21$ for both change-point and change-interval. Although ACT and ACTM show positive $\text{odds}(\hat{\pmb{f}}_{t})$ scores, the scores degrade at $t=21$. Hence, ACT and ACTM methods fail to detect fragment.}
	%  \label{fig:fragmentOdds}
	
	\includegraphics[trim = 50mm 00mm 0mm 0mm, scale=0.22]{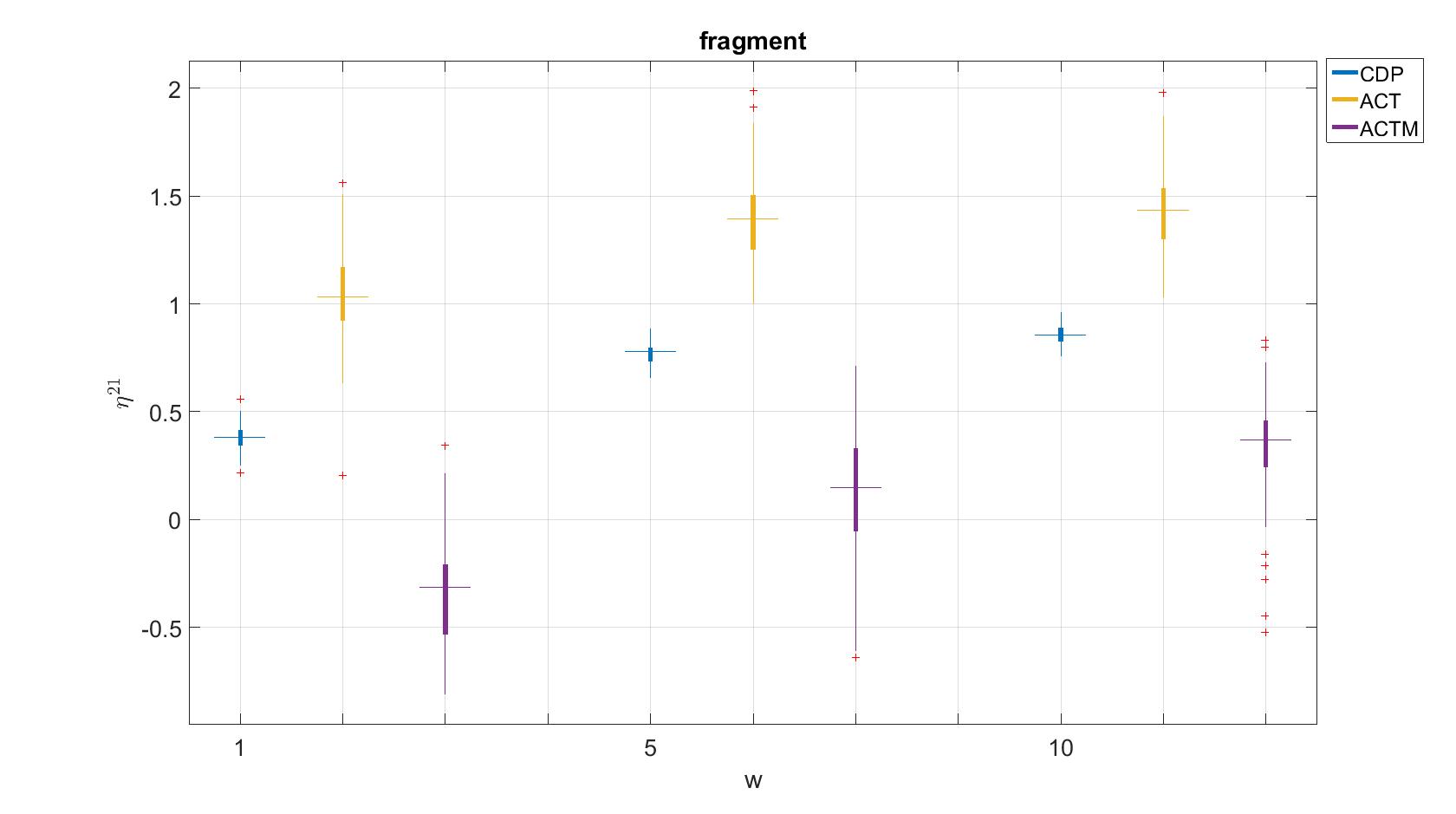}
	\caption{\textbf{Plot of $\pmb{\eta}^{21}$ on CDP, ACT, and ACTM for fragment for $w=1,5,10$ at change-point}. For all $w$, ${\pmb{\eta}^{21}}_{CDP}$ and ${\pmb{\eta}^{21}}_{ACT}$ are positive and increase with $w$. ${\pmb{\eta}^{21}}_{ACTM}$ is mostly negative at $w=1$, but increases with $w$. However, ${\pmb{\eta}^{21}}_{ACT}$ and ${\pmb{\eta}^{21}}_{ACTM}$ are wider and show more outliers.}
	\label{fig:fragmentOddsWindow}
	\includegraphics[trim = 60mm 0mm 0mm 0mm, scale=0.25]{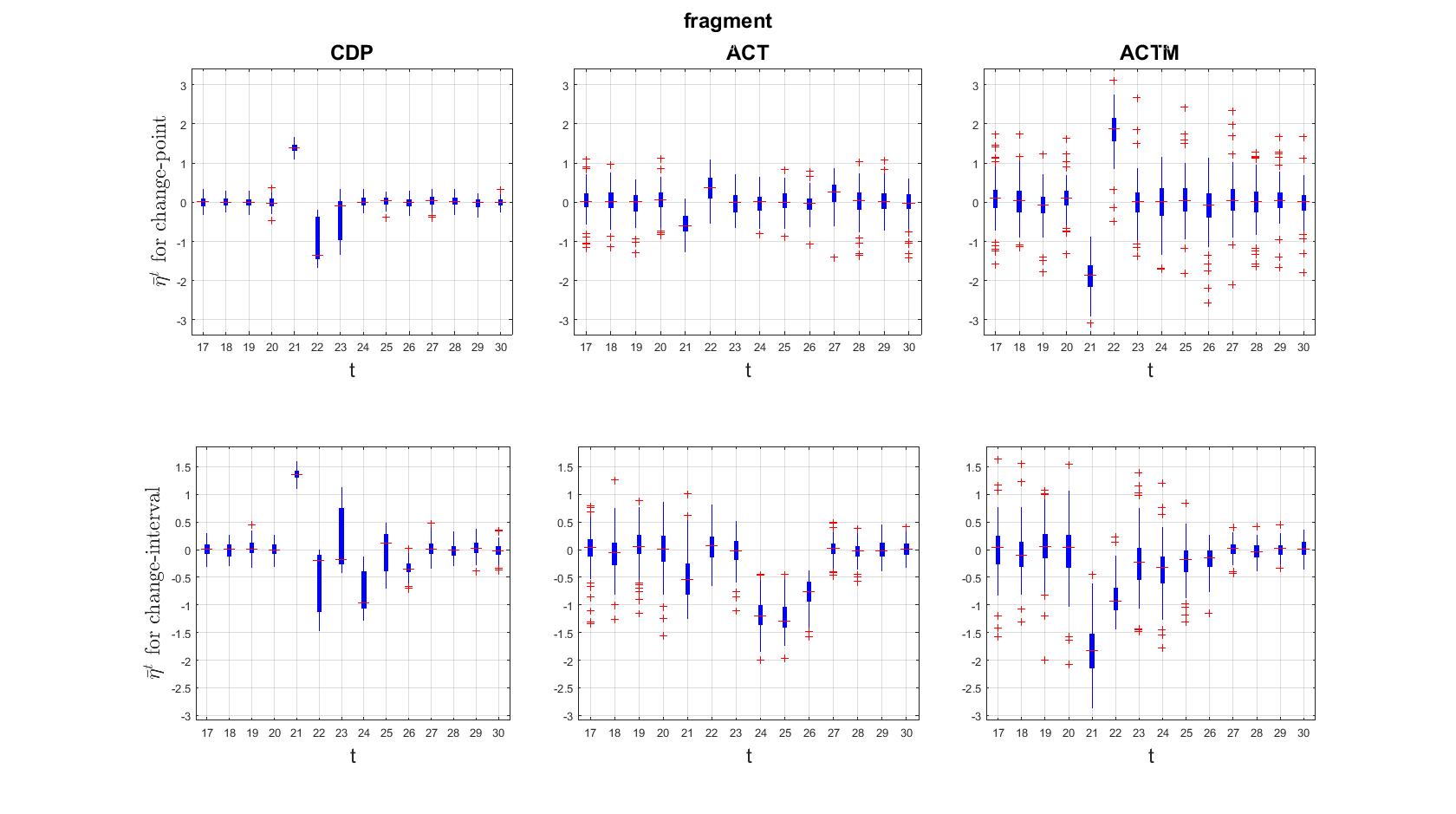}
	\caption{\textbf{Observing $\bar{\pmb{\eta}}^t$ on CDP (left), ACT (middle), and ACTM (right) over time on fragment for $w=5$ change point(top) and change-interval (bottom)}. ${\bar{\pmb{\eta}}}^{21}_{CDP}$ shows a clear detection at $t=21$ for both change point and change-interval. Both ${\bar{\pmb{\eta}}}^{21}_{ACT}$ and ${\bar{\pmb{\eta}}}^{21}_{ACTM}$ are negative. Furthermore ${\bar{\pmb{\eta}}}^{t}_{ACT}$ and ${\bar{\pmb{\eta}}}^{t}_{ACTM}$ contain a large number of outliers.} 
	\label{fig:fragmentOddsRatio}
	
\end{figure}

\begin{figure}[]
	\centering
	%\includegraphics[trim = 50mm 30mm 0mm 0mm, scale=0.3]{Figures/hetro2homo-odds}
	%  \caption{Observing $\text{odds}(\hat{\pmb{f}}_{t})$ on CDP, ACT and ACTM for hetero-to-homo for $w=5$. CDP scores are positive and shows an increase at $t=21$ for both change-point and change-interval. ACT and ACTM give higher scores for non changed vertices, hence negative $\text{odds}(\hat{\pmb{f}}_{t})$ scores. ACT and ACTM methods fail to detect hetero-to-homo.}
	%  \label{fig:hetro2homoOdds}
	\includegraphics[trim = 50mm 0mm 0mm 0mm, scale=0.22]{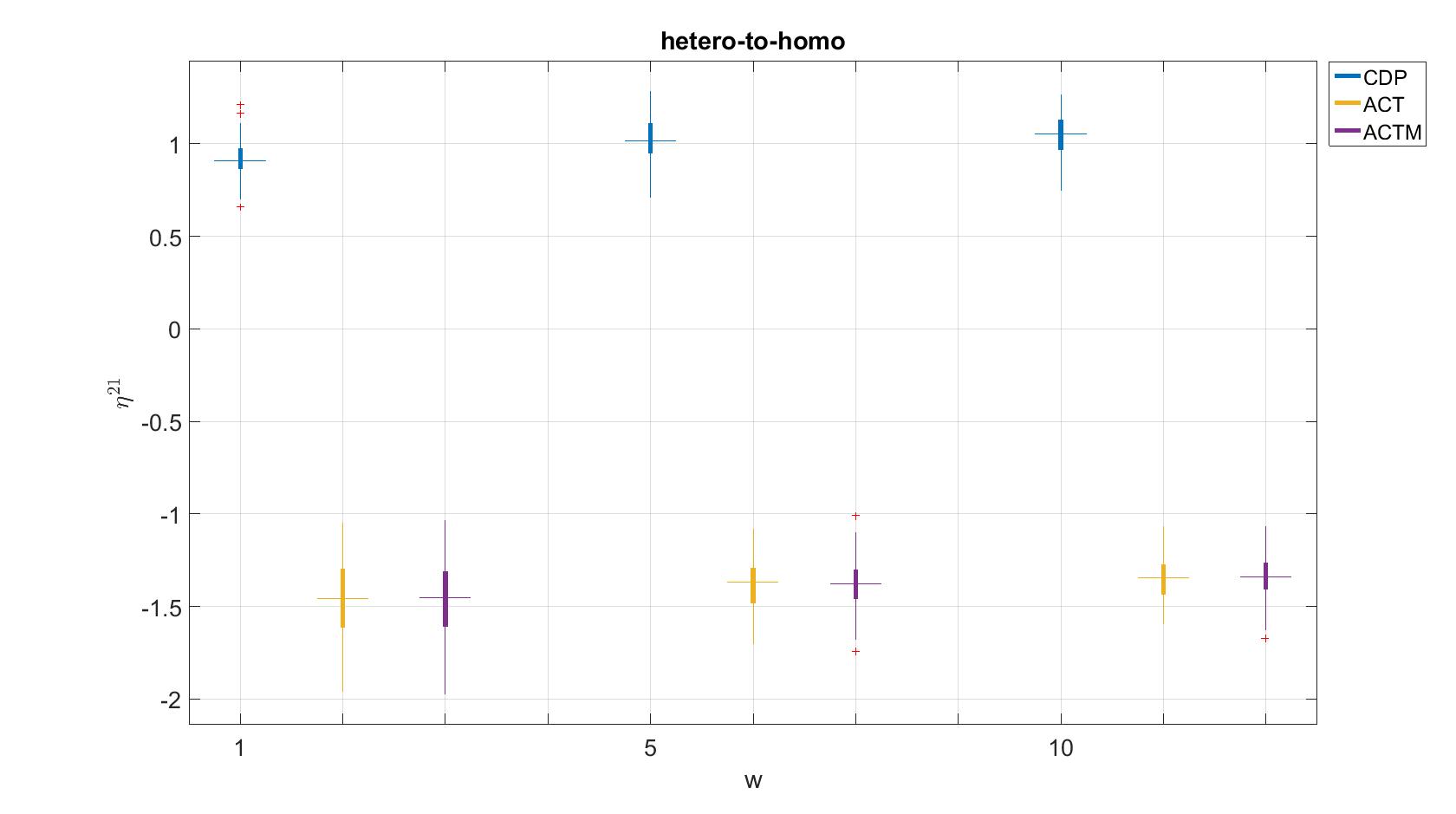}
	\caption{\textbf{Plot of $\pmb{\eta}^{21}$ on CDP, ACT, and ACTM for hetero-to-homo for $w=1,5,10$ at change-point}. For all $w$, ${\pmb{\eta}^{21}}_{CDP}$ are positive and slightly increase with $w$. The ${\pmb{\eta}^{21}}_{ACT}$ and ${\pmb{\eta}^{21}}_{ACTM}$ are negative for all $w$.}
	\label{fig:hetro2homoOddsWindow}
	\includegraphics[trim = 60mm 0mm 0mm 0mm, scale=0.25]{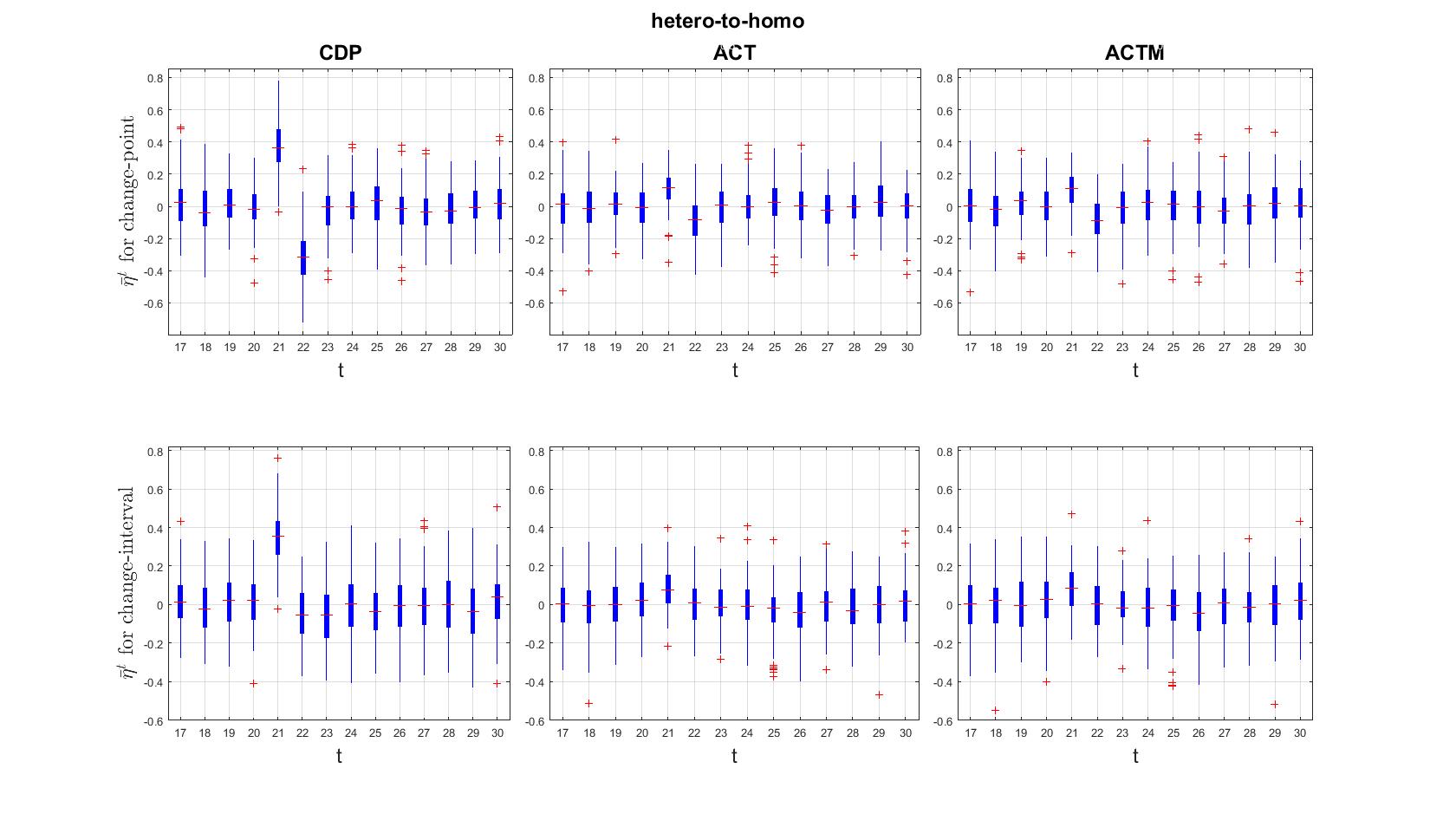}
	\caption{\textbf{Observing $\bar{\pmb{\eta}}^t$ on CDP (left), ACT (middle), and ACTM (right) over time on hetero-to-homo for $w=5$ change point(top) and change-interval (bottom)}. ${\bar{\pmb{\eta}}}^{21}_{CDP}$ shows a clear detection at $t=21$ for both change point and change-interval. Both ${\bar{\pmb{\eta}}}^{21}_{ACT}$ and ${\bar{\pmb{\eta}}}^{21}_{ACTM}$ show a slight increase, but still some values lie below zero.} 
	\label{fig:hetro2homoOddsRatio}
	
\end{figure}

\begin{figure}[]
	\centering
	%\includegraphics[trim = 50mm 30mm 0mm 0mm, scale=0.3]{Figures/homo2hetero-odds}
	%  \caption{Observing $\text{odds}(\hat{\pmb{f}}_{t})$ on CDP, ACT and ACTM for homo-to-hetero for $w=5$. CDP scores are positive and shows an increase at $t=21$ for both change-point and change-interval. ACT and ACTM give higher scores for non changed vertices, hence negative $\text{odds}(\hat{\pmb{f}}_{t})$ scores. ACT and ACTM methods fail to detect homo-to-hetero.}
	%  \label{fig:homo2heteroOdds}
	\includegraphics[trim = 50mm 0mm 0mm 0mm, scale=0.22]{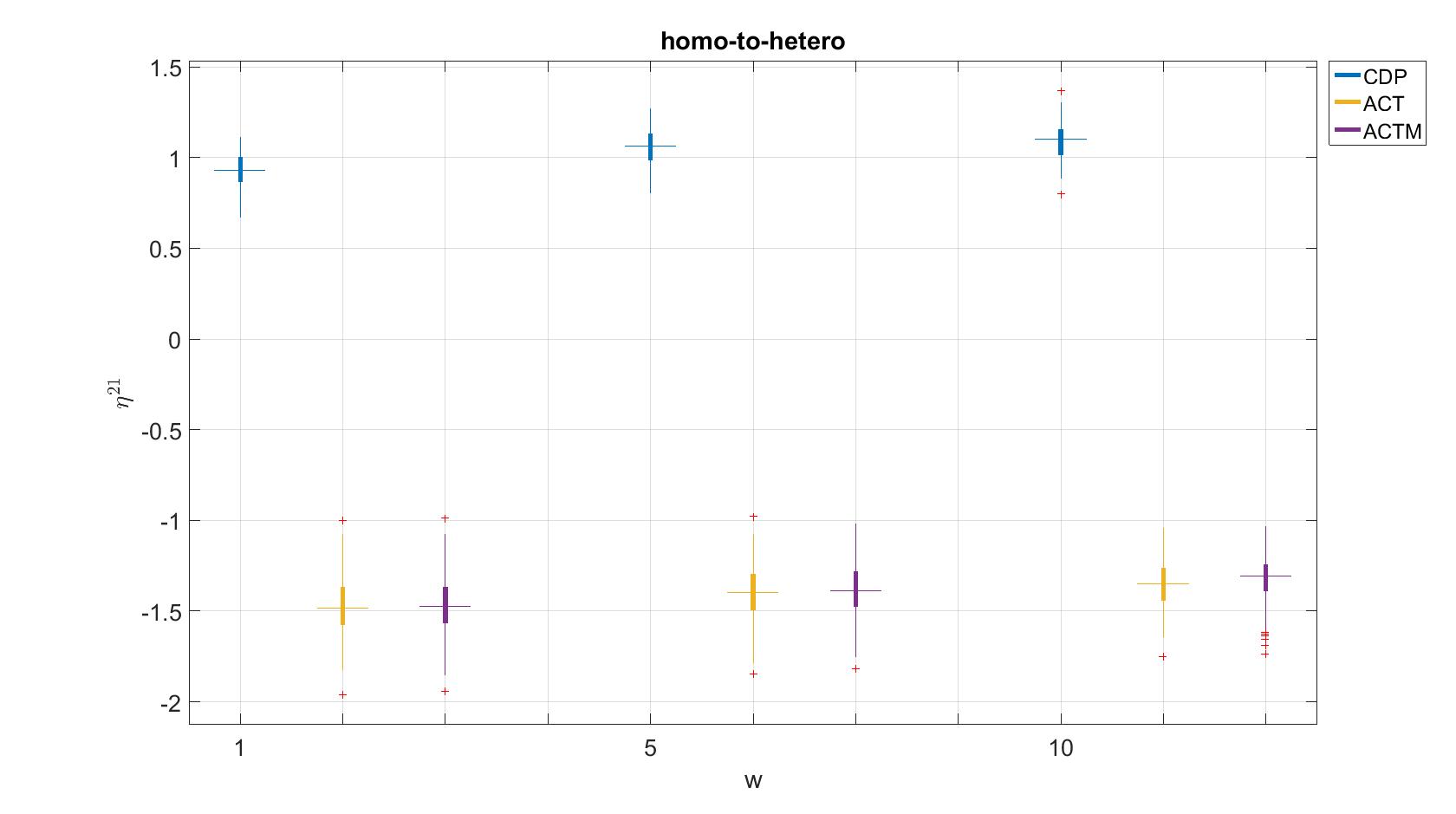}
	\caption{\textbf{Plot of $\pmb{\eta}^{21}$ on CDP, ACT, and ACTM for homo-to-hetero for $w=1,5,10$ at change-point}. For all $w$, ${\pmb{\eta}^{21}}_{CDP}$ are positive and slightly increase with $w$. The ${\pmb{\eta}^{21}}_{ACT}$ and ${\pmb{\eta}^{21}}_{ACTM}$ are negative for all $w$.}
	\label{fig:homo2heteroOddsWindow}
	\includegraphics[trim = 60mm 0mm 0mm 0mm, scale=0.25]{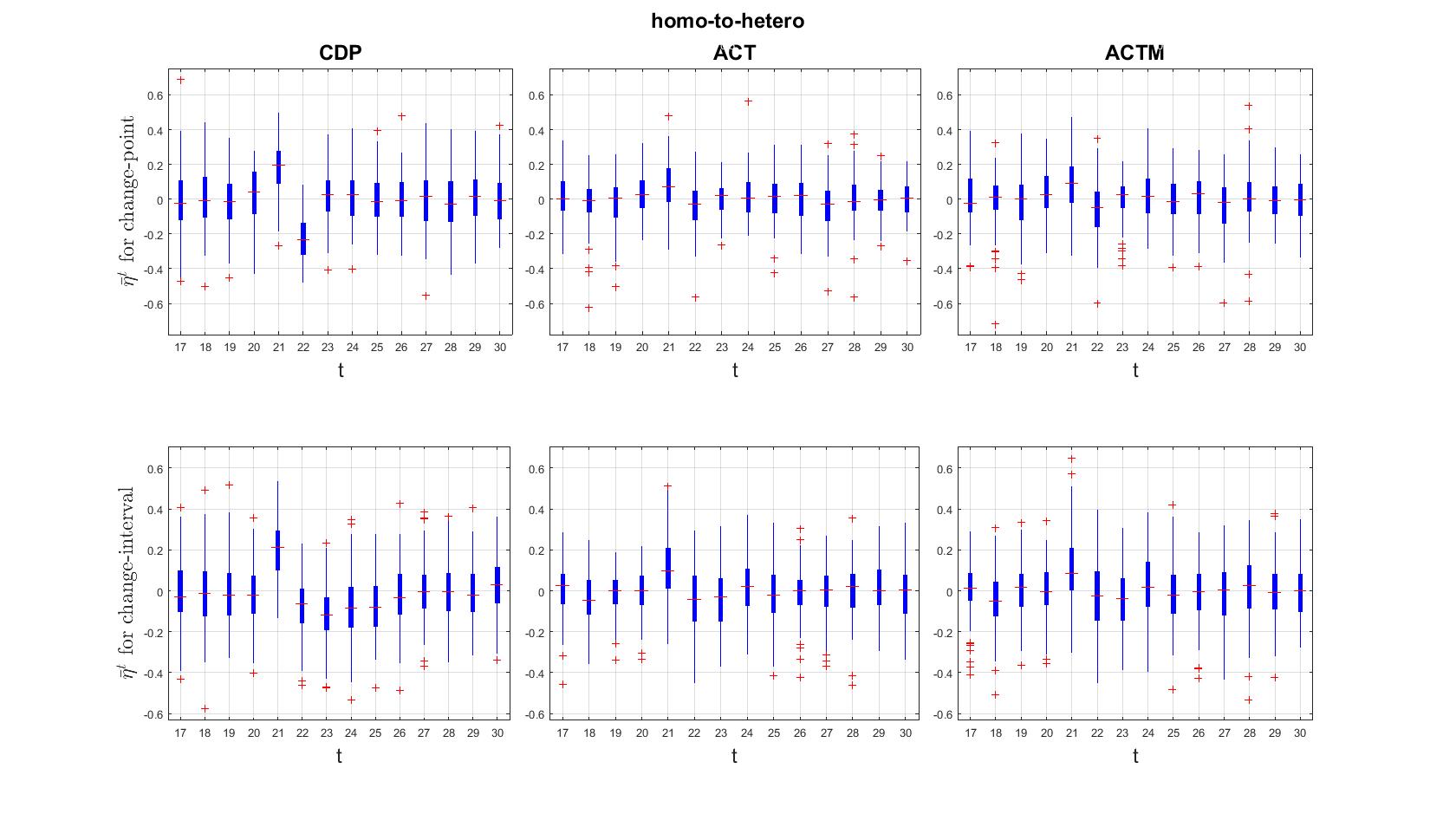}
	\caption{\textbf{Observing $\bar{\pmb{\eta}}^t$ on CDP (left), ACT (middle), and ACTM (right) over time on homo-to-hetero for $w=5$ change point(top) and change-interval (bottom)}. ${\bar{\pmb{\eta}}}^{21}_{CDP}$ shows a clear detection at $t=21$ for both change point and change-interval, but ${\bar{\pmb{\eta}}}^{21}_{CDP}$ slightly extends below zero. Both ${\bar{\pmb{\eta}}}^{21}_{ACT}$ and ${\bar{\pmb{\eta}}}^{21}_{ACTM}$ also show a slight increase, with the intervals extending below zero.} 
	\label{fig:homo2heteroOddsRatio}
	
\end{figure}

\begin{figure}[]
	\centering
	%\includegraphics[trim = 50mm 30mm 0mm 0mm, scale=0.3]{Figures/simple2complex-odds}
	%  \caption{Observing $\text{odds}(\hat{\pmb{f}}_{t})$ on CDP, ACT and ACTM for simple-to-complex for $w=5$. CDP scores are positive and shows an increase at $t=21$ for both change-point and change-interval. ACT and ACTM give higher scores for non changed vertices, hence negative $\text{odds}(\hat{\pmb{f}}_{t})$ scores. ACT and ACTM methods fail to detect simple-to-complex.}
	%  \label{fig:simple2complexOdds}
	\includegraphics[trim = 50mm 0mm 0mm 0mm, scale=0.22]{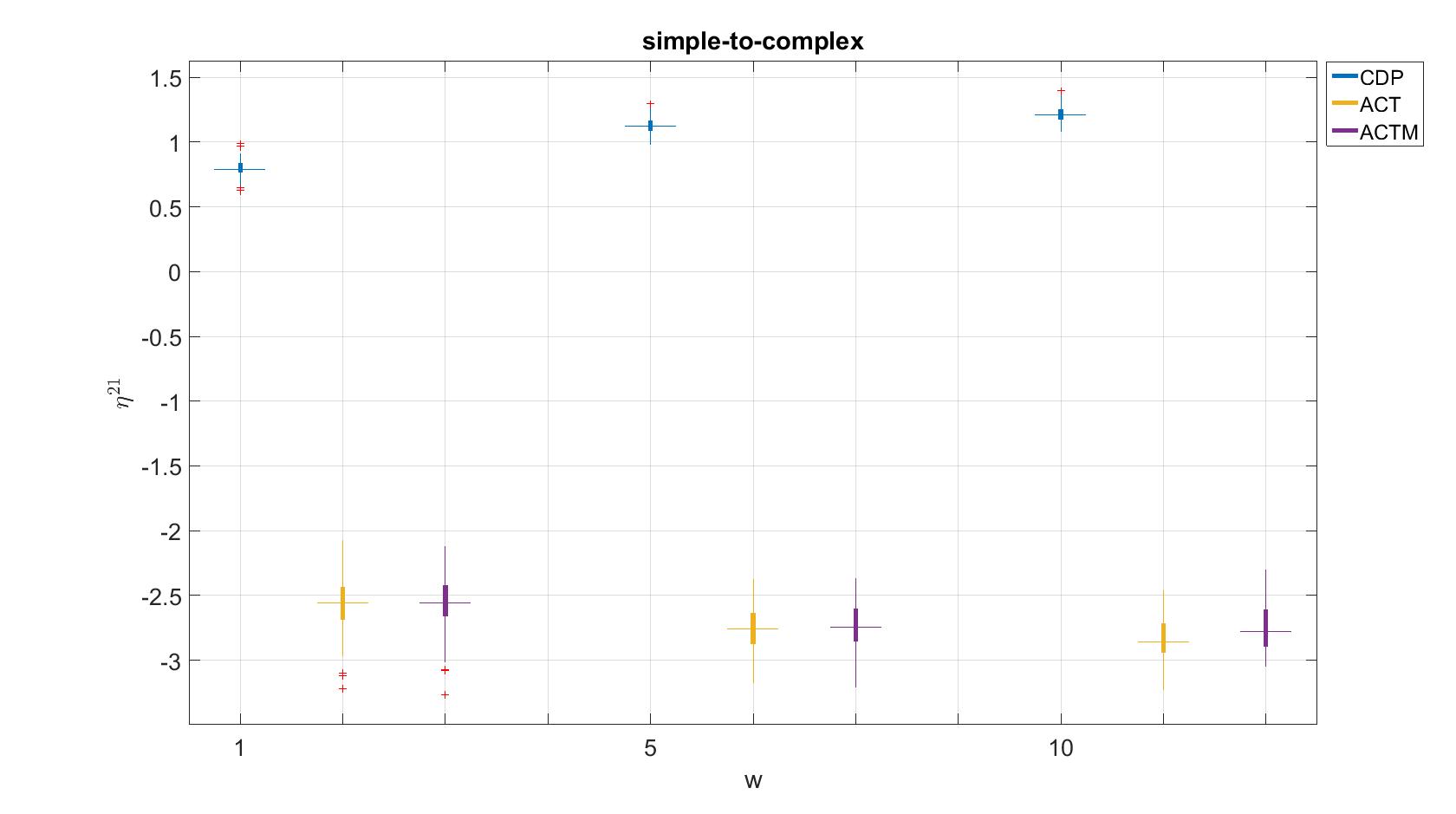}
	\caption{\textbf{Plot of $\pmb{\eta}^{21}$ on CDP, ACT, and ACTM for simple-to-complex for $w=1,5,10$ at change-point}. For all $w$, ${\pmb{\eta}^{21}}_{CDP}$ are positive and increase with $w$. The ${\pmb{\eta}^{21}}_{ACT}$ and ${\pmb{\eta}^{21}}_{ACTM}$ are negative for all $w$.}
	\label{fig:simple2complexOddsWindow}
	\includegraphics[trim = 60mm 00mm 0mm 0mm, scale=0.25]{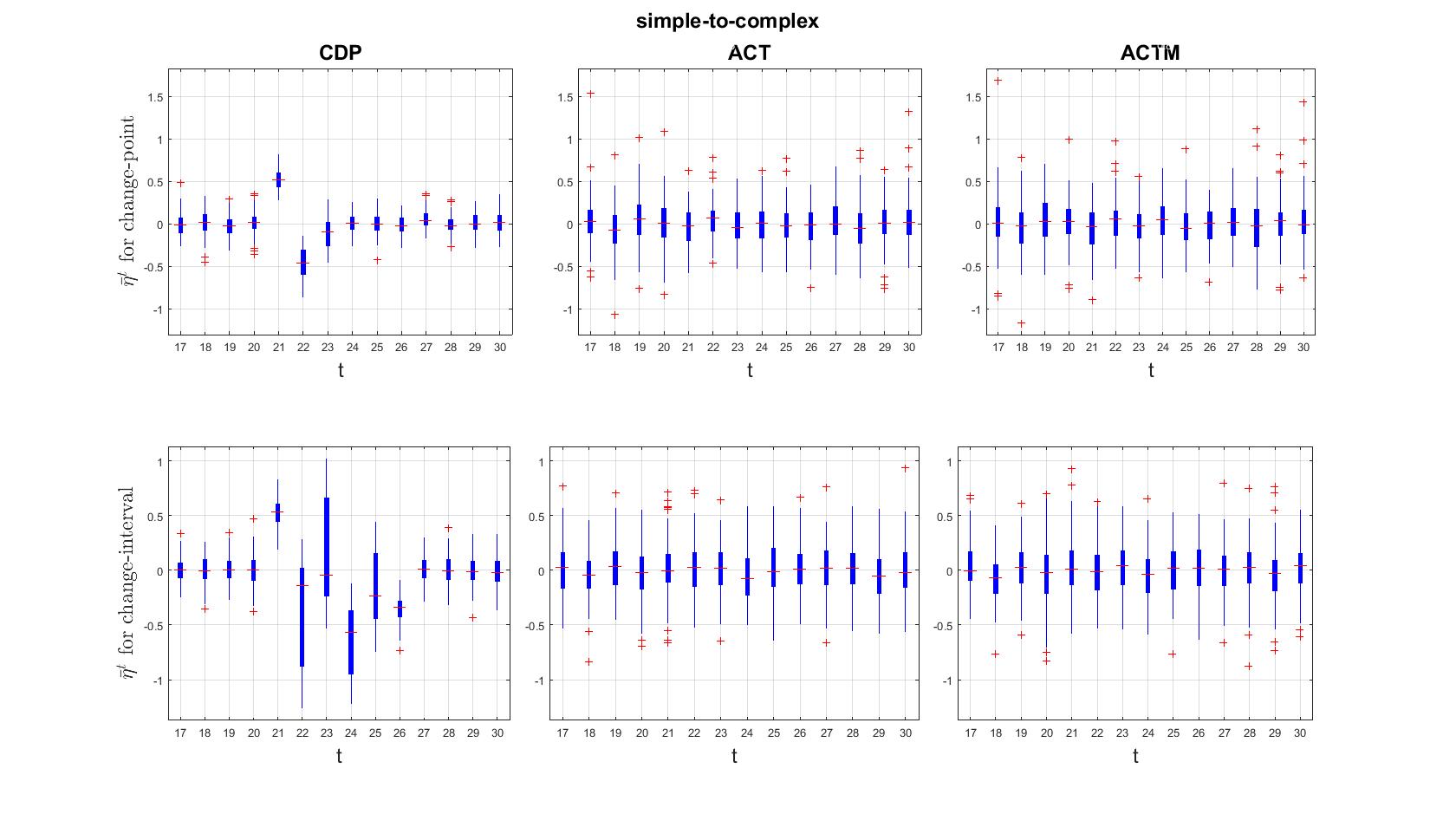}
	\caption{\textbf{Observing $\bar{\pmb{\eta}}^t$ on CDP (left), ACT (middle), and ACTM (right) over time on simple-to-complex for $w=5$ change point(top) and change-interval (bottom)}. We observe a clear detection at ${\bar{\pmb{\eta}}}^{21}_{CDP}$ for change-point. For change-interval we observe ${\bar{\pmb{\eta}}}^{t}_{CDP}$  for $t=22,\ldots,25$ to have high variance. A detection is not observed on both ${\bar{\pmb{\eta}}}^{21}_{ACT}$ and ${\bar{\pmb{\eta}}}^{21}_{ACTM}$.}
	\label{fig:simple2complexOddsRatio}
	
\end{figure}

\begin{figure}[]
	\centering
	%\includegraphics[trim = 50mm 0mm 0mm 0mm, scale=0.3]{Figures/complex2simple-odds}
	%  \caption{Observing $\text{odds}(\hat{\pmb{f}}_{t})$ on CDP, ACT and ACTM for complex-to-simple for $w=5$. CDP scores are positive and shows an increase at $t=21$ for both change-point and change-interval. ACT and ACTM give higher scores for non changed vertices, hence negative $\log \mathcal{OR}$ scores. ACT and ACTM methods fail to detect complex-to-simple.}
	%  \label{fig:complex2simpleOdds}
	\includegraphics[trim = 50mm 0mm 0mm 0mm, scale=0.22]{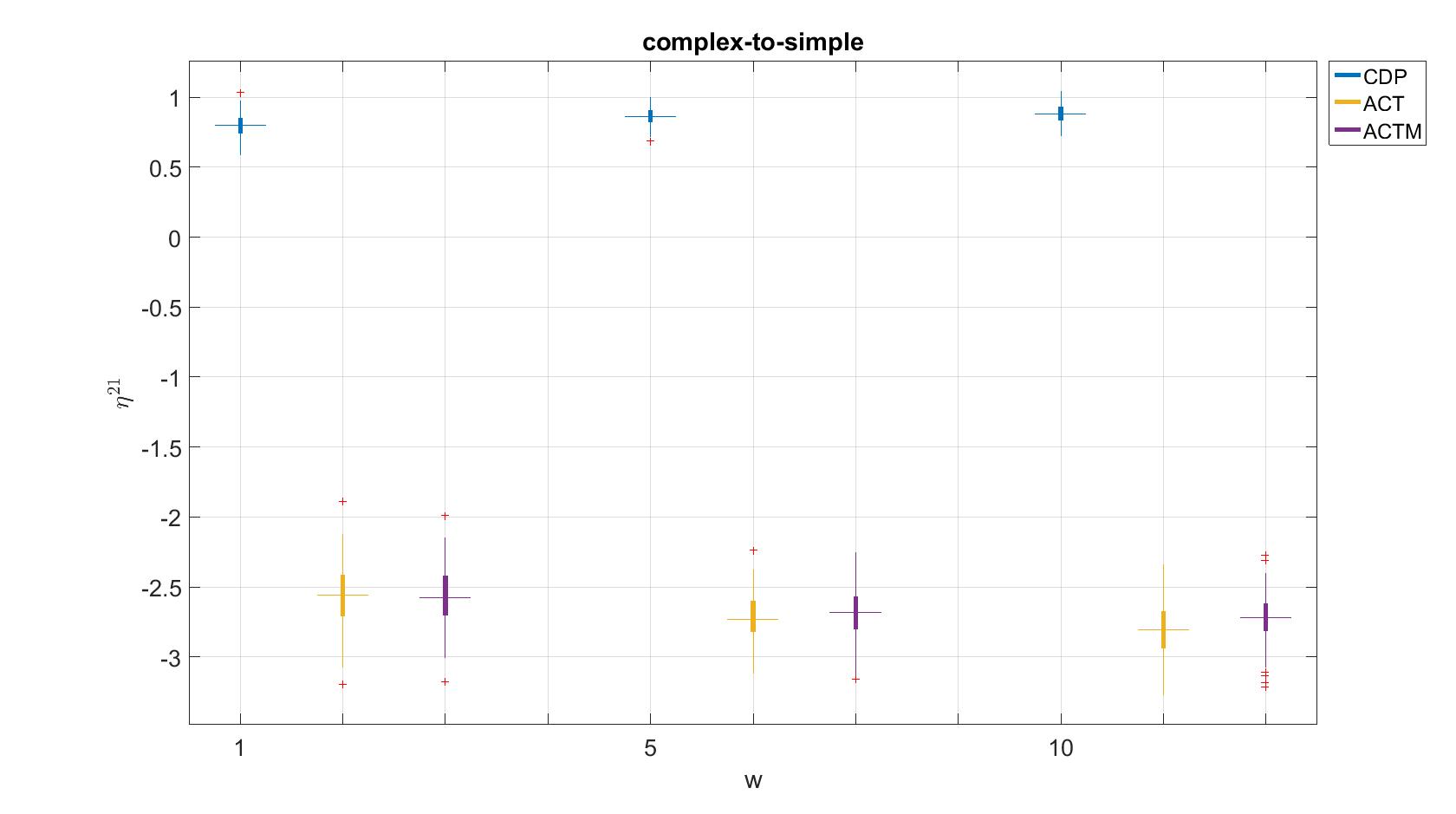}
	\caption{\textbf{Plot of $\pmb{\eta}^{21}$ on CDP, ACT, and ACTM for complex-to-simple for $w=1,5,10$ at change-point}. For all $w$, ${\pmb{\eta}^{21}}_{CDP}$ are positive and slightly increase with $w$. The ${\pmb{\eta}^{21}}_{ACT}$ and ${\pmb{\eta}^{21}}_{ACTM}$ are negative for all $w$.}
	\label{fig:complex2simpleOddsWindow}
	\includegraphics[trim = 60mm 0mm 0mm 0mm, scale=0.25]{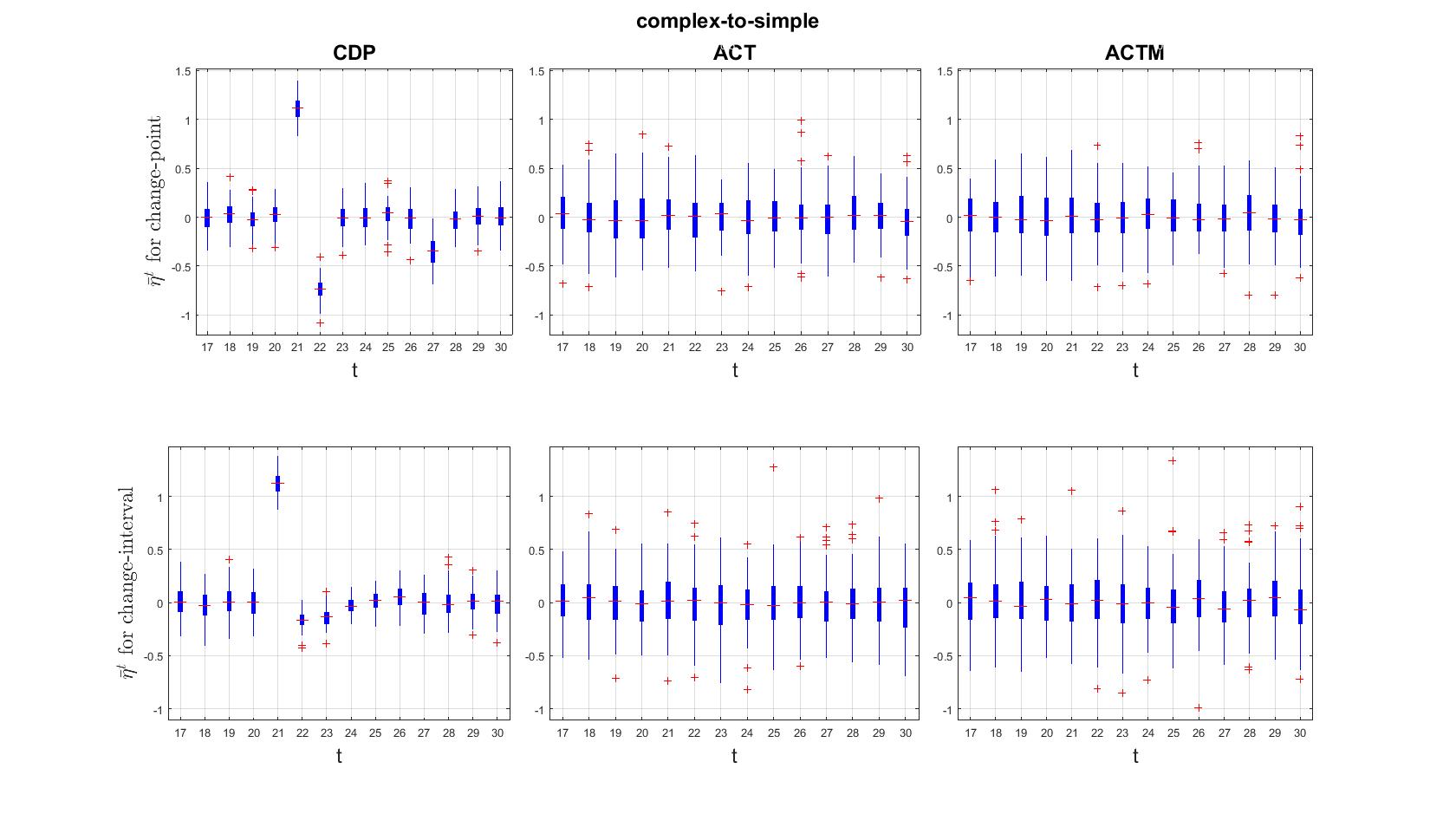}
	\caption{\textbf{Observing $\bar{\pmb{\eta}}^t$ on CDP (left), ACT (middle), and ACTM (right) over time on complex-to-simple for $w=5$ change point(top) and change-interval (bottom)}. We observe a clear detection at ${\bar{\pmb{\eta}}}^{21}_{CDP}$ for both change-point and change-interval. A detection is not observed on both ${\bar{\pmb{\eta}}}^{21}_{ACT}$ and ${\bar{\pmb{\eta}}}^{21}_{ACTM}$.}
	\label{fig:complex2simpleOddsRatio}
	
\end{figure}

Note that the graphs generated at each time instant are independent samples from a given generative model ($F_0$ or $F_1$). Thus, within the same generative model, edge weights can change from one time instant to another, also causing the connectivity patterns of vertices to change. For example, in Figure \ref{fig:grpOdds} (Top), we observe that the ${\pmb{\eta}}^{t}$'s are centred at a positive level even within the generative model, $\mathcal{M}1$. This shows that the set of vertices, $V_c$, for the group-change scenario (Table \ref{tab:scenarios}) undergo a higher change in their connectivity patterns for independent graph realizations under $\mathcal{M}1$. However, when calculating our performance measure, the set of vertices, $V_c$, does not necessarily contain those vertices whose connectivity patterns have changed between independent realizations from a given generative model. For example, in Figure \ref{fig:grpOdds} (Bottom), we observe that  ${\pmb{\eta}}^{t}$'s are centred at a negative level within generative model, $\mathcal{M}4$. This shows that the vertices in the set, $V_{\bar{c}}$, are the ones that change more during independent graph realizations under $\mathcal{M}4$. Despite these changes occurring in connectivity patterns within a given generative model, our interest lies in detecting a change during model transitions. At $t=21$, we expect ${\pmb{\eta}}^{21}$ to be larger than the ${\pmb{\eta}}^{t}$'s observed for time instants corresponding to the same model.  Thus, in order to clearly observe this, we calculate the performance measure $\bar{\pmb{\eta}}^t$ (Equation \ref{eq:perf}). In Figure \ref{fig:grpOddsRatio}, we plot $\bar{\pmb{\eta}}^t$ for CDP, ACT, and ACTM  with $w=5$ for the group-change scenario for $t=17,18,\ldots,30$. We observe that $\bar{\pmb{\eta}}^t$ provides a clearer picture than ${\pmb{\eta}}^t$ on a method's ability to detect change caused by model transitions. For the rest of the scenarios, we only plot $\bar{\pmb{\eta}}^t$ over time and compare the performance measure, $\pmb{\eta}^{t}$ (Equation \ref{eq:odds}), for CDP, ACT, and ACTM for all window sizes only at the time instant corresponding to a change, i.e., we only compare $\pmb{\eta}^{21}$.
We compare the $\pmb{\eta}^{21}$'s for only those change scores obtained on change-point scenarios since it is sufficient to calculate $\pmb{\eta}^{21}$ for either change-point or change-interval as both involve similar changes when considering only $t=21$. If $\pmb{\eta}^{21}$ is positive, then this indicates that the vertices in the set, $V_c$, have higher change scores compared to the rest of the vertices in $V_{\bar{c}}$, at the time instant of change ($t=21$). Figure \ref{fig:grpOddsWindow} shows $\pmb{\eta}^{21}$ returned by CDP, ACT, and ACTM for the group-change scenario using various window sizes, $w=1,5,10$. The $\pmb{\eta}^{21}$ returned by CDP for all window sizes are clearly positive. For $\pmb{\eta}^{21}$ returned by ACT and ACTM, we see the bulk of the interval lying below zero for all window sizes, showing failure in detection for those methods. 
%\begin{figure}[]
%	\centering
%	\includegraphics[trim = 50mm 0mm 0mm 0mm, scale=0.22]{Figures/groupchange-logOddsWindow}
%	\caption{\textbf{Plot of $\pmb{\eta}^{21}$ on CDP, ACT, and ACTM for group-change for $w=1,5,10$ at change-point}. For all $w$, ${\pmb{\eta}^{21}}_{CDP}$ is positive and increases with $w$. For all $w$, a majority of elements of ${\pmb{\eta}^{21}}_{ACT}$ and ${\pmb{\eta}^{21}}_{ACTM}$ are negative.}
%	\label{fig:grpOddsWindow}
%\end{figure}
We also observe $\pmb{\eta}^{21}$ for the other change scenarios, split (Figure \ref{fig:splitOddsWindow}), merge (Figure \ref{fig:mergeOddsWindow}), form (Figure \ref{fig:formOddsWindow}), fragment (Figure \ref{fig:fragmentOddsWindow}), hetero-to-homo (Figure \ref{fig:hetro2homoOddsWindow}), homo-to-hetero (Figure \ref{fig:homo2heteroOddsWindow}), simple-to-complex (Figure \ref{fig:simple2complexOddsWindow}), and complex-to-simple (Figure \ref{fig:complex2simpleOddsWindow}). Our results show that CDP successfully detects the change in all the scenarios considered. ACT shows failure in detection for all change scenarios except form and fragment, while ACTM shows failure in detection for all change scenarios except form.

We further observe $\bar{\pmb{\eta}}^{t}$ for split (Figure \ref{fig:splitOddsRatio}), merge (Figure \ref{fig:mergeOddsRatio}), form (Figure \ref{fig:formOddsRatio}), fragment (Figure \ref{fig:fragmentOddsRatio}), hetero-to-homo (Figure \ref{fig:hetro2homoOddsRatio}), homo-to-hetero (Figure \ref{fig:homo2heteroOddsRatio}), simple-to-complex (Figure \ref{fig:simple2complexOddsRatio}), and complex-to-simple (Figure \ref{fig:complex2simpleOddsRatio}). CDP shows a clear detection at $t=21$ for change scenarios form, fragment, hetero-to-homo, homo-to-hetero, simple-to-complex, and complex-to-simple. In the case of split and merge, we observe an increase at $\bar{\pmb{\eta}}^{21}$, with the intervals being wide. ACT and ACTM do not show a clear increase at $\bar{\pmb{\eta}}^{21}$ for split, merge, simple-to-complex, and complex-to-simple cases. For homo-to-hetero and hetero-to-homo, we observe a slight increase in $\bar{\pmb{\eta}}^{21}$ for ACT and ACTM. For fragment, $\bar{\pmb{\eta}}^{21}$ is highly negative for both ACT and ACTM methods. Thus, although we observed in Figure \ref{fig:fragmentOddsWindow} that ACT shows good performance in terms of $\pmb{\eta}^{21}$, Figure \ref{fig:fragmentOddsRatio} shows that the change scores have decreased at $t=21$. Thus, ACT shows failure in detecting the fragment scenario.  

In Table \ref{tab:signTestResults}, we perform the sign test to assess the statistical significance of the observed results. We compare $\pmb{\eta}^{21}$ calculated for group-change, split, merge, form, hetero-to-homo, homo-to-hetero, simple-to-complex, and complex-to-simple at $w=5$ for change-point. We do not perform the sign test for fragment scenario as we already observed a decrease in $\bar{\pmb{\eta}}^{21}$ for ACT and ACTM compared to previous time instants in Figure \ref{fig:fragmentOddsRatio} (this clearly shows how CDP outperforms these two methods). The leftmost column in Table \ref{tab:signTestResults} gives the alternative hypothesis tested. Subsequently in Table \ref{tab:propResults}, we show the proportion of values in $\pmb{\eta}^{21}$, that correspond to the hypothesis tested in Table \ref{tab:signTestResults}. CDP outperforms ACT and ACTM for all change scenarios except form. ACTM outperforms ACT for group-change, form, and homo-to-hetero. For the other scenarios tested, there is no difference in $\pmb{\eta}^{21}$ for ACT and ACTM. However, when we consider the proportions in Table \ref{tab:propResults}, the majority of the entries in ${\bar{\pmb{\eta}}}^{21}_{ACTM}$ are greater than ${\bar{\pmb{\eta}}}^{21}_{ACT}$ for all change scenarios except hetero-to-homo. 

\begin{landscape}
\begin{table}[!]
%\begin{sidewaystable}[]
	\centering
	\caption{{Sign test results for comparing $\pmb{\eta}^{21}$ calculated using CDP, ACT, and ACTM at $w=5$ for change-point}. }
	\label{tab:signTestResults}
	\begin{tabular}{|c|c|c|c|c|c|c|c|c|}
		\toprule
		Alternative Hypothesis & group- & split & merge & form & homo-to- & hetero- & simple-to & complex- \\
		& change &  &  &  & hetero & to-homo & -complex & to-simple \\
		\midrule
		${\pmb{\eta}}^{21}_{CDP} > {\pmb{\eta}}^{21}_{ACT}$ & 2.08e-23 & 2.08e-23 & 2.08e-23 & 1.0000 & 2.08e-23 & 2.08e-23 & 2.08e-23 & 2.08e-23\\
		${\pmb{\eta}}^{21}_{CDP} < {\pmb{\eta}}^{21}_{ACT}$ & 1.0000 & 1.0000 & 1.0000 & 2.08e-23 & 1.0000 & 1.0000 & 1.0000 & 1.0000\\
		${\pmb{\eta}}^{21}_{CDP} \neq {\pmb{\eta}}^{21}_{ACT}$ & 4.16e-23 & 4.16e-23 & 4.16e-23 & 4.16e-23 & 4.16e-23 & 4.16e-23 & 4.16e-23 & 4.16e-23\\
		\midrule
		${\pmb{\eta}}^{21}_{CDP} > {\pmb{\eta}}^{21}_{ACTM}$ & 2.08e-23 & 2.08e-23 & 2.08e-23 & 1.0000 & 2.08e-23 & 2.08e-23 & 2.08e-23 & 2.08e-23\\
		${\pmb{\eta}}^{21}_{CDP} <{\pmb{\eta}}^{21}_{ACTM}$ & 1.0000 & 1.0000 & 1.0000 & 2.08e-23 & 1.0000 & 1.0000 & 1.0000 & 1.0000\\
		${\pmb{\eta}}^{21}_{CDP} \neq {\pmb{\eta}}^{21}_{ACTM}$ & 4.16e-23 & 4.16e-23 & 4.16e-23 & 4.16e-23 & 4.16e-23 & 4.16e-23 & 4.16e-23 & 4.16e-23\\
		\midrule
		
		${\pmb{\eta}}^{21}_{ACTM} > {\pmb{\eta}}^{21}_{ACT}$ & 4.02e-11 & 0.3085 & 0.3821 & 2.08e-23 & 0.00347 & 0.6915 & 0.2420 & 0.0968\\
		${\pmb{\eta}}^{21}_{ACTM} < {\pmb{\eta}}^{21}_{ACT}$ & 1.0000 & 0.7580 & 0.6915 & 1.0000 & 0.9981 & 0.3821 & 0.8159 & 0.9332\\
		${\pmb{\eta}}^{21}_{ACTM} \neq {\pmb{\eta}}^{21}_{ACT}$ & 8.03e-11 & 0.6171 & 0.7642 & 4.16e-23 & 0.0069 & 0.7642 & 0.4840 & 0.1936\\
		
		\bottomrule
	\end{tabular}
\end{table}
%\end{sidewaystable}
\end{landscape}

\setlength\tabcolsep{1pt}
\begin{table}[!]
	\centering
	\caption{Proportions for comparing $\pmb{\eta}^{21}$ calculated using CDP, ACT, and ACTM at $w=5$ for change-point}
	\label{tab:propResults}
	\begin{tabular}{|c|c|c|c|c|c|c|c|c|}
		\toprule
		Categories & group- & split & merge & form & homo-to- & hetero- & simple-to & complex- \\
		& change &  &  &  & hetero & to-homo & -complex & to-simple \\
		\midrule
		${\pmb{\eta}}^{21}_{CDP} > {\pmb{\eta}}^{21}_{ACT}$  & 1.00 & 1.00 & 1.00 & 0.00 & 1.00 & 1.00 & 1.00 & 1.00\\
		${\pmb{\eta}}^{21}_{CDP} < {\pmb{\eta}}^{21}_{ACT}$  & 0.00 & 0.00 & 0.00 & 1.00 & 0.00 & 0.00 & 0.00 & 0.00\\
		${\pmb{\eta}}^{21}_{CDP} = {\pmb{\eta}}^{21}_{ACT}$  & 0.00 & 0.00 & 0.00 & 0.00 & 0.00 & 0.00 & 0.00 & 0.00\\
		\midrule
		${\pmb{\eta}}^{21}_{CDP} > {\pmb{\eta}}^{21}_{ACTM}$  & 1.00 & 1.00 & 1.00 & 0.00 & 1.00 & 1.00 & 1.00 & 1.00\\
		${\pmb{\eta}}^{21}_{CDP} <{\pmb{\eta}}^{21}_{ACTM}$  & 0.00 & 0.00 & 0.00 & 1.00 & 0.00 & 0.00 & 0.00 & 0.00\\
		${\pmb{\eta}}^{21}_{CDP} = {\pmb{\eta}}^{21}_{ACTM}$  & 0.00 & 0.00 & 0.00 & 0.00 & 0.00 & 0.00 & 0.00 & 0.00\\
		\midrule
		
		${\pmb{\eta}}^{21}_{ACTM} > {\pmb{\eta}}^{21}_{ACT}$  & 0.83 & 0.53 & 0.52 & 1.00 & 0.64 & 0.48 & 0.54 & 0.57\\
		${\pmb{\eta}}^{21}_{ACTM} < {\pmb{\eta}}^{21}_{ACT}$  & 0.17 & 0.47 & 0.48 & 0.00 & 0.36 & 0.52 & 0.46 & 0.43\\
		${\pmb{\eta}}^{21}_{ACTM} = {\pmb{\eta}}^{21}_{ACT}$  & 0.00 & 0.00 & 0.00 & 0.00 & 0.00 & 0.00 & 0.00 & 0.00\\
		
		\bottomrule
	\end{tabular}
\end{table}

Each change scenario discussed in Section \ref{Simulation Experiments}, consists of a change in the behaviour of a subset of vertices in the DCSBM graph: (i) split, merge, homo-to-hetero, and hetero-to-homo involve the set of vertices, $\{v_1,v_2,\ldots, v_{300}\}$, in the most sparsely connected block in the graph, (ii) group-change, simple-to-complex, and complex-to-simple involve the set of vertices, $\{v_{1},v_{2},\ldots, v_{600}\}$, that can be considered to be moderately connected in the graph, and (iii) form and fragment involve the set of vertices, $\{v_{601},v_{602},\ldots, v_{900}\}$, from the most dense block in the graph. From the results of the experiments conducted, we observe how CDP detects changes involving  each of these subsets of vertices. ACT and ACTM could only detect the form scenario that involves a change in the vertices from the most dense block in the graph. However still, ACT and ACTM failed to detect fragment, which is the reverse of form. From the overall simulation results we see that, while ACTM is slightly better than ACT, CDP is the best of the three.

\begin{figure}[]
	\centering
	\includegraphics[trim = 60mm 0mm 0mm 0mm, scale=0.15]{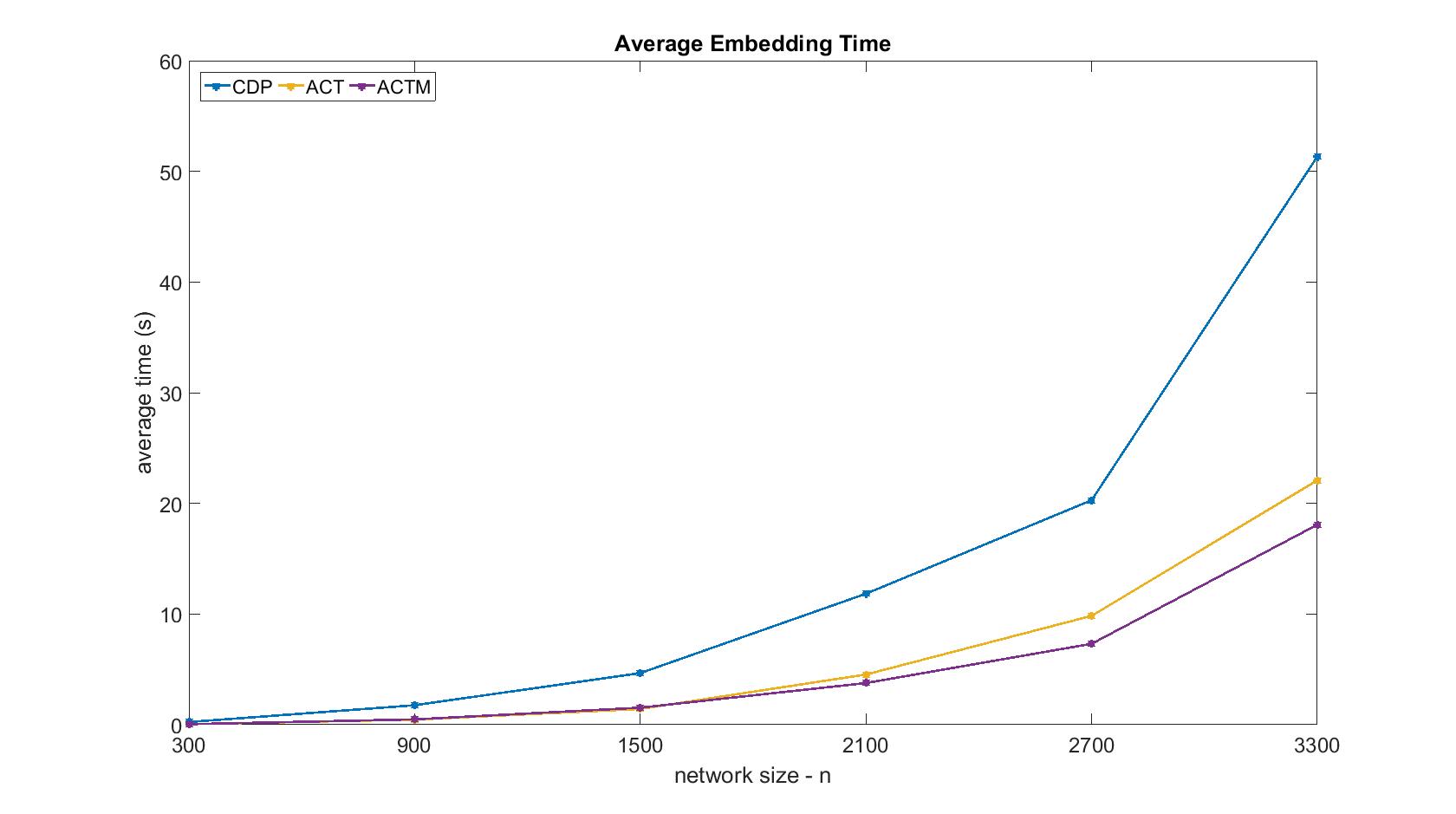}
	\caption{Comparison of average CPU time taken to embed a graph using different methods.}
	\label{fig:TimeEmbed}
\end{figure}

\begin{figure}[]
	\centering
	\includegraphics[trim = 60mm 0mm 0mm 0mm, scale=0.15]{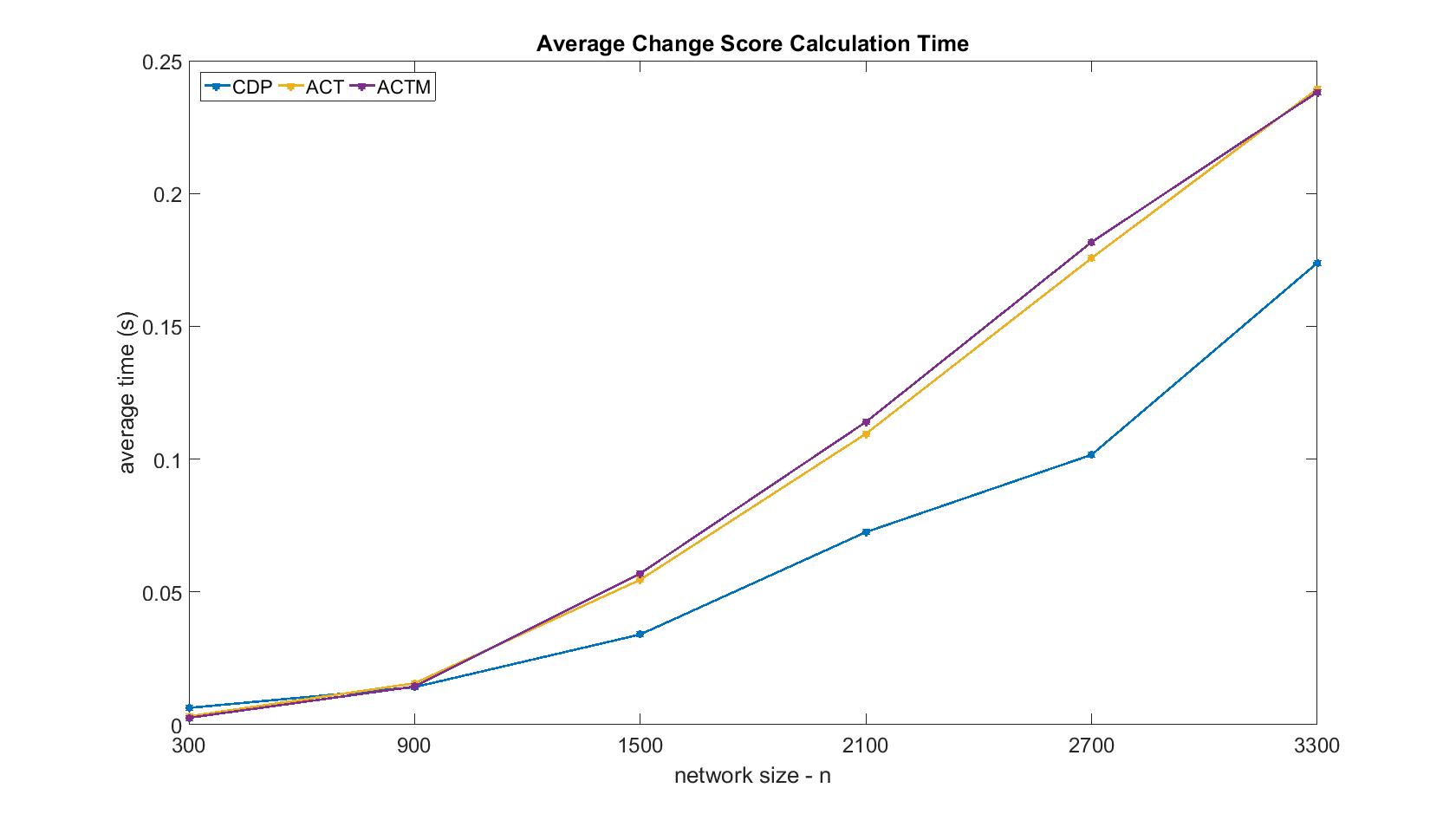}
	\caption{Comparison of average CPU time taken to calculate change scores using different methods.}
	\label{fig:timeCalc}
\end{figure}

Next, we conduct experiments to evaluate the scalability of CDP and the other two baseline methods. We select one change scenario, and generate a sequence of six graphs. At the sixth time instant, we calculate vertex change scores using a window of size five. We repeat this over $100$ simulation runs. We calculate the average CPU time taken to embed a single graph, and the average CPU time taken to calculate profile behaviour and change scores at a given time instant. Following the same procedure, we conduct experiments on graphs with several sizes. All experiments are implemented on a Windows server Intel Xeon with two 3.3GHz processors of 128 GB RAM. Our results are given in Table \ref{tab:scalability2}. Figures \ref{fig:TimeEmbed} and \ref{fig:timeCalc} plot the average computational time taken by each method for the embedding step and change score calculation step, respectively for different graph sizes. 

\begin{table}[]
	\centering
	\caption{Average CPU  time (seconds) taken by each method}
	\label{tab:scalability2}
	\begin{tabular}{|r|r|r|r|r|}
		\toprule
		Task                     & n    & CDP & ACT & ACTM \\
		\midrule
		Spectral embedding       & 300  & 0.2425 & 0.0474 & 0.0475 \\
		& 900  & 1.7584 & 0.4554  &  0.4832 \\
		& 1500 & 4.6635 & 1.4223  & 1.5394  \\
		& 2100 & 11.8364 & 4.5364  & 3.7637  \\
		& 2700 & 20.2857 & 9.8368  & 7.3115   \\
		& 3300 & 51.3395 & 22.0778  & 18.0710 \\
		
		\midrule 
		Profile behaviour and change score calculation & 300  & 0.0063 & 0.0032  & 0.0026     \\
		& 900  &  0.0142  & 0.0156    & 0.0144 \\
		& 1500 &  0.0340  & 0.0546    & 0.0570  \\
		& 2100 &  0.0725  & 0.1095    & 0.1140  \\
		& 2700 &  0.1016  & 0.1757    & 0.1817  \\
		& 3300 & 0.1739 & 0.2394  & 0.2382  \\
		\bottomrule  
	\end{tabular}
\end{table}

The most computationally efficient methods for embedding a graph are the ACT and ACTM methods. These methods both perform SVD of the weighted adjacency matrix to extract a single singular vector. However, as shown in our previous results, keeping only one singular vector does not provide a good representation of all vertices in the network. CDP performs SVD to extract $d$ singular vectors from the representation matrix to obtain an optimal representation, hence taking more computational time. When comparing change score calculation times (Figure \ref{fig:timeCalc}), CDP is clearly more efficient than ACT or ACTM for graph sizes greater than $900$. However, we observe that the change score times are negligible compared to the respective embedding times. 
From the overall simulation results, it can be concluded that CDP method outperforms the other two baseline methods, and is the most reliable method in detecting the different types of change scenarios considered.


%\begin{figure}
%	\centering
%	\begin{subfigure}{.5\textwidth}
%		\centering
%		\includegraphics[trim = 60mm 0mm 0mm 0mm, scale=0.25]{Figures/TimeEmbed}
%		\caption{Comparison of average CPU time taken to embed a graph using different methods.}
%		\label{fig:TimeEmbed}	
%	\end{subfigure}%
%	\begin{subfigure}{.5\textwidth}
%		\centering
%			\includegraphics[trim = 60mm 0mm 0mm 0mm, scale=0.25]{Figures/timeCalc}
%		\caption{Comparison of average CPU time taken to calculate change scores using different methods.}
%		\label{fig:timeCalc}
%	\end{subfigure}
%	\caption{\textbf{Illustration of scalability of CDP compared to ACT and ACTM.}
%	\label{fig:scalability}
%\end{figure}  

\section{Case Study: The Enron E-mail Network}
\label{Real World Data}

The Enron dataset is used in various publications for community detection and anomaly detection \cite{priebe2005scan,peel2015detecting,rossi2013modeling}. In this paper, we use the cleaned and processed version of the dataset created by \cite{tang2008community}. Based on the sent and received email addresses in the original Enron corpus, \cite{tang2008community} extract a total of $2359$ user email addresses, and construct a time sequence of email-sender networks and a time sequence of email-receivers networks for each month from  December 1999 to March 2002. Based on these two dynamic networks, we construct a time sequence of 28 undirected graphs (one for each month),  where the vertices denote user email addresses, and the edge weights denote the number of emails communicated (either sent or received) by the corresponding pair of users. Each graph is then represented by an $n\times n$ symmetric weighted adjacency matrix, where $n=2359$. 
%We disregard self loops as an email from a person to his own self is not considered important. 

We applied CDP, ACT, and ACTM on this data with a window of length $1$ (we used $w=1$ as the time instants correspond to months). Each of the methods, CDP, ACT, and ACTM returns an $n \times 1$ vector of vertex change scores for each time instant. Our goal is to detect vertices which have changed most during a given time instant. To achieve this goal, we employ the following simple procedure: Let $\mathbf{z}^t$ be the vector of vertex change scores obtained for a given method. Each $z_i^t$ is converted to a z-score, $\hat{z}_i^t=\frac{(z_i^t-\bar{z}^t)}{\sigma_{\mathbf{z}^t}}$, where $\bar{z}^t=\frac{1}{n}\sum_{i=1}^{n}z_i^t$, and $\sigma^2_{\mathbf{z}^t}=\frac{1}{n-1}\sum_{i=1}^{n}(z_i^t-\bar{z}^t)^2$. We threshold each z-score distribution, $\hat{\mathbf{z}}^t$, at $5$ to detect the vertices which changed the most at that time instant (we investigate and find that the threshold $5$ for this dataset ensures the percentage of vertices detected at each time instant is less than two percent for all three methods). The Enron time-line contains a detailed description of the key players, and events that took place during the rise and fall of the Enron company \cite{thomas2002rise}. Based on the assumption that the email communication patterns within the company were affected by the events associated with the scandal, we evaluate the performance of our change detection method.

\begin{figure}[] 
	\centering
	\includegraphics[trim = 60mm 0mm 0mm 0mm, scale=0.15]{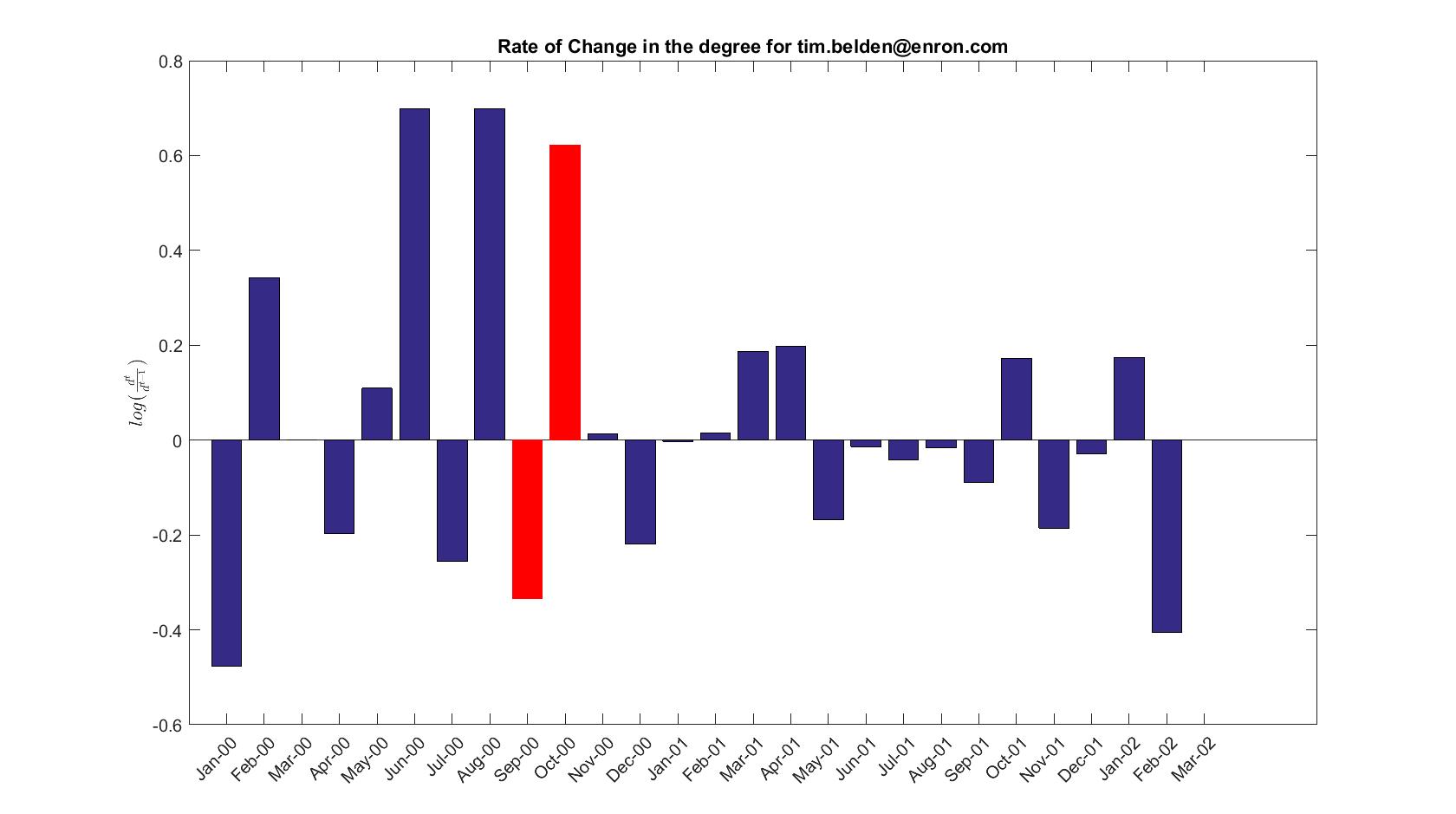}
	\caption{Rate of change in the number of emails sent and received by Timothy Beldon between consecutive months\label{fig:enron_timothy}}
\end{figure}

\begin{figure}[] 
	\centering
	\includegraphics[trim = 60mm 0mm 0mm 0mm, scale=0.15]{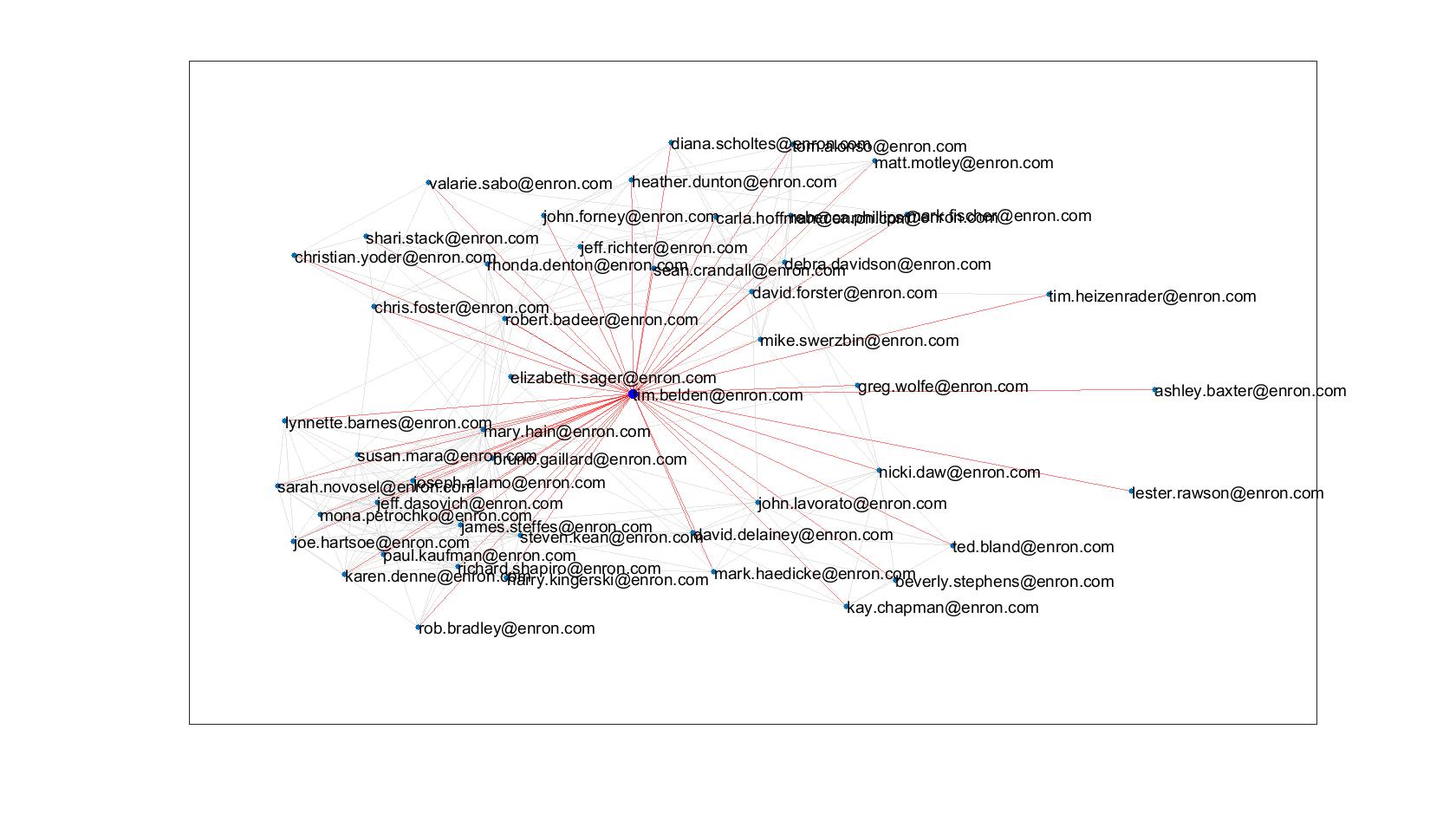}
	\caption{\textbf{Subgraph of the vertex corresponding to Timothy Beldon for August-2000}. Beldon is represented by the enlarged blue vertex in the centre, and the edges connected to it are highlighted in red.}
	\label{fig:beldon_9}
\end{figure}

\begin{figure}[] 
	\centering
	\includegraphics[trim = 60mm 00mm 0mm 0mm, scale=0.15]{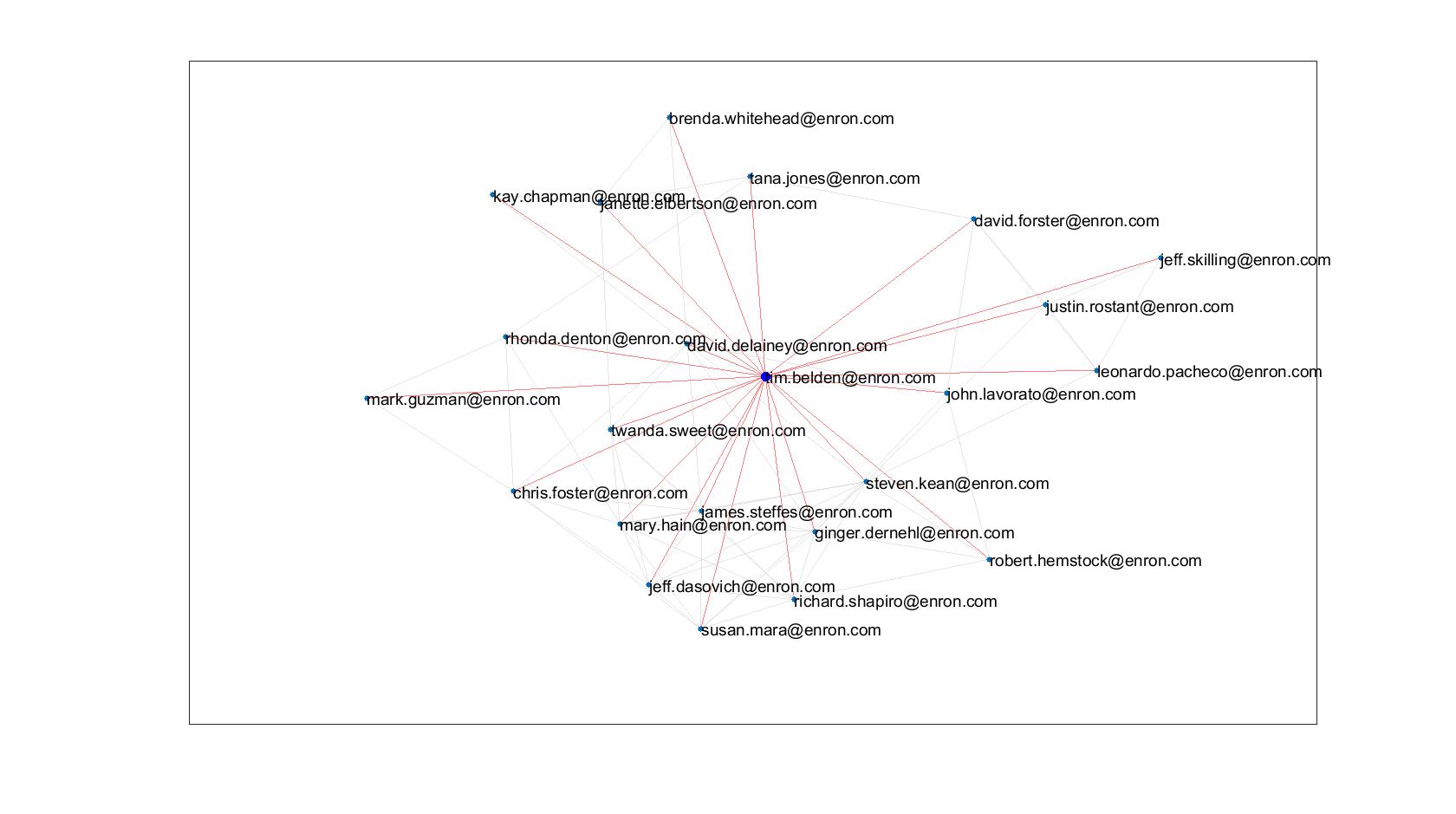}
	\caption{\textbf{Subgraph of the vertex corresponding to Timothy Beldon for September-2000}. Beldon is represented by the enlarged blue vertex in the centre, and the edges connected to it are highlighted in red.}
	\label{fig:beldon_10}
\end{figure}

\begin{figure}[] 
	\centering
	\includegraphics[trim = 60mm 00mm 0mm 0mm, scale=0.15]{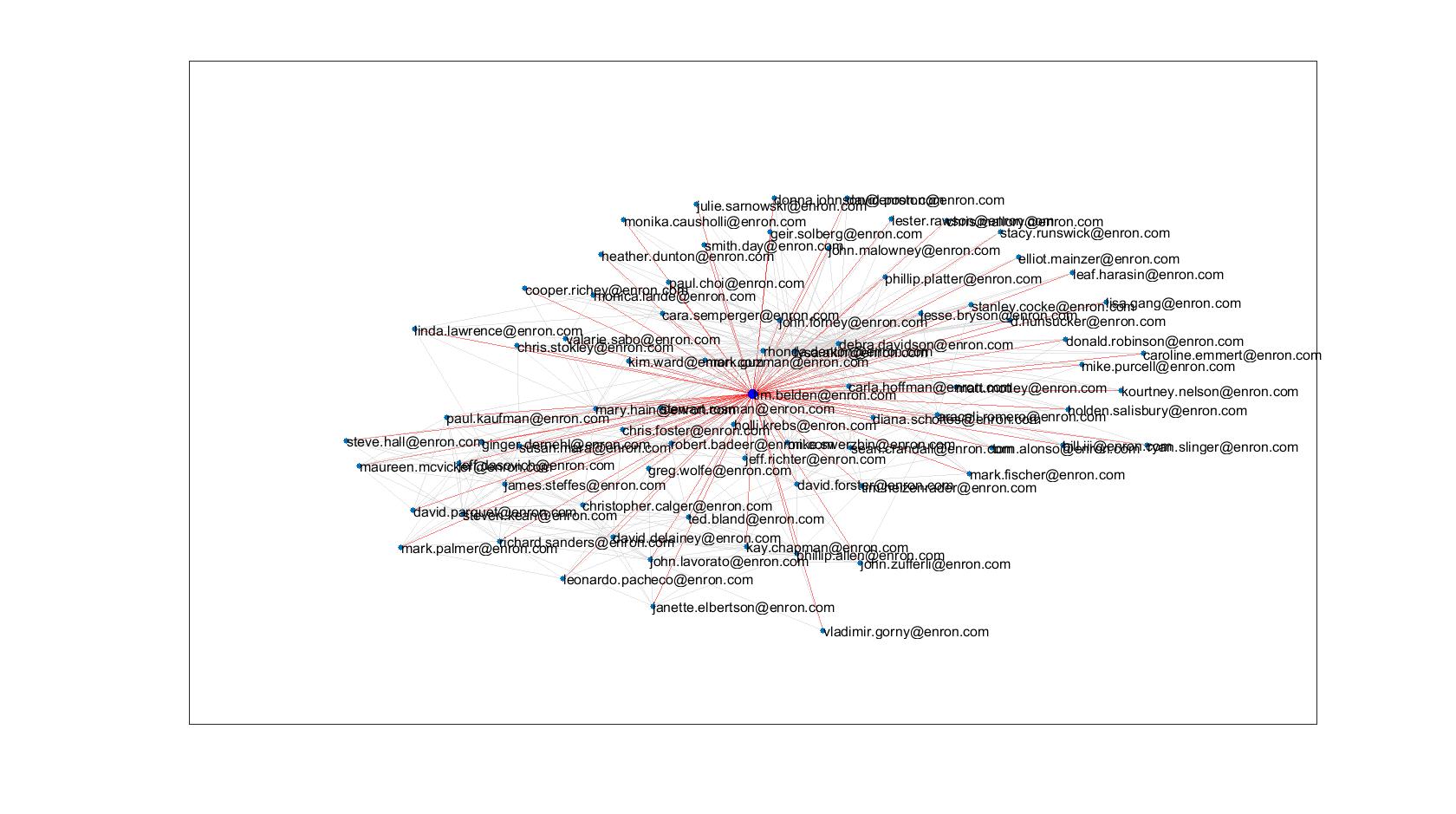}
	\caption{\textbf{Subgraph  of the vertex corresponding to Timothy Beldon for October-2000}. Beldon is represented by the enlarged blue vertex in the centre, and the edges connected to it are highlighted in red.}
	\label{fig:beldon_11}
\end{figure}

We find Timothy Beldon (chief trader of Enron's West Coast power desk and convicted of wire fraud) to be one of the entities which changed the most, using CDP for the time instants corresponding to September-2000 and October-2000. According to the Enron time-line, this is the time when an attorney from Enron travelled to Portland to discuss Timothy Beldon's strategies of boosting energy prices. Figure \ref{fig:enron_timothy} shows the rate of change in the number of emails sent and received by Beldon throughout the whole time period considered. First, in September-2000, we observe a noticeable drop in the number of emails communicated, which then suddenly increases in October-2000. Figures \ref{fig:beldon_9}, \ref{fig:beldon_10}, and \ref{fig:beldon_11} further show the subgraphs consisting of the vertices corresponding to Timothy Beldon and his connections for time instants, August-2000, September-2000 and October-2000 respectively. It is observed that Beldon, who communicated with many employees in various job roles in August-2000, limited his communications mostly to the top level executives and CEO's during September-2000. He then starts communicating with many employees in different job roles in October-2000. CDP successfully detects this change in degree as well as community membership by giving high change scores to Timothy Beldon during the respective time instants. 

\begin{figure}[] 
	\centering
	\includegraphics[trim = 60mm 0mm 0mm 0mm, scale=0.15]{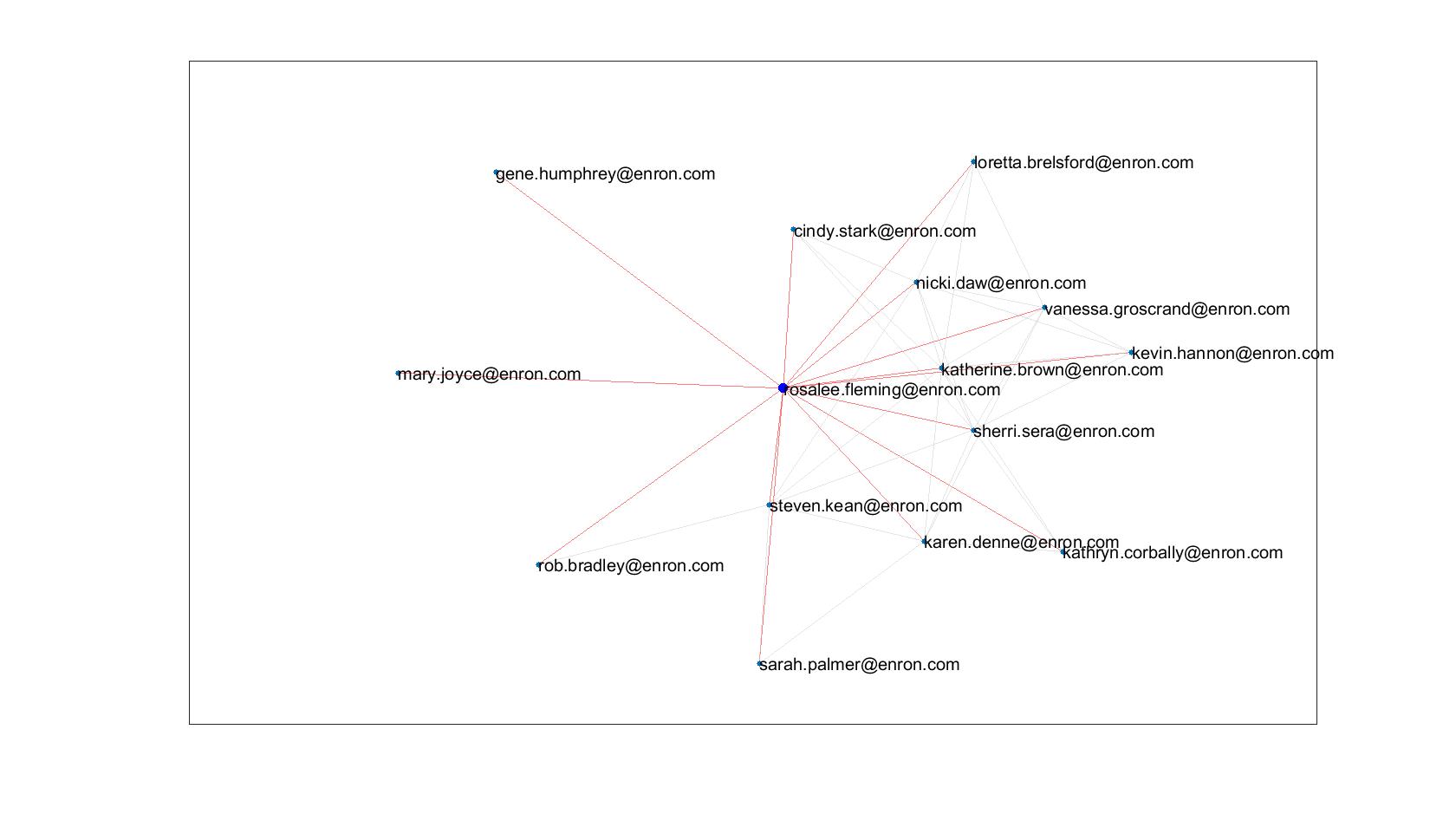}
	\caption{\textbf{Subgraph of the vertex corresponding to Rosalie Fleming for November-2000}. Fleming is represented by the enlarged blue vertex in the centre, and the edges connected to it are highlighted in red.}
	\label{fig:fleming_12}
\end{figure}

\begin{figure}[] 
	\centering
	\includegraphics[trim = 60mm 00mm 0mm 0mm, scale=0.15]{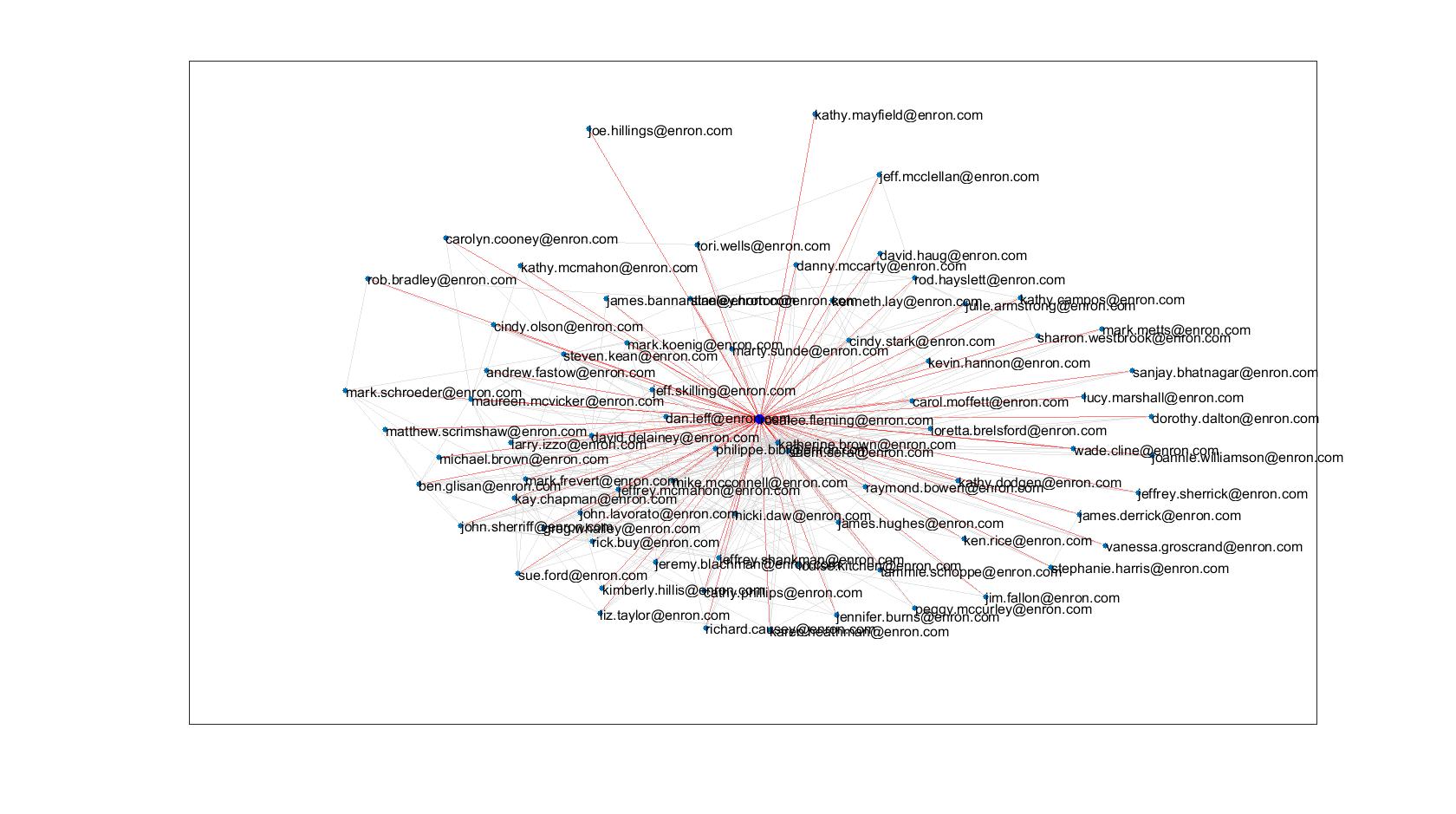}
	\caption{\textbf{Subgraph of the vertex corresponding to Rosalie Fleming for December-2000}. Fleming is represented by the enlarged blue vertex in the centre, and the edges connected to it are highlighted in red.}
	\label{fig:fleming_13}
\end{figure}

Enron announced that current CEO Kenneth Lay was to be replaced by Jeffry Skilling in December-2000. CDP successfully detects Rosalie Fleming, who was the assistant of Kenneth Lay, as one of the top changed entities for this time instant. The subgraph in Figure \ref{fig:fleming_12} show Fleming's connections in November-2000 which are mostly with the employees in the company. Figure \ref{fig:fleming_13} shows how she starts communicating with people with different job roles such as CEO's, directors, etc. in December-2000. We observe how Fleming's connectivity patterns drastically change during this transition. CDP successfully detects this change at the corresponding time instant.

During the time instants between October-2001 to February-2002, CDP gives high change scores to employees in job roles such as risk analysts, senior specialists, presidents, vice presidents, and traders. This is justifiable as this was the time period where Enron's stocks started to fall, and bankruptcy was declared. 

\begin{figure}[] 
	\centering
	\includegraphics[trim = 60mm 0mm 0mm 0mm, scale=0.15]{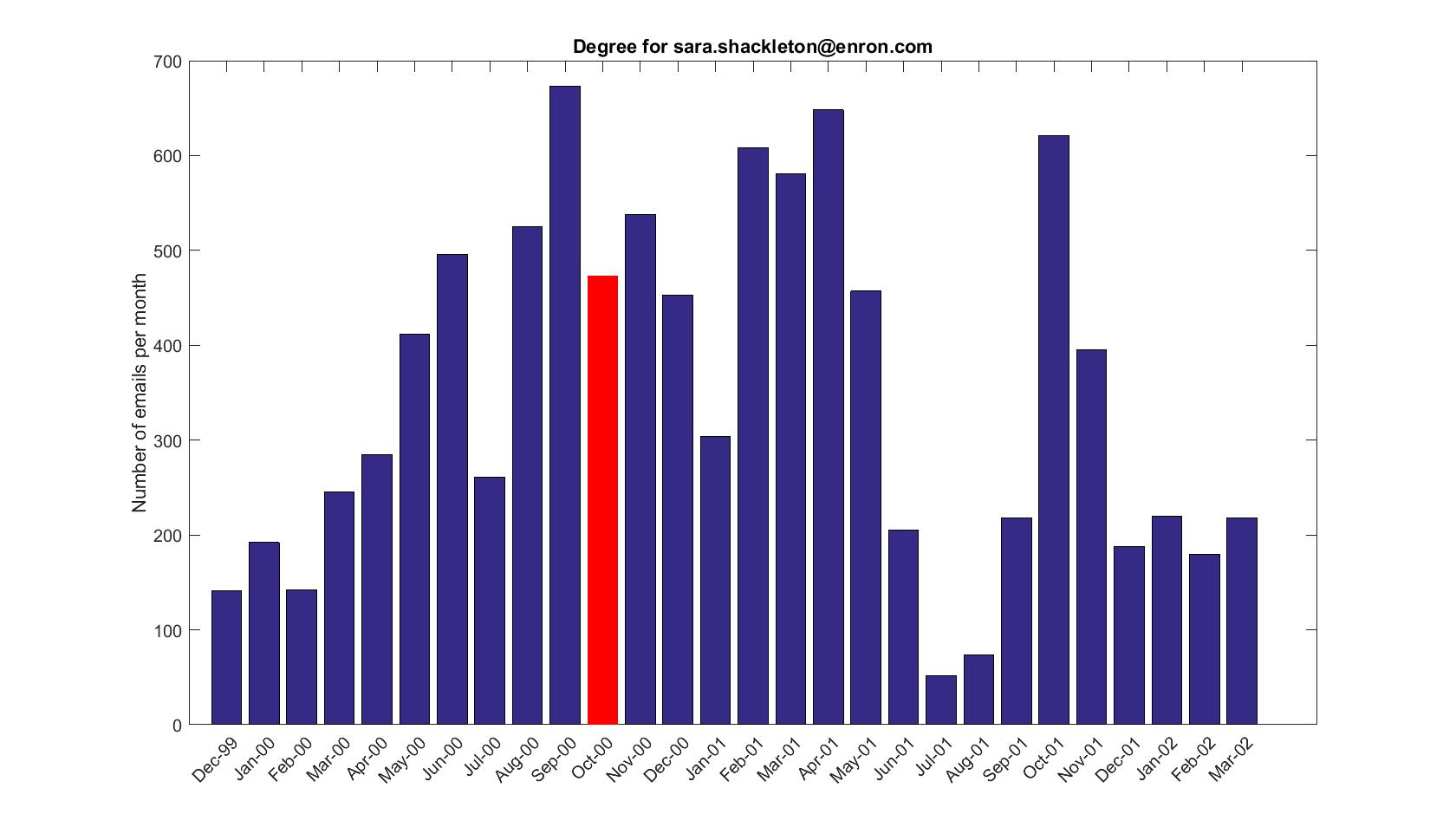}
	\caption{Number of emails sent and received by Sarah Shackleton during the 28 months\label{fig:enron_sarah}}
\end{figure}

For the transition from September-2000 to October-2000, both ACT and ACTM give high change scores to Sara Shackleton (previous vice president to Enron North America). Figure \ref{fig:enron_sarah} shows the histogram of the emails sent and received by Shackleton throughout this time period considered. We observe how Shackleton maintains a large number of overall communications, hence acts as a hub in the network. We observe a decrease in the degree at the time instant denoting October-2000 with respect to September-2000. Furthermore, our investigations show that there is approximately a $66 \%$ overlap in Shackleton's connections for these two time instants. Hence it is justifiable to assume that the change detected by ACT and ACTM simply  reflects the change in degree for the corresponding vertex. 

\begin{figure}[] 
	\centering
	\includegraphics[trim = 60mm 0mm 0mm 0mm, scale=0.15]{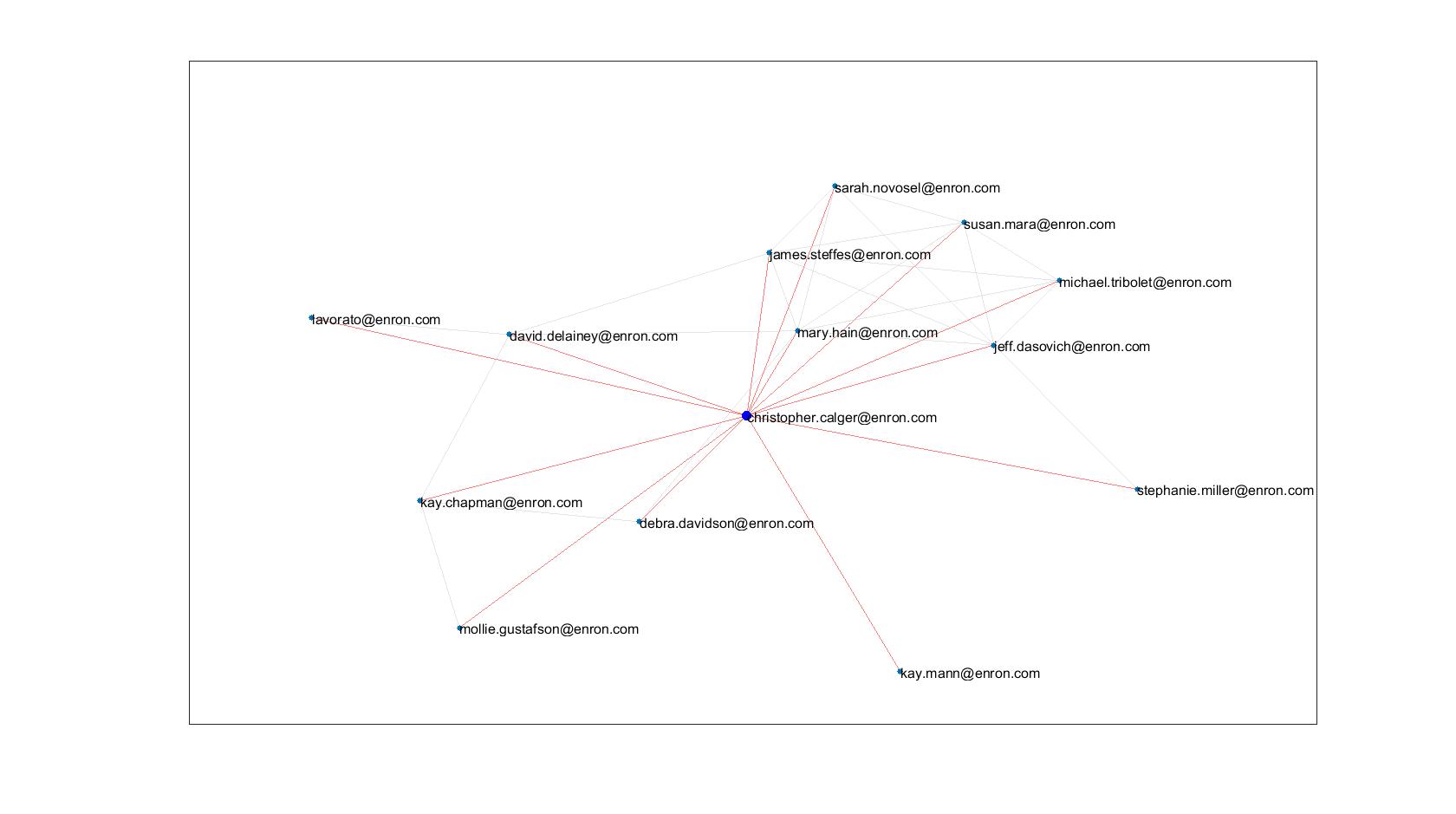}
	\caption{\textbf{Subgraph of the vertex corresponding to Christopher Calger for January-2001}. Calger is represented by the enlarged blue vertex in the centre and the edges connected to it are highlighted in red.}
	\label{fig:calger_14}
\end{figure}

\begin{figure}[] 
	\centering
	\includegraphics[trim = 60mm 00mm 0mm 0mm, scale=0.15]{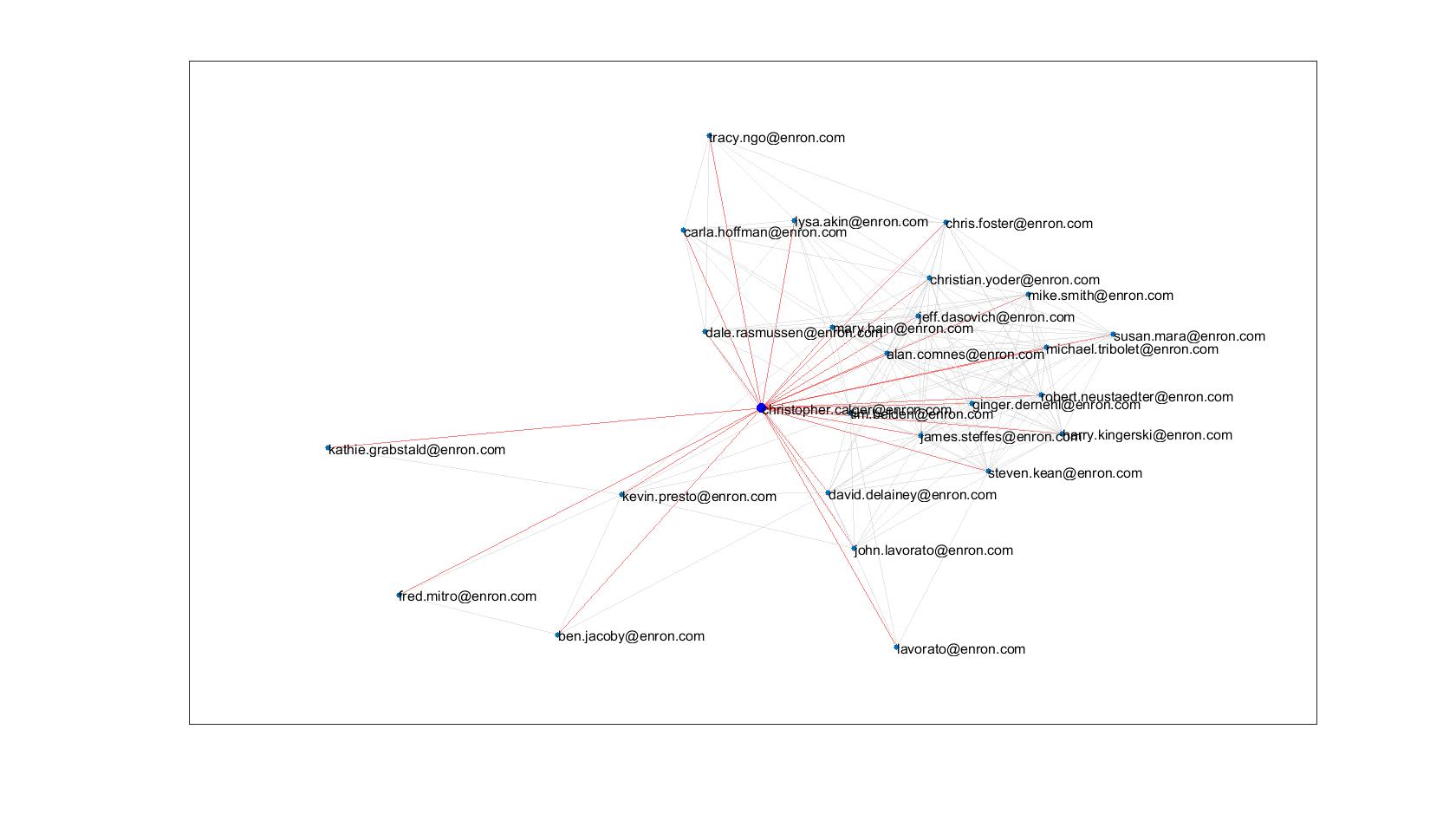}
	\caption{\textbf{Subgraph of the vertex corresponding to Christopher Calger for February-2001}. Calger is represented by the enlarged blue vertex in the centre and the edges connected to it are highlighted in red.}
	\label{fig:calger_15}
\end{figure}

\begin{figure}[] 
	\centering
	\includegraphics[trim = 60mm 0mm 0mm 0mm, scale=0.15]{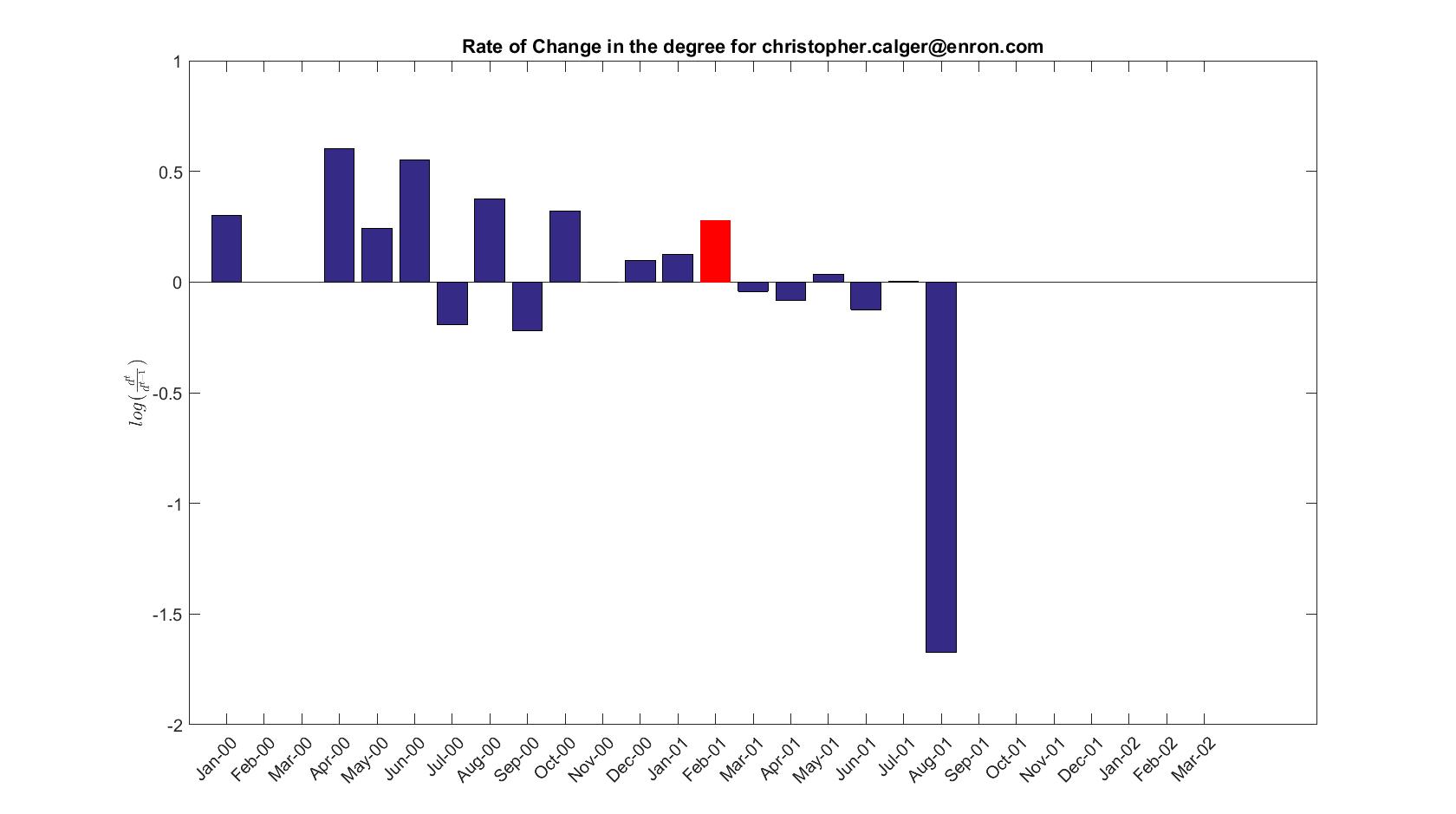}
	\caption{Rate of change in the number of emails sent and received by Christopher Calger during the 28 months\label{fig:calger_degree}}
\end{figure}

Christopher Calger (former executive in Enron's trading business) is also detected by ACT and ACTM for the transition from January-2001 to February-2001. However, when comparing the subgraph around the vertex corresponding to Calger in January-2001 (Figure \ref{fig:calger_14}) and February-2001 (Figure \ref{fig:calger_15}), we do not see a considerable change in his connections. Our further calculations show a $54\%$ overlap in his connections. However, when observing the rate of change in the degree of the same vertex throughout the whole time period (Figure \ref{fig:calger_degree}), February-2001 shows a slight increase. 

In summary, CDP successfully detects some key players involved in the scandal as vertices which change the most during time instants corresponding to suspicious events in the Enron time-line. ACT and ACTM are focused mostly on change in the degree of the vertices, while CDP detects different types of changes. 

\section{Summary}
\label{Summary}

In this paper, we present a novel method, CDP, to detect changes in vertex behaviour in a dynamic network represented as a time sequence of undirected and weighted graphs. We adopt a spectral embedding approach for this purpose. 

Although there already exist change detection methods based on spectral embedding, such as ACT \cite{ide2004eigenspace}, they are mainly designed to detect changes occurring in dense, well-connected graphs. Hence, when applied to sparse and heterogeneous graphs, they focus on the behaviour of highly active, dominant vertices. Changes occurring in vertices with moderate connectivity go unnoticed by these methods. Our approach adapts spectral techniques, commonly used in the area of spectral clustering, to obtain an embedding that represents all vertices in the graph. Our
graph regularization method addresses sparseness and heterogeneity that are common in most real-world graphs. We apply a Procrustes analysis method to the embeddings to calculate change scores for vertices at each time instant. This is a novel application of Procrustes analysis. According to the results of our simulation experiments and experiments on the Enron email dataset, CDP successfully detects various changes involving vertices in a time evolving graph. These changes include changes in vertex degree, changes in community structure and unusual increases or decreases in edge weights. In all experiments, the performance of CDP was compared against two other spectral-based change detection methods, ACT and ACTM, which did not address sparsity and heterogeneity issues in the embedding stage. These baseline methods failed to detect the changes in the majority of experiments.

Several possible future research directions emerged from the work of this paper. Deciding the dimension of the embedding is one of the most critical steps of our method. An optimal low dimension, $d$, ensures that the embedded points amplify the connectivity structure of the graph and remove noise and redundant information. We adapted \cite{achlioptas2007fast}'s low-rank matrix approximation method to determine $d$. From our discussions and experimental results, this method is a good alternative to the traditional scree-plot method in estimating the correct value for $d$ especially in real-world graphs. However, when applied to large graphs, a drawback is the high computational cost associated with this method.
% This is mainly due to the iterative calculation of the spectral norm of a large matrix. 
It would be interesting to investigate other faster methods to estimate $d$ in our algorithms. Several dimensionality selection methods are summarised in \cite{jackson1993stopping} and \cite{jolliffe2002principal} that can be used for this purpose. However, as the truncation dimension, $d$, plays a major role in the accuracy of the change detection algorithm, careful investigation is required to understand the trade-off between accuracy and scalability in the selection of an alternative method. 

CDP use generalized orthogonal Procrustes analysis techniques to calculate a profile embedding from the embeddings inside the window. If the window contains embedded points that are highly variable across time instants, the noise added by these points may prevent a change from being detected. To address this issue, it is possible to use a weighted Procrustes analysis procedure \cite{cootes1992training}. The weights can be selected to give higher importance to those points that are more stable and lesser importance to those points that have high variability inside the window. We believe that it would be worthwhile to investigate whether the use of weighted Procrustes analysis can improve change detection performance.

To conclude, this paper presents a novel change detection method combining spectral embedding and Procrustes analysis techniques. Our method successfully detects a wide range of vertex-based changes that closely relate to changes occurring in most real-world dynamic networks.

\bibliographystyle{abbrvnat}
\bibliography{Bibliography}   % name your BibTeX data base

\begin{thebibliography}{46}
\providecommand{\natexlab}[1]{#1}
\providecommand{\url}[1]{\texttt{#1}}
\expandafter\ifx\csname urlstyle\endcsname\relax
  \providecommand{\doi}[1]{doi: #1}\else
  \providecommand{\doi}{doi: \begingroup \urlstyle{rm}\Url}\fi

\bibitem[Id{\'e} and Kashima(2004)]{ide2004eigenspace}
Tsuyoshi Id{\'e} and Hisashi Kashima.
\newblock Eigenspace-based anomaly detection in computer systems.
\newblock In \emph{Proceedings of the tenth ACM SIGKDD international conference
  on Knowledge discovery and data mining}, pages 440--449. ACM, 2004.

\bibitem[Hewapathirana(2019)]{hewapathirana2019change}
Isuru~U Hewapathirana.
\newblock Change detection in dynamic attributed networks.
\newblock \emph{Wiley Interdisciplinary Reviews: Data Mining and Knowledge
  Discovery}, 9\penalty0 (3):\penalty0 e1286, 2019.

\bibitem[Sricharan and Das(2014)]{sricharan2014localizing}
Kumar Sricharan and Kamalika Das.
\newblock Localizing anomalous changes in time-evolving graphs.
\newblock In \emph{Proceedings of the 2014 ACM SIGMOD international conference
  on Management of data}, pages 1347--1358. ACM, 2014.

\bibitem[Skillicorn(2007)]{skillicorn2007understanding}
David Skillicorn.
\newblock \emph{Understanding complex datasets: data mining with matrix
  decompositions}.
\newblock CRC press, 2007.

\bibitem[Saerens et~al.(2004)Saerens, Fouss, Yen, and
  Dupont]{saerens2004principal}
Marco Saerens, Francois Fouss, Luh Yen, and Pierre Dupont.
\newblock The principal components analysis of a graph, and its relationships
  to spectral clustering.
\newblock In \emph{European Conference on Machine Learning}, pages 371--383.
  Springer, 2004.

\bibitem[Neil et~al.(2013)Neil, Hash, Brugh, Fisk, and Storlie]{neil2013scan}
Joshua Neil, Curtis Hash, Alexander Brugh, Mike Fisk, and Curtis~B Storlie.
\newblock Scan statistics for the online detection of locally anomalous
  subgraphs.
\newblock \emph{Technometrics}, 55\penalty0 (4):\penalty0 403--414, 2013.

\bibitem[Heard et~al.(2010)Heard, Weston, Platanioti, Hand,
  et~al.]{heard2010bayesian}
Nicholas~A Heard, David~J Weston, Kiriaki Platanioti, David~J Hand, et~al.
\newblock Bayesian anomaly detection methods for social networks.
\newblock \emph{The Annals of Applied Statistics}, 4\penalty0 (2):\penalty0
  645--662, 2010.

\bibitem[Papadimitriou et~al.(2010)Papadimitriou, Dasdan, and
  Garcia-Molina]{papadimitriou2010web}
Panagiotis Papadimitriou, Ali Dasdan, and Hector Garcia-Molina.
\newblock Web graph similarity for anomaly detection.
\newblock \emph{Journal of Internet Services and Applications}, 1\penalty0
  (1):\penalty0 19--30, 2010.

\bibitem[Priebe et~al.(2005)Priebe, Conroy, Marchette, and
  Park]{priebe2005scan}
Carey~E Priebe, John~M Conroy, David~J Marchette, and Youngser Park.
\newblock Scan statistics on enron graphs.
\newblock \emph{Computational \& Mathematical Organization Theory}, 11\penalty0
  (3):\penalty0 229--247, 2005.

\bibitem[Gupta et~al.(2012)Gupta, Gao, Sun, and Han]{gupta2012integrating}
Manish Gupta, Jing Gao, Yizhou Sun, and Jiawei Han.
\newblock Integrating community matching and outlier detection for mining
  evolutionary community outliers.
\newblock In \emph{Proceedings of the 18th ACM SIGKDD international conference
  on Knowledge discovery and data mining}, pages 859--867. ACM, 2012.

\bibitem[Yu et~al.(2018)Yu, Woodall, and Tsui]{yu2018detecting}
Lisha Yu, William~H Woodall, and Kwok-Leung Tsui.
\newblock Detecting node propensity changes in the dynamic degree corrected
  stochastic block model.
\newblock \emph{Social Networks}, 54:\penalty0 209--227, 2018.

\bibitem[Akoglu and Faloutsos(2010)]{akoglu2010event}
Leman Akoglu and Christos Faloutsos.
\newblock Event detection in time series of mobile communication graphs.
\newblock In \emph{Army Science Conference}, pages 77--79, 2010.

\bibitem[Id{\'e} et~al.(2007)Id{\'e}, Papadimitriou, and
  Vlachos]{ide2007computing}
Tsuyoshi Id{\'e}, Spiros Papadimitriou, and Michail Vlachos.
\newblock Computing correlation anomaly scores using stochastic nearest
  neighbors.
\newblock In \emph{Data Mining, 2007. ICDM 2007. Seventh IEEE International
  Conference on}, pages 523--528. IEEE, 2007.

\bibitem[Sun et~al.(2008)Sun, Xie, Zhang, and Faloutsos]{sun2008less}
Jimeng Sun, Yinglian Xie, Hui Zhang, and Christos Faloutsos.
\newblock Less is more: Sparse graph mining with compact matrix decomposition.
\newblock \emph{Statistical Analysis and Data Mining}, 1\penalty0 (1):\penalty0
  6--22, 2008.

\bibitem[Sun et~al.(2006)Sun, Papadimitriou, and Philip]{sun2006window}
Jimeng Sun, Spiros Papadimitriou, and S~Yu Philip.
\newblock Window-based tensor analysis on high-dimensional and multi-aspect
  streams.
\newblock In \emph{ICDM}, volume~6, pages 1076--1080, 2006.

\bibitem[Papalexakis et~al.(2012)Papalexakis, Faloutsos, and
  Sidiropoulos]{papalexakis2012parcube}
Evangelos~E Papalexakis, Christos Faloutsos, and Nicholas~D Sidiropoulos.
\newblock Parcube: Sparse parallelizable tensor decompositions.
\newblock In \emph{Machine Learning and Knowledge Discovery in Databases},
  pages 521--536. Springer, 2012.

\bibitem[Sengupta and Chen(2015)]{sengupta2015Hetro}
Srijan Sengupta and Yuguo Chen.
\newblock Spectral clustering in heterogeneous networks.
\newblock \emph{Statistica Sinica:Vol. 25, No. 3, 1081-1106}, 2015.

\bibitem[Dryden and Mardia(1998)]{dryden1998statistical}
Ian~L Dryden and Kanti~V Mardia.
\newblock \emph{Statistical shape analysis}, volume~4.
\newblock Wiley Chichester, 1998.

\bibitem[Klimt and Yang(2004)]{klimt2004introducing}
Bryan Klimt and Yiming Yang.
\newblock Introducing the enron corpus.
\newblock In \emph{CEAS}, 2004.

\bibitem[Hoff et~al.(2002)Hoff, Raftery, and Handcock]{hoff2002latent}
Peter~D Hoff, Adrian~E Raftery, and Mark~S Handcock.
\newblock Latent space approaches to social network analysis.
\newblock \emph{Journal of the american Statistical association}, 97\penalty0
  (460):\penalty0 1090--1098, 2002.

\bibitem[Nickel(2007)]{nickel2007random}
Christine Leigh~Myers Nickel.
\newblock \emph{Random dot product graphs: A model for social networks},
  volume~68.
\newblock 2007.

\bibitem[Joseph and Yu(2013)]{joseph2013impact}
Antony Joseph and Bin Yu.
\newblock Impact of regularization on spectral clustering.
\newblock \emph{arXiv preprint arXiv:1312.1733}, 2013.

\bibitem[Chung(1997)]{chung1997spectral}
Fan~RK Chung.
\newblock \emph{Spectral graph theory}, volume~92.
\newblock American Mathematical Soc., 1997.

\bibitem[Amini et~al.(2013{\natexlab{a}})Amini, Chen, Bickel, Levina,
  et~al.]{amini2013pseudo}
Arash~A Amini, Aiyou Chen, Peter~J Bickel, Elizaveta Levina, et~al.
\newblock Pseudo-likelihood methods for community detection in large sparse
  networks.
\newblock \emph{The Annals of Statistics}, 41\penalty0 (4):\penalty0
  2097--2122, 2013{\natexlab{a}}.

\bibitem[Brand and Huang(2003)]{brand2003unifying}
Matthew Brand and Kun Huang.
\newblock A unifying theorem for spectral embedding and clustering.
\newblock In \emph{Proceedings of the Ninth International Workshop on
  Artificial Intelligence and Statistics}, 2003.

\bibitem[Achlioptas and McSherry(2007)]{achlioptas2007fast}
Dimitris Achlioptas and Frank McSherry.
\newblock Fast computation of low-rank matrix approximations.
\newblock \emph{Journal of the ACM (JACM)}, 54\penalty0 (2):\penalty0 9, 2007.

\bibitem[Stegmann and Gomez(2002)]{stegmann2002brief}
Mikkel~B Stegmann and David~Delgado Gomez.
\newblock A brief introduction to statistical shape analysis.
\newblock \emph{Informatics and mathematical modelling, Technical University of
  Denmark, DTU}, 15:\penalty0 11, 2002.

\bibitem[Tang et~al.(2012)Tang, Wang, and Liu]{tang2012community}
Lei Tang, Xufei Wang, and Huan Liu.
\newblock Community detection via heterogeneous interaction analysis.
\newblock \emph{Data Mining and Knowledge Discovery}, 25\penalty0 (1):\penalty0
  1--33, 2012.

\bibitem[Mihail and Papadimitriou(2002)]{mihail2002eigenvalue}
Milena Mihail and Christos Papadimitriou.
\newblock On the eigenvalue power law.
\newblock In \emph{International Workshop on Randomization and Approximation
  Techniques in Computer Science}, pages 254--262. Springer, 2002.

\bibitem[Amini et~al.(2013{\natexlab{b}})Amini, Chen, Bickel, and
  Levina]{amini2013}
Arash~A. Amini, Aiyou Chen, Peter~J. Bickel, and Elizaveta Levina.
\newblock Pseudo-likelihood methods for community detection in large sparse
  networks.
\newblock \emph{Ann. Statist.}, 41\penalty0 (4):\penalty0 2097--2122, 08
  2013{\natexlab{b}}.
\newblock \doi{10.1214/13-AOS1138}.
\newblock URL \url{http://dx.doi.org/10.1214/13-AOS1138}.

\bibitem[Ng et~al.(2001)Ng, Jordan, and Weiss]{ng2001spectral}
Andrew~Y Ng, Michael~I Jordan, and Yair Weiss.
\newblock On spectral clustering1 analysis and an algorithm.
\newblock \emph{Proceedings of Advances in Neural Information Processing
  Systems. Cambridge, MA: MIT Press}, 14:\penalty0 849--856, 2001.

\bibitem[Von~Luxburg(2007)]{von2007tutorial}
Ulrike Von~Luxburg.
\newblock A tutorial on spectral clustering.
\newblock \emph{Statistics and computing}, 17\penalty0 (4):\penalty0 395--416,
  2007.

\bibitem[Yu et~al.(2019)Yu, Zwetsloot, Stevens, Wilson, and
  Tsui]{yu2019monitoring}
Lisha Yu, Inez~M Zwetsloot, Nathaniel~T Stevens, James~D Wilson, and Kwok~Leung
  Tsui.
\newblock Monitoring dynamic networks: a simulation-based strategy for
  comparing monitoring methods and a comparative study.
\newblock \emph{arXiv preprint arXiv:1905.10302}, 2019.

\bibitem[Wang et~al.(2017)Wang, Chakrabarti, Sivakoff, and
  Parthasarathy]{wang2017fast}
Yu~Wang, Aniket Chakrabarti, David Sivakoff, and Srinivasan Parthasarathy.
\newblock Fast change point detection on dynamic social networks.
\newblock \emph{arXiv preprint arXiv:1705.07325}, 2017.

\bibitem[Karrer and Newman(2011)]{karrer2011stochastic}
Brian Karrer and Mark~EJ Newman.
\newblock Stochastic blockmodels and community structure in networks.
\newblock \emph{Physical Review E}, 83\penalty0 (1):\penalty0 016107, 2011.

\bibitem[Clauset et~al.(2009)Clauset, Shalizi, and Newman]{clauset2009power}
Aaron Clauset, Cosma~Rohilla Shalizi, and Mark~EJ Newman.
\newblock Power-law distributions in empirical data.
\newblock \emph{SIAM review}, 51\penalty0 (4):\penalty0 661--703, 2009.

\bibitem[De~Ridder et~al.(2016)De~Ridder, Vandermarliere, and
  Ryckebusch]{de2016detection}
Simon De~Ridder, Benjamin Vandermarliere, and Jan Ryckebusch.
\newblock Detection and localization of change points in temporal networks with
  the aid of stochastic block models.
\newblock \emph{Journal of Statistical Mechanics: Theory and Experiment},
  2016\penalty0 (11):\penalty0 113302, 2016.

\bibitem[Martin et~al.(2014)Martin, Zhang, and Newman]{martin2014localization}
Travis Martin, Xiao Zhang, and MEJ Newman.
\newblock Localization and centrality in networks.
\newblock \emph{Physical Review E}, 90\penalty0 (5):\penalty0 052808, 2014.

\bibitem[Poole(2014)]{poole2014linear}
David Poole.
\newblock \emph{Linear algebra: A modern introduction}.
\newblock Cengage Learning, 2014.

\bibitem[Peel and Clauset(2015)]{peel2015detecting}
Leto Peel and Aaron Clauset.
\newblock Detecting change points in the large-scale structure of evolving
  networks.
\newblock In \emph{Twenty-Ninth AAAI Conference on Artificial Intelligence},
  2015.

\bibitem[Rossi et~al.(2013)Rossi, Gallagher, Neville, and
  Henderson]{rossi2013modeling}
Ryan~A Rossi, Brian Gallagher, Jennifer Neville, and Keith Henderson.
\newblock Modeling dynamic behavior in large evolving graphs.
\newblock In \emph{Proceedings of the sixth ACM international conference on Web
  search and data mining}, pages 667--676. ACM, 2013.

\bibitem[Tang et~al.(2008)Tang, Liu, Zhang, and Nazeri]{tang2008community}
Lei Tang, Huan Liu, Jianping Zhang, and Zohreh Nazeri.
\newblock Community evolution in dynamic multi-mode networks.
\newblock In \emph{Proceedings of the 14th ACM SIGKDD international conference
  on Knowledge discovery and data mining}, pages 677--685. ACM, 2008.

\bibitem[Thomas(2002)]{thomas2002rise}
C~William Thomas.
\newblock The rise and fall of enron.
\newblock \emph{JOURNAL OF ACCOUNTANCY-NEW YORK-}, 193\penalty0 (4):\penalty0
  41--52, 2002.

\bibitem[Jackson(1993)]{jackson1993stopping}
Donald~A Jackson.
\newblock Stopping rules in principal components analysis: a comparison of
  heuristical and statistical approaches.
\newblock \emph{Ecology}, pages 2204--2214, 1993.

\bibitem[Jolliffe(2002)]{jolliffe2002principal}
Ian Jolliffe.
\newblock \emph{Principal component analysis}.
\newblock Wiley Online Library, 2002.

\bibitem[Cootes et~al.(1992)Cootes, Taylor, Cooper, and
  Graham]{cootes1992training}
Timothy~F Cootes, Christopher~J Taylor, David~H Cooper, and Jim Graham.
\newblock Training models of shape from sets of examples.
\newblock In \emph{BMVC92}, pages 9--18. Springer, 1992.

\end{thebibliography}

%% Non-BibTeX users please use
%\begin{thebibliography}{}
%%
%% and use \bibitem to create references. Consult the Instructions
%% for authors for reference list style.
%\end{thebibliography}

\end{document}